\documentclass{bmvc2k}

\usepackage{amssymb}
\usepackage{amsmath}
\usepackage{wrapfig}

\title{Can We Predict The Human Preference For Text-to-Image Content Prior To Generation And Is It Even Useful To Do So?}

\addauthor{Joong Ho Kim}{jkim5@lsu.edu}{1}
\addauthor{Keith G. Mills}{keith.mills@lsu.edu}{1}

\addinstitution{
 LSU ATHENA Lab\\
 Baton Rouge, LA, U.S.A. 70803
}

\runninghead{Kim and Mills}{Human Preference Prediction for T2I Generation}

\newcommand{\blue}[1]{\textcolor{blue}{#1}}
\newcommand{\red}[1]{\textcolor{red}{#1}}
\newcommand{\scaleboxratio}{0.8}

\def\etal{\emph{et al}\bmvaOneDot}

\usepackage{graphicx}
\usepackage{booktabs}
\usepackage{xcolor}
\usepackage{multicol}
\usepackage{multirow}
\usepackage{pdflscape}
\usepackage{float}
\usepackage{xfrac}

\begin{document}

\maketitle

\begin{abstract}
Diffusion Models (DM) have revolutionized text-driven generation by enabling the synthesis of high-quality, photorealistic visual content %
from %
user %
prompts. Whereas prior advances in visual generation such as VAEs and GANs were primarily evaluated on perceptual or visual similarity metrics such as FID %
PSNR, DM advances have fostered the development of more advanced Human Preference Metrics (HPM) that %
model and quantify human judgment as scalar values. %

However, DMs synthesize content using an inherently stochastic process where random noise seeds %
generation. %
The initial random noise %
directly affects the quality of generated outputs, both qualitatively and quantitatively. This influence is %
pronounced in smaller models for local deployment scenarios. Given this phenomenon, we first investigate to what extent we can predict scalar HPM %
scores %
prior to committing compute resources for generation. %
Further, we then investigate to what extent we can leverage such prediction to improve the quality of generated images, and also study which HPMs %
are best suited for this task. Our investigation reveals that not only is this possible, but that it is feasible to achieve %
negligible hardware overhead.
Code is available at \url{https://github.com/LSU-ATHENA/HPM-Predict}.

\end{abstract}

\section{Introduction}
\label{sec:intro}

Diffusion Models (DM)~\cite{Rombach_2022_CVPR} have become the de facto tool for facilitating open-source generation of photorealistic visual content~\cite{comfyui_contributors2025, autoamtic1111-stable-diffusion-webui} such as images~\cite{bfl2024Flux} and videos~\cite{yang2024cogvideox, kong2024hunyuanvideo} in local deployment scenarios~\cite{chen2025sana}. Whereas prior visual generation models such as Variational AutoEncoders (VAE)~\cite{kingma2013auto} and Generative Adversarial Networks (GAN)~\cite{goodfellow2014generative} were primarily evaluated on distance and noise metrics such as the Fr\'echet Inception Distance (FID)~\cite{heusel2017gans} and Peak Signal-to-Noise Ratio (PSNR)~\cite{dong2016accelerating}, Text-to-Image (T2I) DMs such as SDXL~\cite{podell2023sdxl, lykon2023dreamshaper} and Diffusion Transformers (DiT)~\cite{chen2024pixartsigma, li2024hunyuandit} excel at generating visual content from textual prompts~\cite{hessel2021clipscore}. These advances in generative performance have given rise to Human Preference Metrics (HPM)~\cite{wu2023human, ma2025hpsv3, xu2023imagereward, kirstain2023pick} which %
attempt to quantify to what extent a human would find a generative image desirable, given the textual prompt used to generate it.

\begin{figure}[t!]
    \centering
    \includegraphics[width=0.8\linewidth]{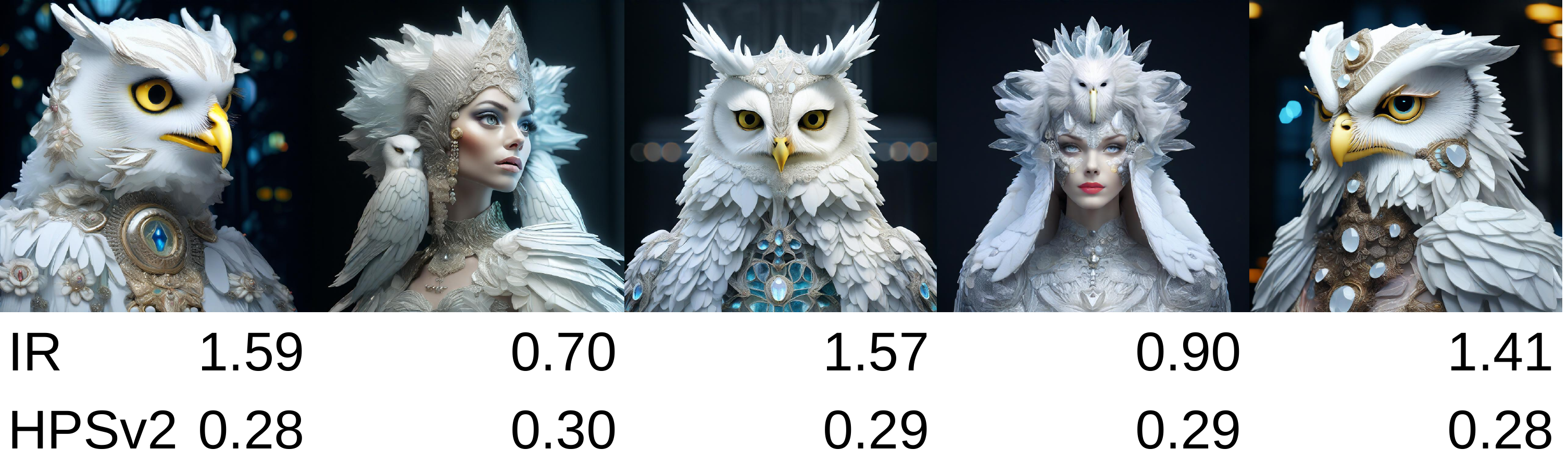}
    \caption{Illustrating the effect of initial random noise on DM output, both qualitatively and quantitatively. Prompt: ``anthropomorphic profile of the white snow owl Crystal priestess...'' from \cite{pixartalpha_prompts}, generated by Hunyuan-DiT, then scored by ImageReward~\cite{xu2023imagereward} and HPSv2~\cite{wu2023human}.}
    \label{fig:noise_example_images}
\end{figure}

However, DMs use a costly iterative process seeded by random noise which has been shown to affect generated outputs both qualitatively~\cite{wang2025silent} and quantitatively~\cite{zhou2025golden, eyring2024reno}, %
i.e., images, as shown in Figure~\ref{fig:noise_example_images}. This motivates a simple question: Can we mitigate the effect of the initial noise on generative quality? Existing techniques in the literature %
range from examining the interaction between prompt and noise inside the DM cross-attention mechanism~\cite{guo2024initno} to adjusting the initial noise based on the prompt~\cite{zhou2025golden} prior to generation to costlier fine-tuning~\cite{li2025noisear} and RLHF-based~\cite{eyring2025noise} approaches. However, such approaches overlook the relationship that exists between the initial random noise and the resultant HPM scores corresponding to the image generated by said noise.

This paper answers the question: Given a textual prompt, how can we preemptively choose a better initial noise prior to performing the costly DM generation process that will yield a superior output from a human preference point of view? We answer this question by separately examining %
the extent to which HPM 
scores can be estimated from DM initial conditions, whether optimizing based on these estimates can yield superior generative output images, alongside an examination of which human preference metrics are best suited to this task. Our detailed contributions are as follows:

First, we question whether DM initial conditions, i.e., the textual prompt and initial noise, carry sufficient signal information to adequately estimate the HPM scores %
of the image they would generate. We answer this question by training performance predictors~\cite{white2021powerful, mills2025qua2sedimo} that leverage existing ideas in initial noise optimization~\cite{zhou2025golden, li2025noisear, eyring2025noise, kalaivanan2025ess, smith2025calibrating, om2025posterior, wang2025source, venkatraman2025outsourced}, such as examining the DM cross-attention map~\cite{guo2024initno}, as well as broader literature in low-cost estimators~\cite{lu2023pinat, han2023general} which provides several solutions at varying compute-cost levels. Further, we study the efficacy of different training losses and evaluation metrics for these predictors. 

Second, leveraging these findings, we integrate our HPM %
predictors into existing T2I DM pipelines. Specifically, given the textual prompt, we then apply a simple Best-of-N (BoN)~\cite{ghasemabadi2025guided, brown2024large} optimization sort using our predictors, then select the best noise to generate an image from. We then measure the HPM scores %
for this image. %
This provides a more meaningful evaluation of our preference score predictors by considering to what extent they actually improve the quality of generated outputs. Furthermore, this allows us to measure and compare the compute cost overhead of our predictors in an end-to-end scenario akin to what a real-world user would face.

Third, we provide a lateral study comparing the efficacy of different HPMs%
~\cite{wu2023human, ma2025hpsv3, xu2023imagereward, kirstain2023pick}. %
Specifically, we question what human preference metrics are easier to train a predictor on, which lead to better DM outputs when integrated into an end-to-end DM pipeline, as well as which HPMs %
tend to agree or disagree with each other when examining images generated via our method compared to a baseline approach.

We conduct extensive experiments on several convolutional U-Net and Transformer~\cite{peebles2023scalable} DMs such as SDXL~\cite{podell2023sdxl}, DreamShaper~\cite{lykon2023dreamshaper}, Hunyuan-DiT~\cite{li2024hunyuandit} and PixArt-$\Sigma$~\cite{chen2024pixartsigma} across a slew of HPMs %
such as HPSv2~\cite{wu2023human}, HPSv3~\cite{ma2025hpsv3}, ImageReward~\cite{xu2023imagereward} and PickScore~\cite{kirstain2023pick}. Our results show that it is indeed possible to broadly improve DM generative quality using a human preference performance predictor; however, care must be taken when selecting which preference metrics to train the predictor on, and that simpler predictors that do not rely on the DM at all when performing BoN not only achieve superior average gains, but do so at a significantly reduced computational cost on multiple hardware devices.

\section{Background and Related Work}
\label{sec:background}
Diffusion Models (DM)~\cite{ho2022classifier} are generative AI models that synthesize content such as images~\cite{podell2023sdxl, chen2024pixartsigma, chen2025sana}, videos~\cite{ho2022video, chen2025rettention}, audio~\cite{huang2023make}, molecules~\cite{wang2025diffusion}, neural network weights~\cite{soro2024diffusion}, etc.~\cite{saxena2023surprising}, by initially sampling a random noise tensor $X_T\sim \mathcal{N}(0, 1)$. Inspired by thermodynamic processes~\cite{sohl2015deep}, a DM gradually refines $X_T$ using a reverse diffusion process (RDP) over a series of discrete timesteps $t\in [T,...,0]$. There are many forms of RDP equations~\cite{daras2024survey, karras2022elucidating}. For example, one of the simplest RDPs is the Denoising Diffusion Implicit Model (DDIM) RDP~\cite{song2020denoising}, formally given by %

\begin{equation}
    \centering
    X_{t-1} = \alpha_{t-1} (\dfrac{X_t - \sigma_t\epsilon_\theta (X_t, t, c)}{\alpha_t}) + \sigma_{t-1}\epsilon_{\theta}(X_t, t, c), 
    \label{eq:ddim}
\end{equation}
where $\alpha_t$ and $\sigma_t$ are determined by a pre-defined scheduler and $\epsilon_\theta$ is the \textit{denoiser} deep neural network (DNN), typically a large U-Net~\cite{kirillov2019panoptic} or Diffusion Transformer (DiT)~\cite{peebles2023scalable}. After a preset number of iterations, the RDP terminates $t=0$, yielding the final output $X_0$. In the case of traditional DMs~\cite{sohl2015deep}, $X_0$ could be a 3-channel $C=3$ RGB image $I$ with specified height $H_I$ and width $W_I$. 

Latent Diffusion Models (LDM)~\cite{Rombach_2022_CVPR} are largely responsible for the recent popularity in DM research and open-source usage by end-users. LDMs operate by performing prediction in the latent space $z$ of a Variational AutoEncoder (VAE)~\cite{kingma2013auto}. In this case, $X_0=z$ would not be an RGB image with %
$C=3$ channels, but would instead be passed through a VAE decoder $g(z)$ in order to generate the final output, i.e., an RGB image $I=g(z)=g(X_0)$. %
Importantly, $X_T \sim \mathcal{N}(0, 1)$ is still a stochastically sampled tensor for LDMs. 

\subsection{Text-Driven Diffusion}
\label{sec:t2i}

Text-to-Image (T2I) generation further involves a user-defined textual prompt, %
i.e., `A crocodile in a sweater'~\cite{pixartalpha_prompts}. Specifically, given a prompt $p$, we first tokenize~\cite{salameh2024autogo} and encode the prompt into the conditioning matrix $c\in \mathbb{R}^{|p_{tok}|\times d_c}$ where $p_{tok}$ is the tokenized prompt and $d_c$ is the token embedding dimension. The format of the denoiser DNN changes from $\epsilon_{\theta}(X_t, t)$ to $\epsilon_{\theta}(X_t, t, c)$ to accommodate the prompt conditioning, but the overarching %
RDP is unchanged as $c$ is constant across all timesteps.

Internally, cross-attention~\cite{vaswani2017attention} facilitates interaction between the noise $X_t$ and prompt conditioning $c$, regardless of whether the denoiser $\epsilon_\theta$ is a convolutional U-Net~\cite{Rombach_2022_CVPR, podell2023sdxl, civitaiIllustriousV20, civitaiJuggernautRagnarok_by_RunDiffusion} or DiT~\cite{chen2024pixartAlpha, chen2024pixartsigma, esser2024scaling, bfl2024Flux, chen2025sana}. Specifically, inside the denoiser DNN, the noise tensor is translated into a patch sequence~\cite{dosovitskiy2020image} to form the \textit{query} aspect of the attention mechanism; $Q_{X}\in \mathbb{R}^{h\times P \times d_h}$ where $P$ is the number of patches while $h$ and $d_h$ are the number of attention heads and head embedding dimension, respectively. %
Likewise, the prompt conditioning forms the \textit{key} and \textit{value} attention tensors, $K_c, V_c\in \mathbb{R}^{h\times|p_{tok}|\times d_h}$. Thus, the noise and prompt interact to compute the softmax attention map $\texttt{softmax}(Q_XK_c^T)\in [0, 1]^{h\times P\times|p_{tok}|}$ which determines what prompt tokens each noise patch will attend to and draw information from. 

The key source of variability in DMs, especially locally-deployable DMs, is that $X_T$ is stochastically sampled and this randomness is propagated into the query $Q_X$ and cross-attention mechanism. 
Figure~\ref{fig:noise_example_images} provides a sample illustration, where given the same prompt, i.e., $c$ is held constant, yet different $X_T$ cause %
drastically different images to form, leading %
to a qualitatively and quantitatively different outputs. %
Such an issue is of importance given that the denoiser $\epsilon_\theta$ is a large DNN, typically consisting of ranging from 600M parameters~\cite{chen2024pixartAlpha, chen2024pixartsigma} to at least several billion~\cite{li2024hunyuandit, podell2023sdxl, esser2024scaling} parameters that, per the RDP, performs multiple rounds of inference to generate an output whose quality varies with $X_T$ which is not even provided by the user. This motivates the need to quantify the quality of a DM output and attempt to control, mitigate, or optimize in the reality where $X_T$ is stochastic.

\subsection{T2I Evaluation via Human Preference Estimation}
\label{sec:human_preference}

The goal of T2I generation is to produce an image $I$ which the end-user believes adequately realizes their provided textual prompt $p$. This objective is easier to describe than execute, as, unlike older visual generation metrics like FID~\cite{hessel2021clipscore}, PSNR~\cite{dong2016accelerating} and CLIPScore~\cite{hessel2021clipscore}, 
human preferences carry a degree of subjectivity and are more difficult to quantify, usually requiring user studies and surveys~\cite{mills2025qua2sedimo, chen2025fp4dit, podell2023sdxl, chen2024pixartAlpha}.

However, the advent of T2I DMs has motivated the development of \textit{learned} human preference reward models or Human Preference Metrics (HPM), such as Human Preference Score v2 (HPSv2)~\cite{wu2023human}, HPSv3~\cite{ma2025hpsv3}, ImageReward (IR)~\cite{xu2023imagereward}, PickScore (PS)~\cite{kirstain2023pick}, etc.~\cite{Zhang_2024_CVPR}. Specifically, human preference metrics perform the heavy lifting involved in executing a user study, such as experimental design and participant recruitment. Once the study is complete, the results are further utilized to train a deep neural network (DNN) predictor model $\phi(p, I)$ which receives a user-provided prompt $p$ and T2I-generated image $I$ as input, and then estimates a \textit{human preference score} $S_{p, I}\in \mathbb{R}$. Formally, 

\begin{equation}
    \centering
    S_{p,I} = \phi(p, I),
    \label{eq:hps}
\end{equation}
where higher $S_{p, I}$ indicates a more preferable image. Recently, these methods have been shown to form a reliable set of benchmarks for accurately evaluating new T2I models~\cite{xie2024sana}, as well as performing Reinforcement Learning via Human Feedback (RLHF)~\cite{wu2026densedpo, li2025noisear, Wallace_2024_CVPR, eyring2025noise, eyring2024reno} and especially for measuring the efficacy of DM efficiency~\cite{chen2025fp4dit, chen2025rettention} and optimization~\cite{jiang2026raise} techniques, such as initial noise optimization~\cite{zhou2025golden}. The goal of this paper is to study to what extent we can further levage these advances to achieve effective initial noise optimization, and how best to do so efficiently, as we next explain.

\section{Methodology}
\label{sec:method}

\begin{figure}[t!]
    \centering
    \includegraphics[width=0.95\linewidth]{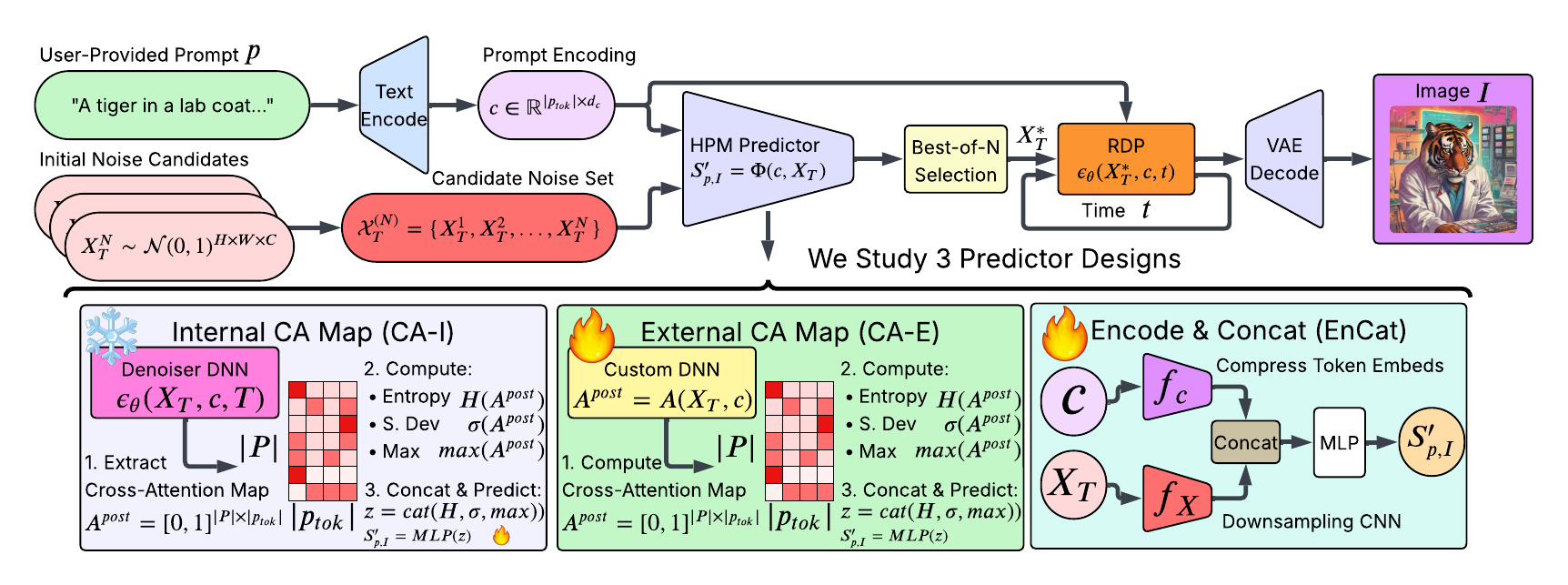}
    \caption{HPM predictor design. Given the DM initial conditions, we consider three types of predictors, abbreviated as CA-I, CA-E and EnCat, which produce $S_{p, I}'$ from $c$ and $X_T$. \blue{Snowflake} and \red{flame} denote if model weights are frozen or trainable, respectively.}
    \label{fig:overview}
    \vspace{-5mm}
\end{figure}

In this section we introduce a technique for achieving initial noise optimization via HPM estimation. Figure~\ref{fig:overview} provides a high-level overview of our technique. %
We first introduce the high-level idea for a predictor architecture, and then discuss how it can be combined with Best-of-N (BoN) in an existing DM pipeline. Next, we take a finer look at the design of our predictor, specifically emphasing how to model the interaction of $X_T$ and $p$, before finally providing an overview of how to train our predictor. 

\subsection{Human Preference Score Prediction from Initial Conditions}

Recall Eq.~\ref{eq:hps} from Sec.~\ref{sec:human_preference} which describes how HPMs %
train a predictor $\phi$ that maps a given prompt $p$ and image $I$ to a scalar score $S_{p, I}$ as way to model how desirable a human would find the image. In the context of latent T2I DMs, the image is the output of a VAE decoder $I=g(X_0)$ where $X_0$ is the final output of the iterative reverse diffusion process (RDP) described in Sec.~\ref{sec:background}. For simplicity, we will use $\texttt{RDP}$ as shorthand for Eq.~\ref{eq:ddim} and as an abstraction for an arbitrary RDP~\cite{ho2020denoising, karras2022elucidating}, i.e., $X_0=\texttt{RDP}(X_T, c)$ where $X_T\sim\mathcal{N}(0, 1)$ is the initial stochastic noise and $c$ is the encoding of prompt $p$.  

Given this setup, we ask a simple question: To what extent can %
we predict the human preference score $S_{p, I}$ from the DM initial conditions $c$ and $X_T$ and how best can we do so? Given the above, we reformulate the T2I generative and evaluation process as a \textit{scalar regression} prediction problem~\cite{white2021powerful, han2023general} as follows:

\begin{equation}
    \centering
    S_{p, I}' = \Phi(c, X_T),
    \label{eq:preference_estimation}
\end{equation}
where $\Phi$ is a \textit{new} human preference predictor model that, in contrast to the original $\phi$, predicts $S_{p, I}'$, the \textit{estimation} of $S_{p, I}$ given the user-provided textual prompt embedding $c$ and the initial noise $X_T$ utilized to generate an image $I$, rather than the image itself as $\phi$ does.

The utility %
behind this approach is straightforward: Generating an RGB image requires substantial compute, e.g., the large denoiser DNN $\varepsilon_\theta$ and RDP. This image, the prompt utilized to generate it, will have a corresponding human preference score $S_{p, I}$, and in fact we can re-write the equation for a human preference predictor to include the RDP by making the %
substitutions $S_{p, I}=\phi(p, I)=\phi(p, g(\texttt{RDP}(X_T, c))$ where $c$ is the encoding of $p$. 

This %
reframing makes the role of the initial stochastic noise $X_T$ on the human preference score $S_{p, I}$ more explicit. Except, whereas an RGB image is a 3D tensor $I\in\mathbb{R}^{3\times H_I \times W_I}$ that contains many different elements, the number of which scales quadratically with $H_IW_I$, the human preference score for the image $I$ is just a scalar value, $S_{p, I}\in\mathbb{R}$, which, according to information theory, should require less computation to estimate $S_{p, I}'$~\cite{tishby2000information, tishby2015deep, shannon1948mathematical, zammit2025neural}, i.e., using a smaller predictor~\cite{white2021powerful, han2023general, mills2025qua2sedimo, lu2023pinat} model which we denote as $\Phi$ in Eq.~\ref{eq:preference_estimation}.

Further, although at first glance, predicting HPM scores directly from Gaussian noise may appear unintuitive, %
in T2I DMs, %
the interaction between $X_T$ and the prompt encoding $c$ is %
established at the earliest stages of the RDP %
via cross-attention~\cite{guo2024initno}. This interaction determines how different regions of the latent representation align with prompt tokens, effectively biasing the trajectory of the RDP. %
As a result, certain noises %
are inherently more compatible with a given prompt than others~\cite{wang2025silent}, leading to systematic variation in downstream HPM scores. %
We hypothesize %
that this early-stage alignment signal is sufficiently informative to enable lightweight predictors $\Phi(c, X_T)$ to approximate the %
preference score without requiring full image generation. Moreover, should this hold, then we can leverage the reduced cost of $\Phi$ compared to $g(\texttt{RDP}(X_T, c))$ to improve existing DM pipelines in an end-to-end manner.

\subsection{Initial Noise Optimization via Best-of-N Sorting}
\label{sec:BoN}

Given a user-provided textual prompt $p$, a T2I DM encodes it into the tensor $c\in \mathbb{R}^{|p_{tok}|\times d_c}$ per Sec.~\ref{sec:t2i} to condition generation. The user does not select the initial noise, but it is instead randomly sampled; $X_T\sim\mathcal{N}(0, 1)$. Thus, prior to generation, instead of sampling a single initial noise, we can instead sample of \textit{pool} of $N$ candidate noises, $\mathcal{X}_T^{(N)}=\{X_T^{1}, X_T^{2},...X_T^{N}\}$. Then, we can use our predictor $\Phi$ to estimate the human preference score for each noise given the prompt per Eq.~\ref{eq:preference_estimation}. This ties a scalar value to each candidate noise, allowing us to perform Best-of-N (BoN)~\cite{brown2024large, ghasemabadi2025guided} selection by sorting the noises according to the score estimations and then select the noise $X_T^*$ with the highest estimated score: %
\begin{equation}
    \centering
    X_T^* = \operatorname*{arg\,max}_{X_T^{(j)} \in \mathcal{X}_T^{(N)}} \Phi(c, X_T^{(j)}).
    \label{eq:BoN}
\end{equation}
Then, only the selected noise $X_T^*$ is passed through the full DM RDP to generate an image which we can then evaluate. The idea is that by preemptitively optimizing the noise based on the estimated score $S_{p, I}'$, we will generate a more desirable image with a higher actual human preference score %
$S_{p, I}$. 

However, this procedure imposes additional overhead on the already-costly DM RDP. Depending on the exact DNN architecture specification of $\Phi$ and choice of $N$, that overhead can range from a negligible amount to something that has a noticeable impact on end-to-end generation times. Next, we discuss several potential options.

\subsection{Preference Predictor Design}
\label{sec:predictor_design}

Our predictor $\Phi(c, X_T)$ produces an estimate $S_{p, I}'$ of the human preference score $S_{p, I}$ for an image $I$ and prompt $p$ using the prompt conditioning $c$ and initial noise $X_T\sim\mathcal{N}(0, 1)$. The challenge in designing this predictor is two-fold: First, both $c$ and $X_T$ are multi-dimensional tensors whereas $S_{p, I}'$ will be a single scalar value. Second, the accuracy of $S_{p, I}'$, and therefore, its downstream utility per Sec.~\ref{sec:BoN} crucially depends on providing sufficient interaction between $c$ and $X_T$ so as to extract as much useful information~\cite{shannon1948mathematical} as possible. In this paper, we highlight three distinct formats of $\Phi$. Figure~\ref{fig:overview} provides a high-level illustration. %

\noindent\textbf{Internal Cross-Attention Mapping.} Initial Noise Optimization (InitNO)~\cite{guo2024initno} is one of the earliest works that improves DM generative quality by adjusting $X_T$. InitNO do not adjust $X_T$ \textit{prior} to generation, but instead apply edits to the DM's internal cross-attention map from Sec.~\ref{sec:t2i}. Taking inspiration from this, our first type of predictor receives as input, the first cross-attention map $A^{post}=\texttt{softmax}(Q_XK_c^T)\in [0, 1]^{h\times P\times|p_{tok}|}$ computed by the DM. We then compress the noise patch dimension $P$ by computing the entropy $A^{post}_H\in\mathbb{R}^{h\times|p_{tok}|}$, standard deviation $A^{post}_\sigma\in\mathbb{R}^{h\times|p_{tok}|}$ and max value $A^{post}_{max}\in [0, 1]^{h\times|p_{tok}|}$ of each row of the attention matrix. Next, for each of these tensors we apply a head compression factor $r; h\bmod r=0$, compressing the head dimension from $h$ to $\sfrac{h}{r}$ by summing each group of $r$ maps. We then flatten the compressed head dimension and concatenate these matrices. This gives rise to a vector $\vec{a}\in\mathbb{R}^{\sfrac{3h}{r}|p_{tok}|}$ which we 
feed into a simple multi-layer perception (MLP) to produce $S_{p, I}'$. We call this predictor type \textbf{Cross-Attention-Internal} or \textbf{CA-I} for short.

\noindent\textbf{External Cross-Attention Mapping.} Another way to model the interaction between $c$ and $X_T$ is to still rely on a cross-attention map but to manually compute it outside of the DM. %
Specifically, we apply a streamlined patchify convolution which converts $X_T$ into a patch sequence, positional encoding~\cite{dosovitskiy2020image}, and a simple cross-attention mechanism which pairs the patched $X_T$ with $c$. We apply the same set of transformers and reductions to the cross-attention map as CA-I to reduce it from a multi-dimensional tensor down to a vector, and then likewise feed it into an MLP. We dub this type of predictor \textbf{Cross-Attention-External} or \textbf{CA-E} for short. Unlike CA-I, we do not need to invoke partial inference on the DM %
to compute the $A^{post}$ used to produce $S_{p, I}'$.

\noindent\textbf{Traditional Encode-and-Concatenate.} Finally, we also consider a traditional form of multi-input, multi-modal predictor that encodes each input into a vector, then concatenates them together and feeds the result into an MLP. Specifically, this predictor %
contains a noise encoder $f_X$ which uses a simple ResNet-style~\cite{he2016identity} CNN to aggressively downsamples the noise from a 2D tensor into a vector. In parallel, a prompt encoder $f_c$ uses a series of linear layers to progressively shrink the initial prompt token embedding dimension $d_c$ down to $1$ to produce a vector of length $|p_{tok}|$. Finally, we concatenate the outputs of $f_X$ and $f_c$ together and feed the resultant vector into an MLP to produce $S_{p, I}'$. We dub this predictor \textbf{Encode-and-Concat} or \textbf{EnCat} for short.  

In the supplementary materials, we provide extensive implementation details and hyperparameters. Further, we also provide ablation studies comparing how well each of these predictors performs, comparing design, training loss functions and evaluation metrics. 

\subsection{Training Objective}
\label{sec:obj}

Regardless of the form our predictor $\Phi$ takes, its efficacy when integrated into an existing DM pipeline will depend on the choice of training loss. To simplify this choice, we consider a differentiable ranking/information retrieval loss paired with a traditional regression loss. This is formally given by 

\begin{equation}
    \centering
    \mathcal{L}_\Phi=\mathcal{L}_{rank} + \mathcal{L}_{reg},
    \label{eq:loss}
\end{equation}
where $\mathcal{L}_{rank}$ is either the Lambdaloss~\cite{wang2018lambdaloss} or a differentiable Spearman Rank Correlation Coefficient (SRCC)~\cite{blondel2020fast} loss computed over a given training batch, and $\mathcal{L}_{reg}$ is a regression loss, typically the Mean Absolute Error (MAE); $\mathcal{L}_{reg}=|S_{p, I}-S_{p, I}'|$. %

\section{Results}
\label{sec:results}
In this section, we present our experimental setup and results. We structure our results around three research questions (RQ):  
\begin{itemize}
    \item \textbf{RQ1:} Given only the DM initial conditions, the prompt encoding $c$ and %
    stochastic noise $X_T$, can we accurately estimate the human preference score $S_{p, I}$ using a predictor $\Phi(c, X_T)$? Further, if this is possible, how is this estimation best achieved? 
    \item \textbf{RQ2:} By pairing our predictor $\Phi(c, X_T)$ with a simple Best-of-N (BoN) sort to optimize $X_T$ prior to generation, can we increase the quality of generated outputs compared to an unoptimized baseline?
    \item \textbf{RQ3:} Which HPMs %
    should we aim to estimate? That is, if we compare different metrics, are some of them easier to predict or better for downstream noise optimization? 

\end{itemize}

To answer these questions, we consider several popular T2I DMs. First, we consider the 2.6B parameter U-Net SDXL~\cite{podell2023sdxl}, a very common base model in the open-source T2I scene that has given rise to many popular fine-tunes~\cite{civitaiCyberRealistic_Pony, civitaiIllustriousV20, civitaiJuggernautRagnarok_by_RunDiffusion} such as DreamShaper~\cite{lykon2023dreamshaper}, which we also consider. In terms of DiTs, we consider the lightweight 600M parameter PixArt-$\Sigma$~\cite{chen2024pixartsigma} as well as the larger 1.6B parameter Hunyuan~\cite{li2024hunyuandit}. For each of these DMs, we use the default checkpoint and hyperparameters from the HuggingFace Diffusers~\cite{von-platen-etal-2022-diffusers} library to generate $1024\times1024$ resolution images.

In terms of %
metrics we consider Human Preference Score v2 (HPSv2)~\cite{wu2023human}, HPSv3~\cite{ma2025hpsv3}, ImageReward (IR)~\cite{xu2023imagereward} and PickScore (PS)~\cite{wang2020picking}. Specifically, we train our predictor $\Phi$ on one of these, but then evaluate the end-to-end pipeline performance by generating images from different sets of prompts and using these metrics to judge the human preference for the generated images. In terms of downstream evaluation prompts, we consider two sets of prompts: First, following Golden Noise~\cite{zhou2025golden}, we consider the first 100 Pick-a-Pick validation set prompts as well as the larger HPSv2 benchmark, which contains 3.2k prompts spread over 4 categories.

\subsection{\textbf{RQ1:} Can Initial Conditions Predict Human Preference?}
\label{sec:offline_results}

\begin{table}[!t]
\begin{center}
\caption{Human preference predictor performance on DM %
initial conditions. We experiment across 4 DMs and use 2 HPMs as targets. For each DM and target, we train a CA-I, CA-E and EnCat predictor with on the LambdaLoss and MAE. We then evaluate the NDCG@$3$, NDCG@$5$ and MAE on the held-out test set and also report the predictor parameters. \textbf{Best} and \textit{second best} results in \textbf{bold} and \textit{italics}, respectively. }
\label{table:offline_prediction}
\resizebox{\linewidth}{!}{
\begin{tabular}{cclcccc}
\toprule
Model & Target & Predictor & Params & NDCG@$3$~(↑) & NDCG@$5$~(↑) & MAE~(↓) \\

\midrule
\multirow{6}{*}{SDXL}
& \multirow{3}{*}{HPSv2}
& {CA-I} & 1.33M & \textit{0.592} & 0.594 & \textbf{0.035} \\
& & {CA-E} & 1.11M & 0.589 & \textit{0.614} & 0.038 \\
& & {EnCat} & 3.50M & \textbf{0.599} & \textbf{0.628} & \textit{0.036} \\

\cline{2-7} 

& \multirow{3}{*}{PickScore}
& CA-I & 1.33M & \textit{0.570} & 0.607 & 1.243 \\
& & {CA-E} & 1.11M & 0.566 & \textit{0.614} & \textit{0.899} \\
& & {EnCat} & 3.50M & \textbf{0.609} & \textbf{0.625} & \textbf{0.811} \\
\midrule

\multirow{6}{*}{DreamShaper}
& \multirow{3}{*}{HPSv2}
& CA-I & 1.33M & 0.595 & 0.626 & 0.037 \\
& & {CA-E} & 1.11M & \textbf{0.612} & \textit{0.640} & \textit{0.028} \\
& & {EnCat} & 3.50M & \textit{0.608} & \textbf{0.647} & \textbf{0.026} \\

\cline{2-7} 

& \multirow{3}{*}{PickScore}
& {CA-I} & 1.33M & \textit{0.588} & \textit{0.627} & 1.179 \\
& & CA-E & 1.11M & 0.580 & 0.614 & \textit{1.001} \\
& & {EnCat} & 3.50M & \textbf{0.590} & \textbf{0.630} & \textbf{0.971} \\
\midrule

\multirow{6}{*}{Hunyuan-DiT}
& \multirow{3}{*}{HPSv2}
& CA-I & 8.33M & \textbf{0.594} & 0.622 & \textit{0.035} \\
& & {CA-E} & 1.48M & 0.589 & \textbf{0.626} & \textbf{0.024} \\
& & {EnCat} & 3.59M & \textit{0.591} & \textit{0.625} & \textbf{0.024} \\

\cline{2-7} 

& \multirow{3}{*}{PickScore}
& {CA-I} & 8.33M & \textbf{0.585} & \textbf{0.624} & 1.201 \\
& & CA-E & 1.48M & 0.576 & 0.611 & \textit{0.981} \\
& & {EnCat} & 3.59M & \textit{0.580} & \textit{0.620} & \textbf{0.872} \\
\midrule

\multirow{6}{*}{PixArt-$\Sigma$}
& \multirow{3}{*}{HPSv2}
& CA-I & 7.52M & 0.581 & 0.617 & 0.035 \\
& & {CA-E} & 2.09M & \textbf{0.586} & \textit{0.619} & \textit{0.031} \\
& & {EnCat} & 12.01M & \textit{0.583} & \textbf{0.620} & \textbf{0.030} \\

\cline{2-7} 

& \multirow{3}{*}{PickScore}
& {CA-I} & 7.52M & \textit{0.564} & \textit{0.604} & 1.242 \\
& & {CA-E} & 2.09M & 0.575 & \textbf{0.614} & \textit{1.016} \\
& & EnCat & 12.00M & \textbf{0.554} & 0.595 & \textbf{0.979} \\
\bottomrule
\end{tabular}
}
\vspace{-5mm}
\end{center}
\end{table}

First, we consider whether it is feasible to estimate human preference scores $S_{p, I}$ for a given HPM from the initial prompt $c$ and noise $X_T$. For each DM, we train several predictors 
$\Phi$ on tuples of prompt-noise-score $(c, X_T, S_{p, I})$ data. Following \cite{zhou2025golden}, we sample 5k prompts from the Pick-a-Pick~\cite{kirstain2023pick} training dataset and generate 20 initial noises per prompt, for 100k tuples in total per DM. We partition each dataset $80\%/10\%/10\%$ into disjoint train-validation-test sets along prompt indices; i.e, the prompts themselves are disjoint with respect to our dataset partitions. Following Sec.~\ref{sec:obj} we train using Eq.~\ref{eq:loss} on MAE and LambdaLoss, while reserving the differentiable SRCC for the supplementary. We also vary our measurements across two separate HPMs: %
HPSv2 and PickScore. Additional training details can be found in the supplementary. %

Table~\ref{table:offline_prediction} reports our results. Between the choice of DM, predictor training target and evaluation metric, there are 24 trials and EnCat achieves the best performance over half the time (15), and the second best performance in a significant number of the remaining trials (7). However, with exception to Hunyuan, the EnCat encoder contains significantly more parameters than both the CA-I and CA-E predictors. Ironically, the CA-E predictor, which externally computes the cross-attention map always contains the fewest number of parameters, yet is arguably the 2nd best design by predictor metrics, as it achieves the best and second best results more frequently than CA-I. Overall, these results demonstrate that it is indeed feasible to predict human preference scores $S_{p, I}$ from initial conditions. The question is whether accurate prediction entails improvement in generative quality.

%\begin{table}[!t]
\begin{wraptable}{lt}{0.75\textwidth}
\begin{center}
\vspace{-5mm}
\caption{Comparison of regression prediction and downstream HPM results on DreamShaper. We use MAE+LambdaLoss to train each predictor, varying the HPM from PickScore to HPSv2. The final column reports the downstream generation score corresponding to the target metric. `Standard' is an unoptimized baseline without using a predictor. Best and second best results in \textbf{bold} and \textit{italics}, respectively.}
\label{table:dreamshaper_loss_ablation}
\scalebox{\scaleboxratio}{
\begin{tabular}{cllccccc}
\toprule
Model & Target & Method
& \multicolumn{2}{c}{MAE+LambdaLoss}
& Downstream \\
\cmidrule(lr){4-5}
\cmidrule(lr){6-6}
& & & NDCG@$5$ & SRCC & Target Score \\
\midrule
\multirow{8}{*}{Dreamshaper}
& \multirow{4}{*}{PickScore}
& Standard & -- & -- & 22.6041 \\
& & CA-I  & 0.6266 & 0.2327 & 22.6096 \\
& & CA-E  & \textit{0.6135} & \textit{0.5637} & \textbf{22.6238} \\
& & EnCat  & \textbf{0.6295} & \textbf{0.5901} & \textit{22.6146} \\
\cmidrule(lr){2-6}
& \multirow{4}{*}{HPSv2}
& Standard & -- & -- & 0.3461 \\
& & CA-I & 0.6255 & 0.1614 & 0.3456 \\
& & CA-E & \textit{0.6397} & \textit{0.6756} & \textbf{0.3467} \\
& & EnCat & \textbf{0.6467} & \textbf{0.7337} & \textit{0.3464} \\
\bottomrule
\end{tabular}
}
\end{center}
\vspace{-8mm}
%\end{table}
\end{wraptable}

Further, Table~\ref{table:dreamshaper_loss_ablation} provides this result for DreamShaper.\\Specifically, we integrate our predictors into the existing DM pipeline and use the BoN sorting from Sec.~\ref{sec:BoN} with $N=100$ to generate images. We then measure the score $S_{p, I}$ for these images according to the source HPM %
the predictor %
trained on (i.e., PickScore on top, HPSv2 on bottom). We then report the average score across all prompts. We compare this to a baseline `Standard' that does not utilize BoN sorting at all. In contrast, both CA-E and EnCat achieve respectable SRCC despite training on LambdaLoss and also consistently outperform the standard baseline. %

\subsection{RQ2: Does Predictor-Guided Noise Selection Improve $S_{p, I}$?}
\label{sec:downstream_results}

\begin{table}[!b]
\begin{center}
\caption{Downstream Best-of-N evaluation using predictor-guided initial noise selection with $N=100$. We report aligned evaluations: HPSv2-trained predictors on HPSv2 benchmark prompts and PickScore-trained predictors on PickScore/Pick-a-Pic prompts. Higher is better for all metrics. Best and second-best results within each model, target, and prompt set are shown in \textbf{bold} and \textit{italics}, respectively.}
\label{table:downstream_bon}
\resizebox{\linewidth}{!}{
\begin{tabular}{ccclcccc}
\toprule
Model & Target & Dataset & Method & HPSv2 & HPSv3 & ImageReward & PickScore \\
\midrule
\multirow{8}{*}{Hunyuan-DiT}
& \multirow{4}{*}{HPSv2} & \multirow{4}{*}{HPSv2}
& Standard & 0.3146 & 11.8764 & 1.0782 & 22.5782 \\
& & & CA-I & \textit{0.3157} & 11.8274 & 1.0817 & 22.5668 \\
& & & CA-E & \textbf{0.3165} & \textbf{11.9431} & \textit{1.0950} & \textit{22.5790} \\
& & & EnCat    & 0.3162 & \textit{11.8807} & \textbf{1.0996} & \textbf{22.5918} \\
\cmidrule(lr){2-8}
& \multirow{4}{*}{PickScore} & \multirow{4}{*}{PickScore}
& Standard & 0.3043 & 7.3080 & 0.8778 & 21.8130 \\
& & & CA-I & 0.3025 & 7.3220 & 0.8819 & \textit{21.8508} \\
& & & CA-E & \textit{0.3069} & \textbf{7.8130} & \textbf{0.9361} & 21.8163 \\
& & & EnCat    & \textbf{0.3086} & \textit{7.5530} & \textit{0.8912} & \textbf{21.8836} \\

\midrule
\multirow{8}{*}{DreamShaper}
& \multirow{4}{*}{HPSv2} & \multirow{4}{*}{HPSv2}
& Standard & 0.3461 & 13.8647 & 1.3152 & 23.7061 \\
& & & CA-I & 0.3456 & 13.8820 & 1.3094 & 23.6733 \\
& & & CA-E & \textbf{0.3467} & \textit{13.9103} & \textbf{1.3191} & \textit{23.7084} \\
& & & EnCat    & \textit{0.3464} & \textbf{13.9292} & \textit{1.3181} & \textbf{23.7201} \\
\cmidrule(lr){2-8}
& \multirow{4}{*}{PickScore} & \multirow{4}{*}{Pick-a-Pic}
& Standard & 0.3251 & 9.2526 & 1.0202 & 22.6041 \\
& & & CA-I & 0.3224 & 9.2587 & \textit{1.0626} & 22.6096 \\
& & & CA-E & \textit{0.3296} & \textbf{9.3564} & 1.0491 & \textbf{22.6238} \\
& & & EnCat    & \textbf{0.3332} & \textit{9.3111} & \textbf{1.0757} & \textit{22.6146} \\

\midrule
\multirow{8}{*}{SDXL}
& \multirow{4}{*}{HPSv2} & \multirow{4}{*}{HPSv2}
& Standard & 0.2869 & \textit{10.5942} & \textit{0.8506} & 22.5161 \\
& & & CA-I & 0.2865 & 10.5408 & \textbf{0.8575} & \textit{22.5166} \\
& & & CA-E & \textit{0.2871} & 10.5515 & 0.8399 & 22.5051 \\
& & & EnCat    & \textbf{0.2888} & \textbf{10.6055} & 0.8447 & \textbf{22.5177} \\
\cmidrule(lr){2-8}
& \multirow{4}{*}{PickScore} & \multirow{4}{*}{PickScore}
& Standard & 0.2840 & 6.6744 & 0.5794 & 21.6467 \\
& & & CA-I & \textit{0.2866} & \textbf{6.9240} & 0.5733 & \textit{21.6788} \\
& & & CA-E & 0.2848 & 6.8311 & \textit{0.5955} & \textbf{21.7104} \\
& & & EnCat    & \textbf{0.2879} & \textit{6.8435} & \textbf{0.6419} & 21.6777 \\

\midrule
\multirow{8}{*}{PixArt-$\Sigma$}
& \multirow{4}{*}{HPSv2} & \multirow{4}{*}{HPSv2}
& Standard & 0.3010 & 12.3826 & 1.0335 & 22.6759 \\
& & & CA-I & 0.2994 & 12.3722 & 1.0327 & \textit{22.6911} \\
& & & CA-E & \textit{0.3041} & \textit{12.4000} & \textbf{1.0363} & \textbf{22.6964} \\
& & & EnCat    & \textbf{0.3049} & \textbf{12.4401} & \textit{1.0355} & 22.6894 \\
\cmidrule(lr){2-8}
& \multirow{4}{*}{PickScore} & \multirow{4}{*}{PickScore}
& Standard & 0.2976 & \textit{8.1384} & 0.8013 & \textbf{21.7987} \\
& & & CA-I & 0.2973 & 7.9863 & \textbf{0.8614} & 21.7339 \\
& & & CA-E & \textbf{0.2981} & 8.1271 & 0.7880 & 21.7875 \\
& & & EnCat    & \textit{0.2980} & \textbf{8.2353} & \textit{0.8557} & \textit{21.7971} \\

\bottomrule
\end{tabular}
}
\vspace{-6mm}
\end{center}
\end{table}

We further evaluate whether 
offline prediction translates into improved image generation on other HPMs. %
Specifically, given our LambdaLoss+MAE predictors trained on either HPSv2 or PickScore, we further consider whether the improvements on the source metric utilized to train the predictor translate to other metrics in a zero-shot sense. In other words, we would ask `If we train a predictor on PickScore, does that also translate into $S_{p, I}$ gains for HPSv2, HPSv3 or ImageReward?'

Table~\ref{table:downstream_bon} reports our findings, which are largely positive. 
The results show that predictor-guided selection can improve generated images, but the gains depend on both the predictor family and the reward model used for evaluation. Across the aligned settings, EnCat and CA-E are the most consistently competitive designs that most consistently outperform the standard baseline across numerous combinations of DMs, predictor targets and preference metrics. EnCat often improves PickScore oriented evaluations, while CA-E frequently improves HPSv2 or HPSv3 evaluations. In contrast, CA-I is less consistent, falls behind CA-E and EnCat and often struggles to even outperform the baseline. %

Across the HPMs, %
we find that HPSv2, HPSv3 and PickScore largely agree with each other. Typically, if one predictor design is best on one of these metrics, it is also the best, or second best, on the others. By contrast, ImageReward stands as the most disagreeable HPM, %
as it is not uncommon for one predictor like CA-I or the baseline to perform well on it, yet stumble on the other metrics. We further analyze this phenomenon in Section~\ref{sec:metric_results}. 

\begin{figure}[b!]
    \centering
    \setlength{\tabcolsep}{2pt}
    \renewcommand{\arraystretch}{1.15}
    \small

    \newcommand{\imgw}{0.24\textwidth}
    
    \begin{tabular}{@{}c c c c@{}}
        Standard & CA-I & CA-E & EnCat \\

        \includegraphics[width=\imgw]{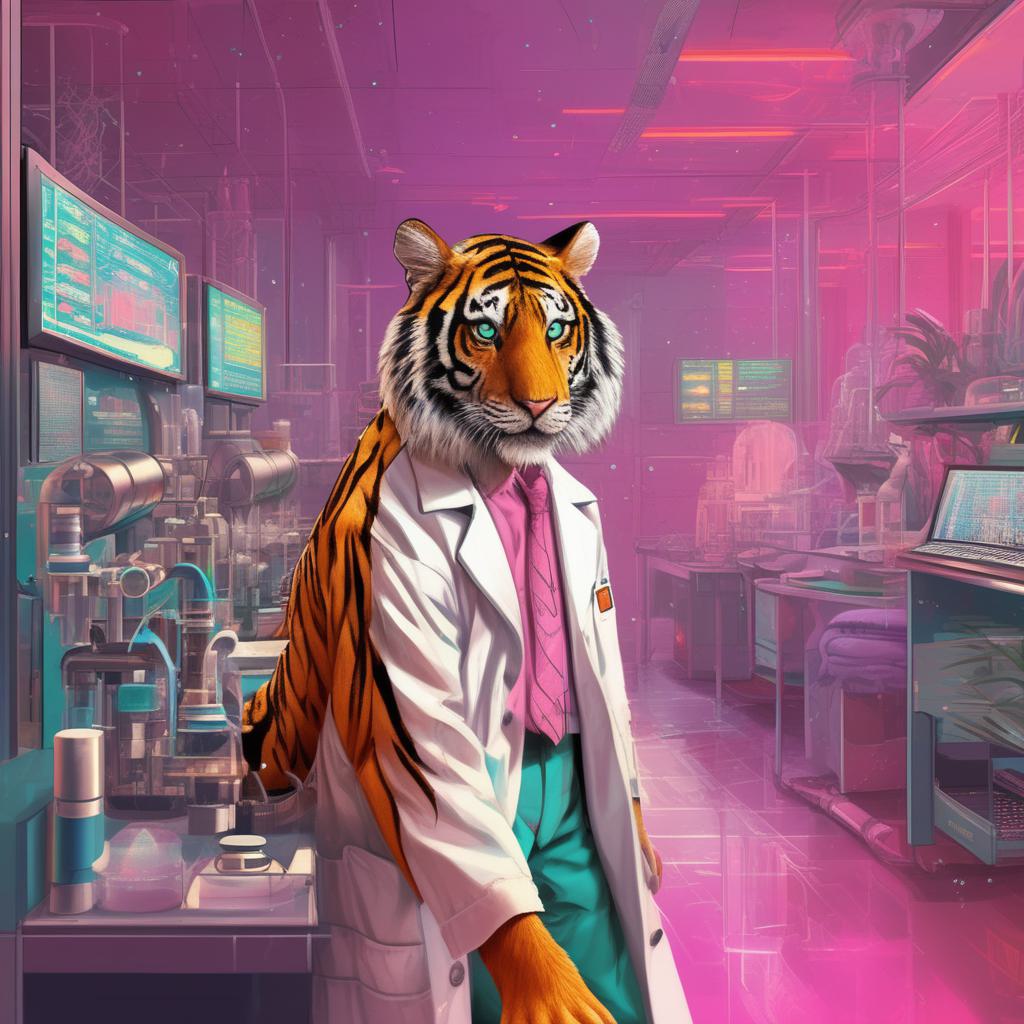} &
        \includegraphics[width=\imgw]{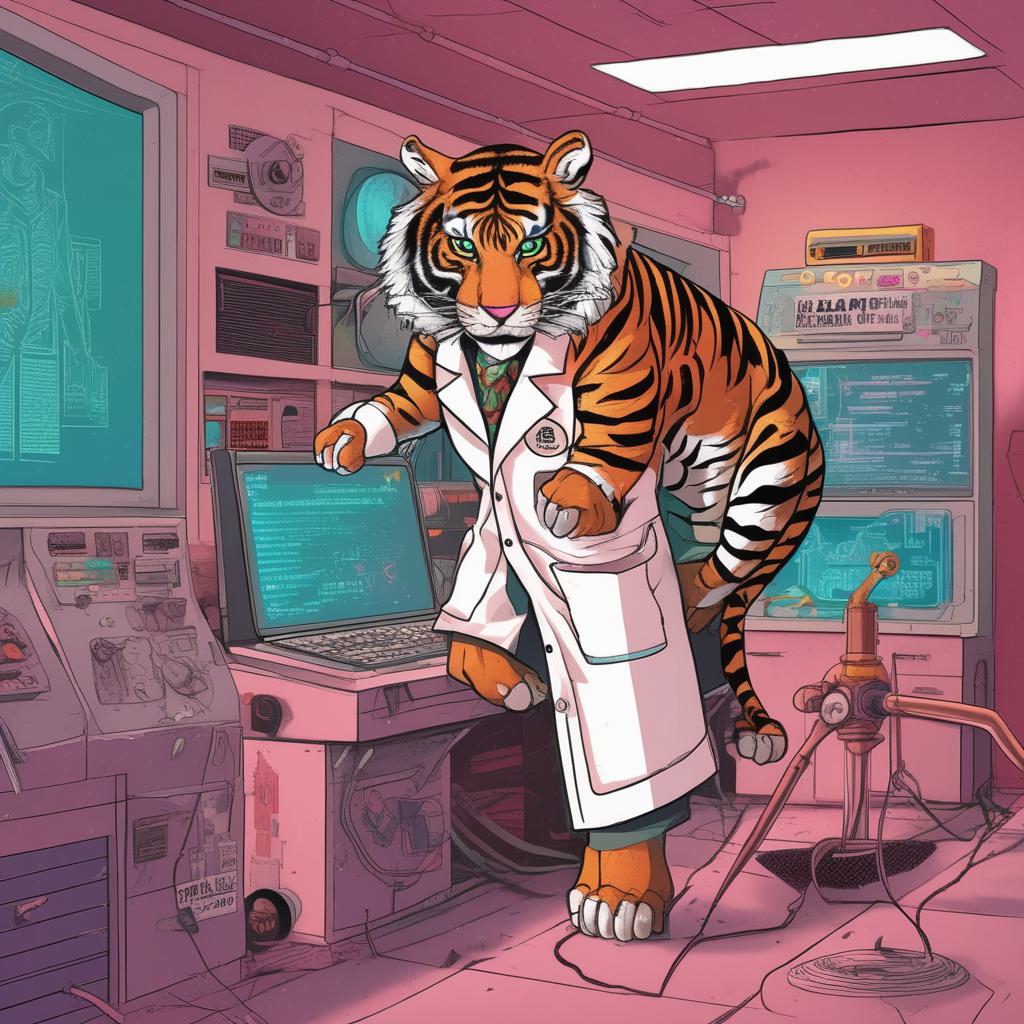} &
        \includegraphics[width=\imgw]{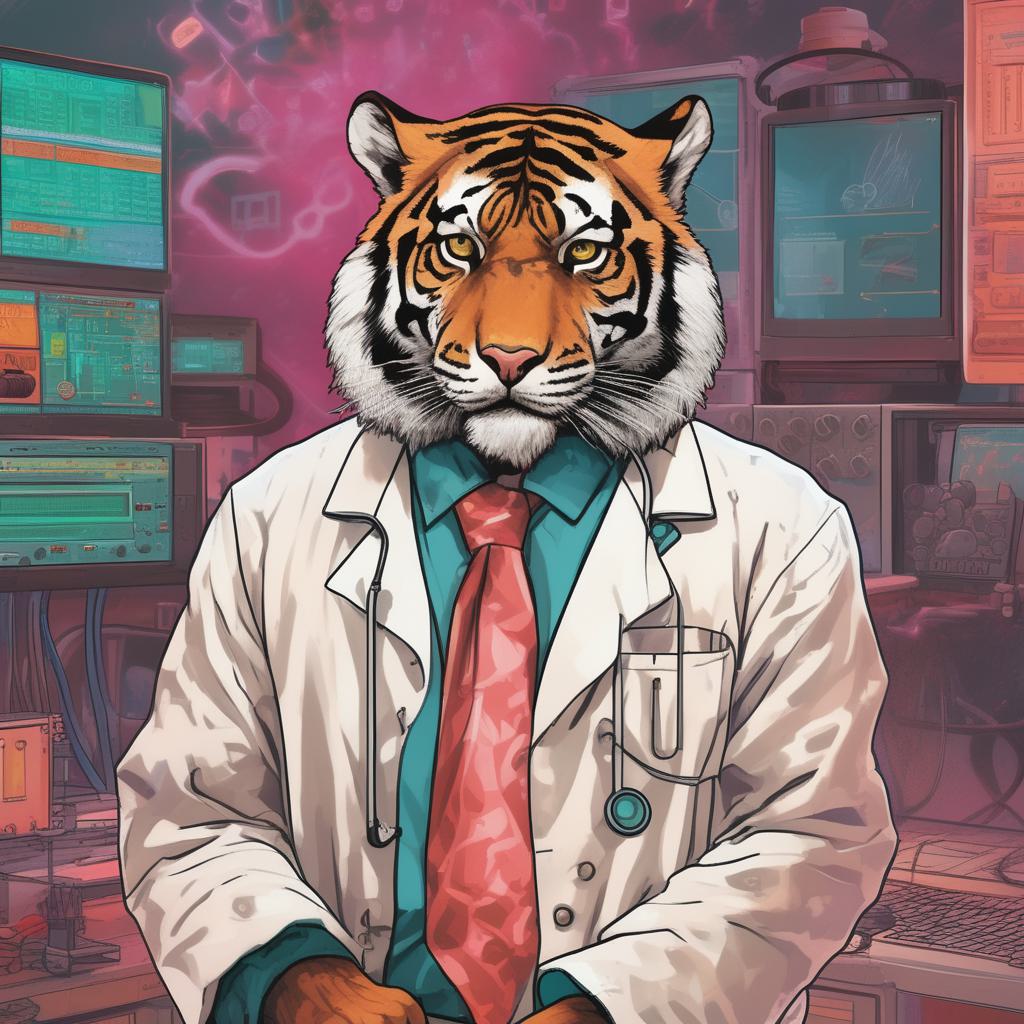} &
        \includegraphics[width=\imgw]{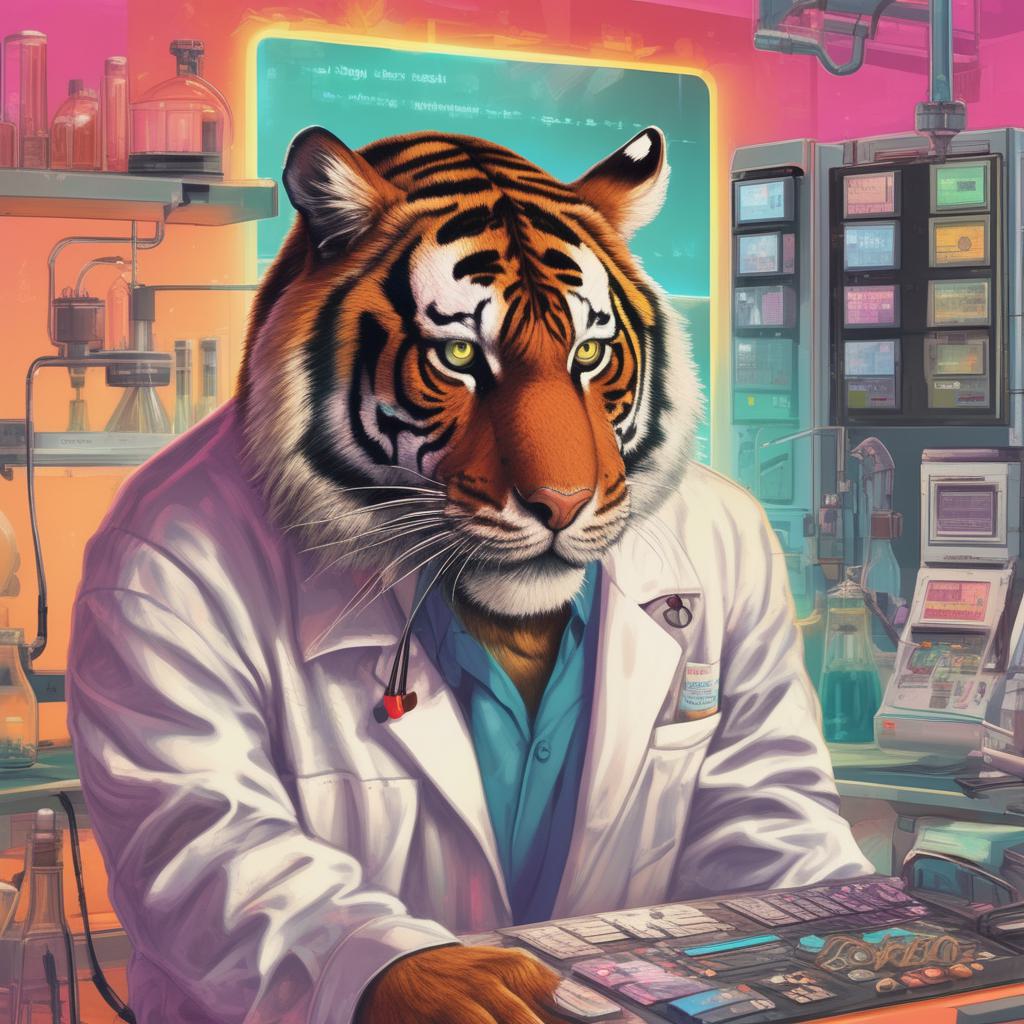} \\
        \multicolumn{4}{@{}p{0.99\textwidth}@{}}{
            SDXL: ``A tiger in a lab coat with a 1980s Miami vibe, turning a well oiled science content machine, digital art.''
        } \\[2mm]

        \includegraphics[width=\imgw]{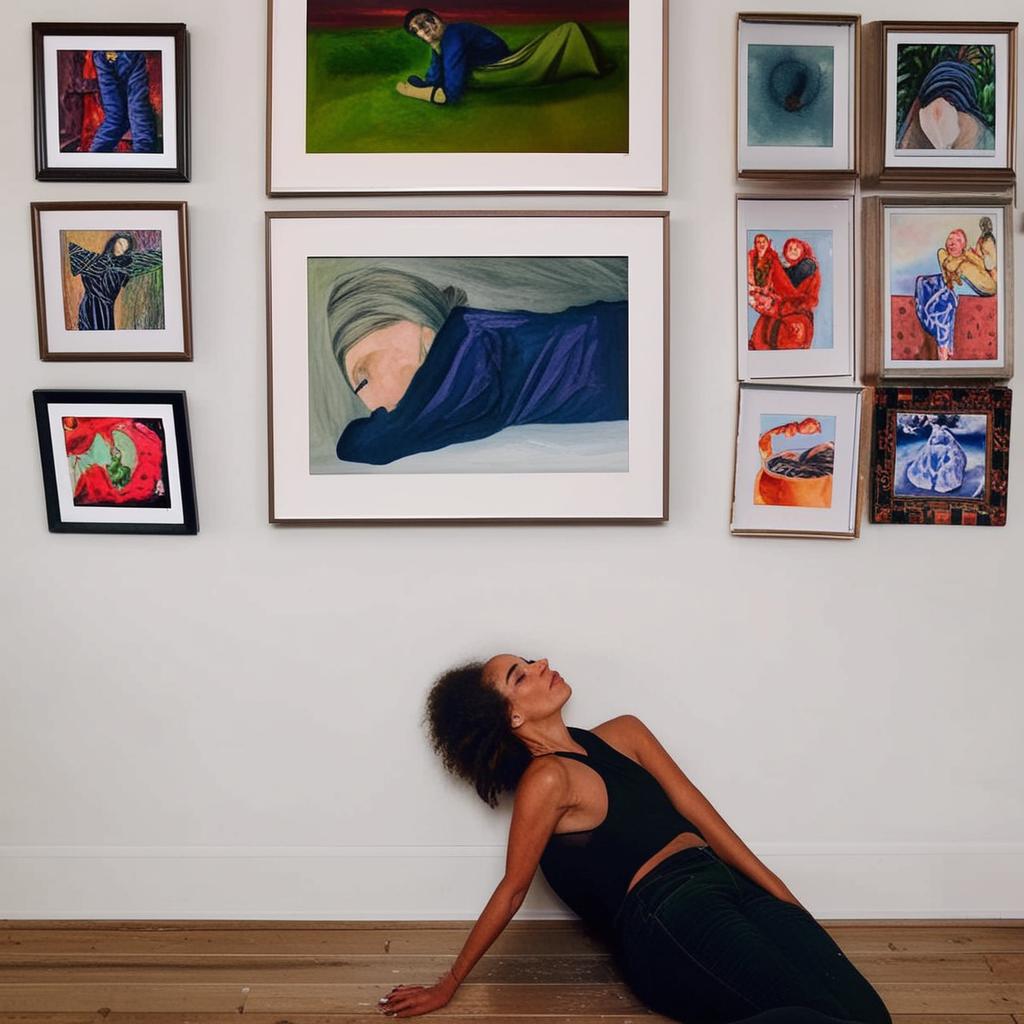} &
        \includegraphics[width=\imgw]{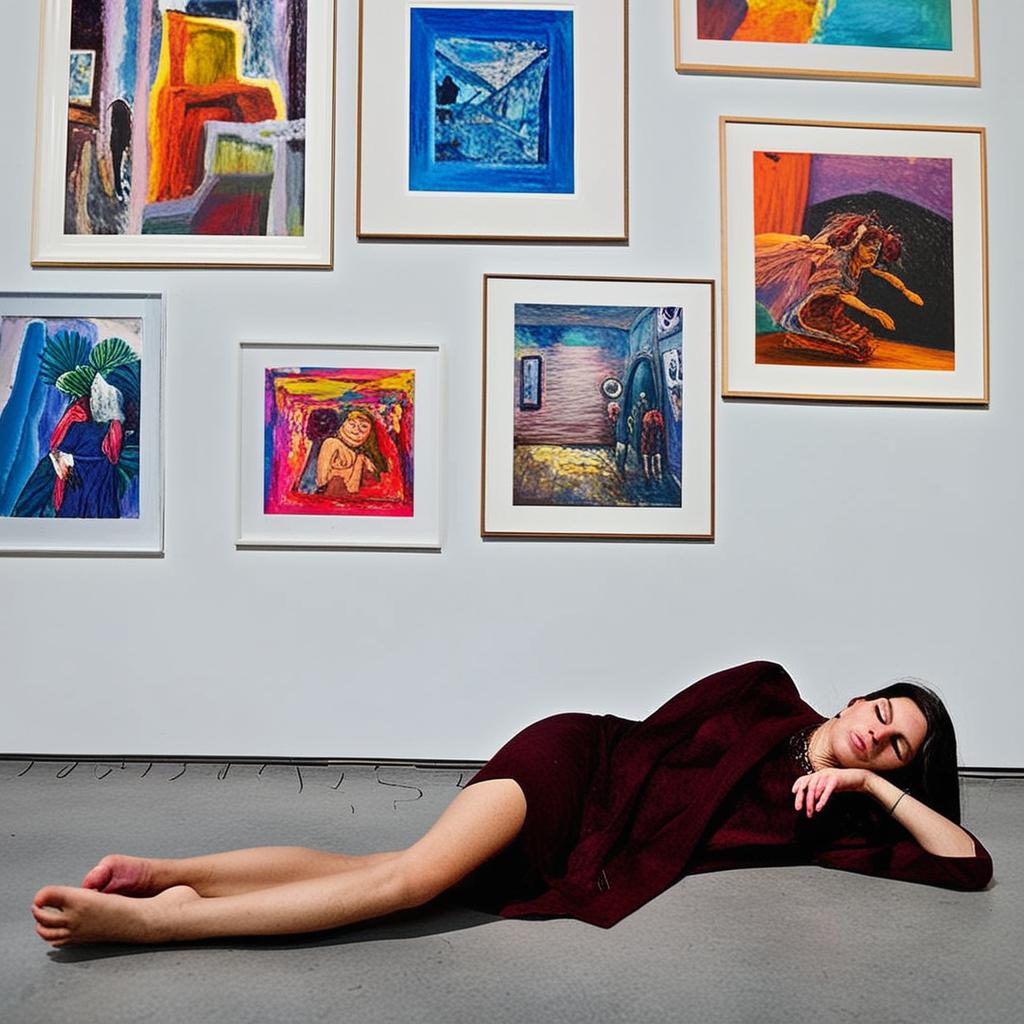} &
        \includegraphics[width=\imgw]{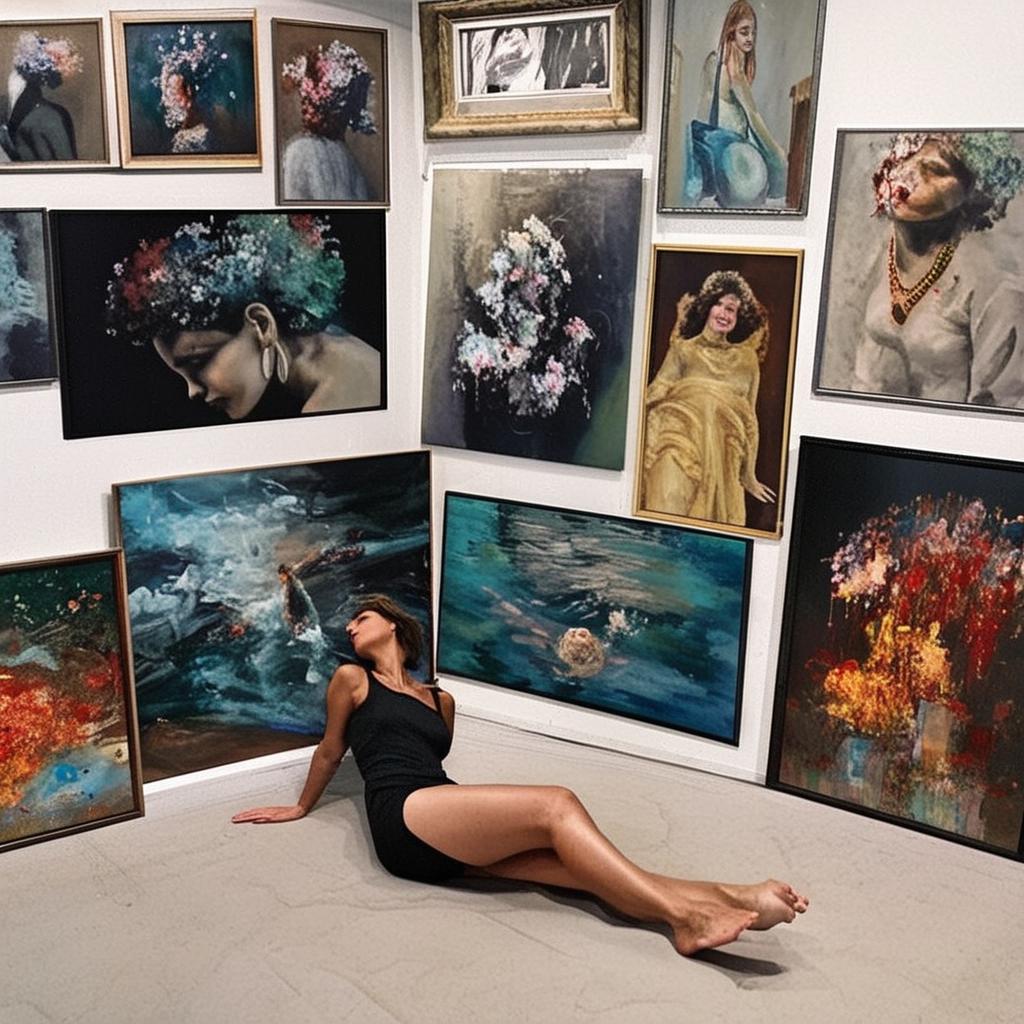} &
        \includegraphics[width=\imgw]{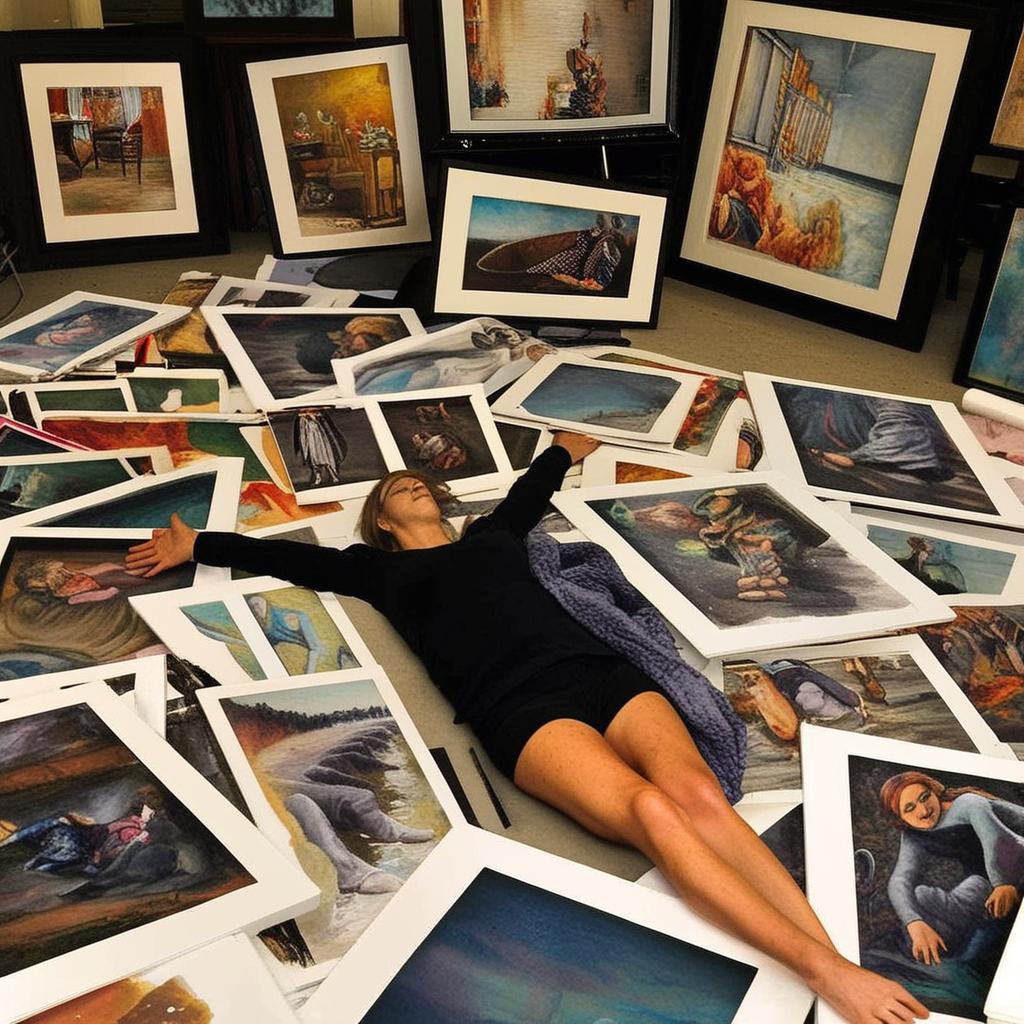} \\
        \multicolumn{4}{@{}p{0.99\textwidth}@{}}{
            Hunyuan-DiT: ``A woman lies down surrounded by piles of paintings by Greg Rutkowski.''
        } \\[2mm]

        \includegraphics[width=\imgw]{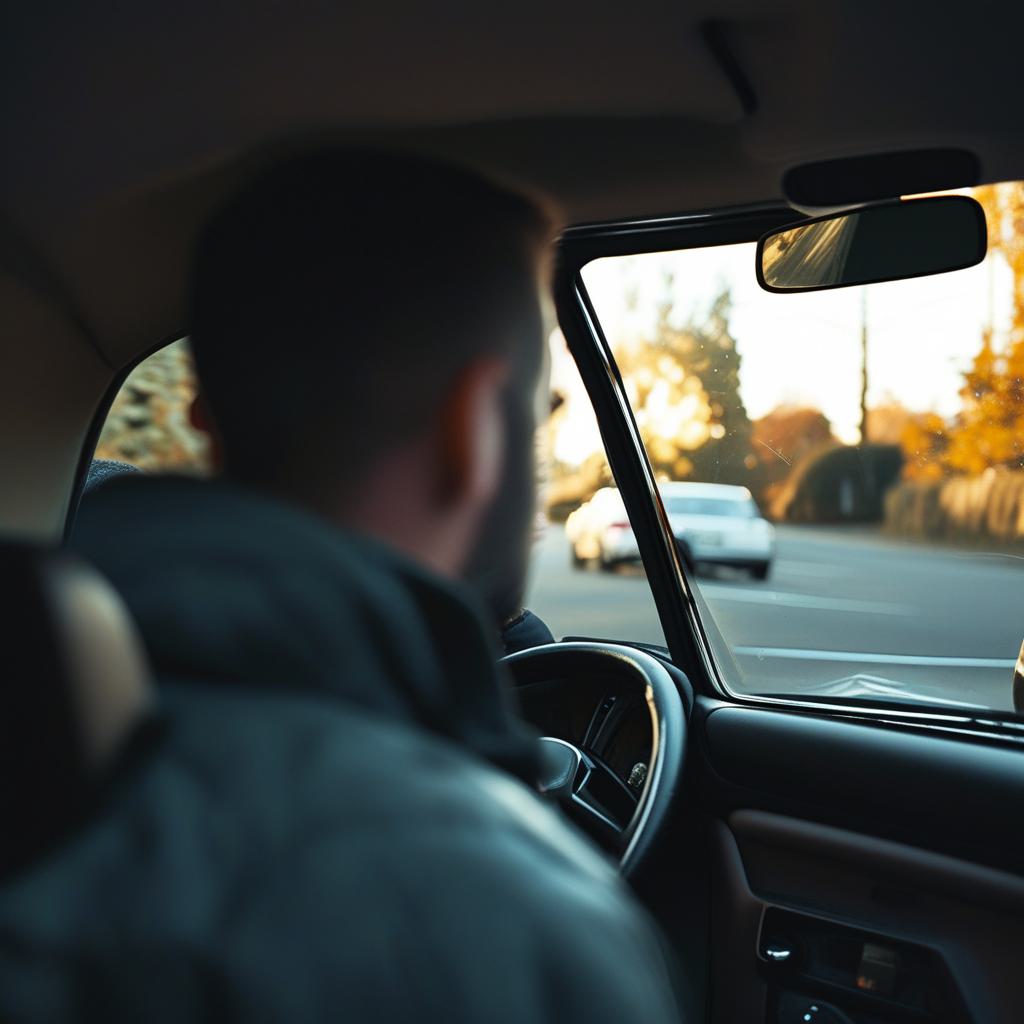} &
        \includegraphics[width=\imgw]{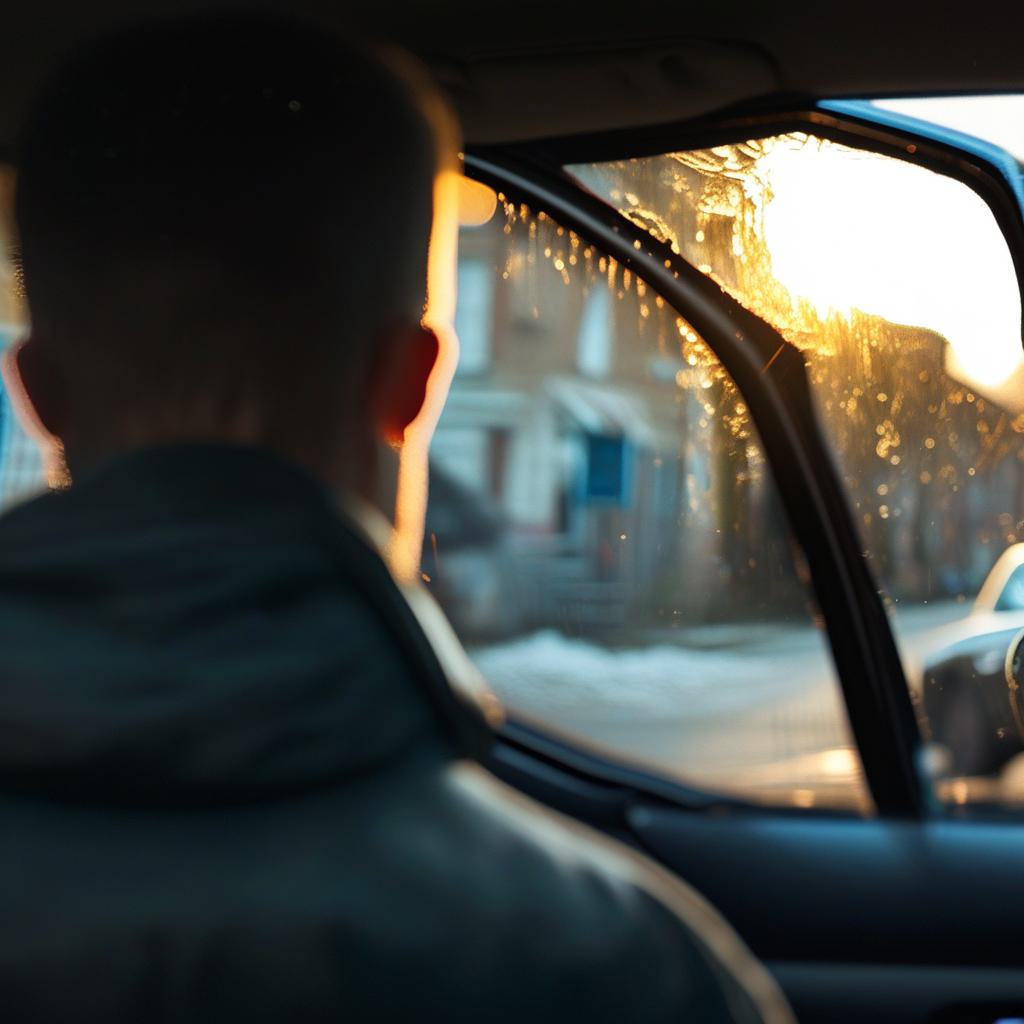} &
        \includegraphics[width=\imgw]{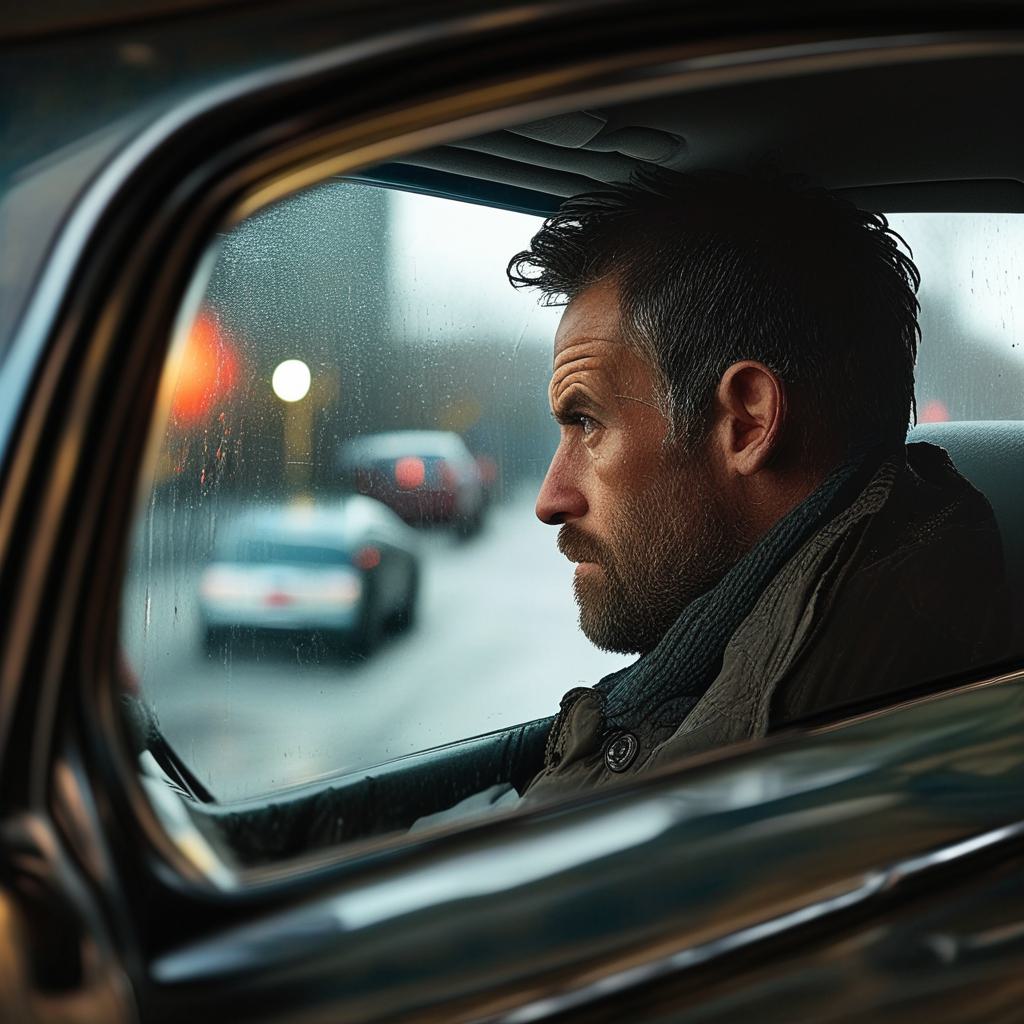} &
        \includegraphics[width=\imgw]{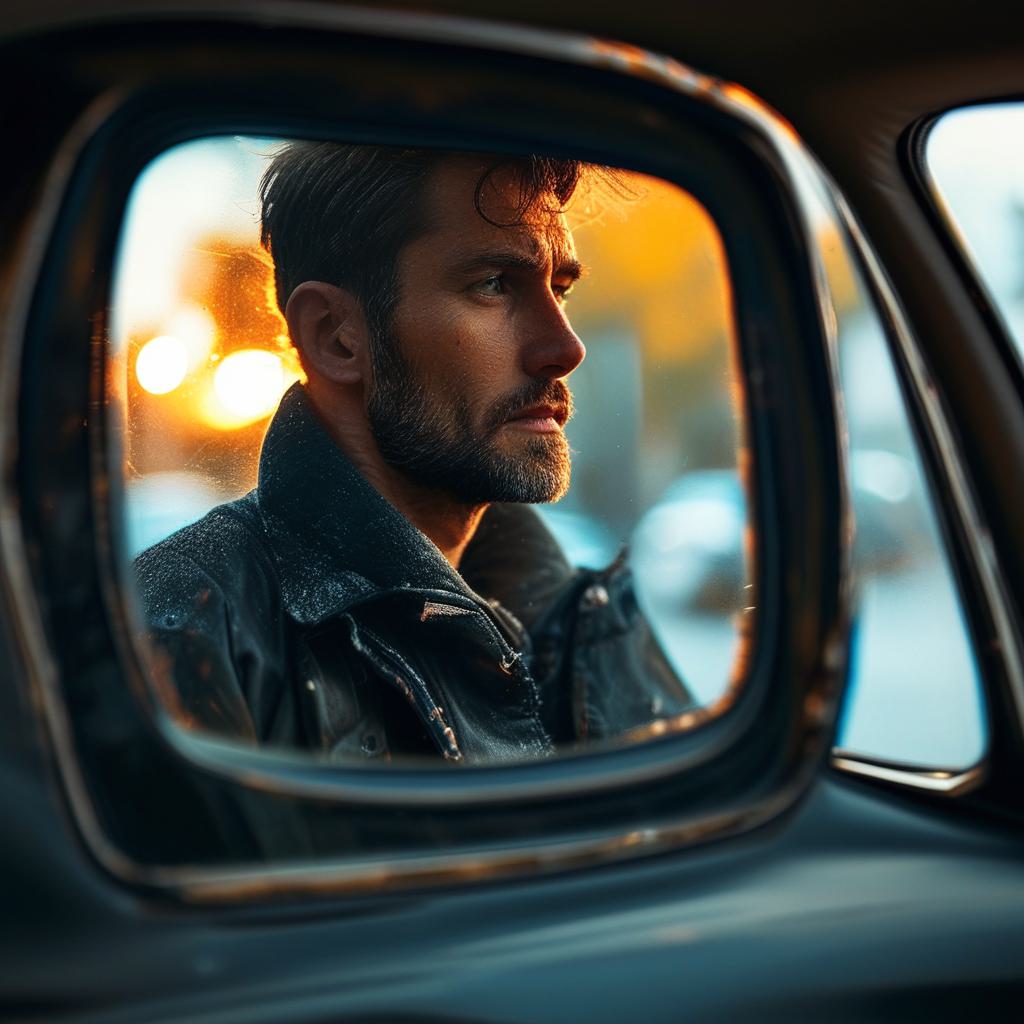} \\
        \multicolumn{4}{@{}p{0.99\textwidth}@{}}{
            PixArt-$\Sigma$: ``A mans reflection in a side view mirror.''
        } \\

    \end{tabular}

    \caption{Qualitative visual examples. DM and prompt details provided. We provide further examples in the supplementary materials.}
    \label{fig:mainbody_examples}
    \vspace{-3mm}
\end{figure}

Next, Figure~\ref{fig:mainbody_examples} shows qualitative examples on several DMs. %
Each row fixes a DM, prompt and then shows the image produced by each method: the `Standard' baseline and then our predictor architectures. %
For the SDXL prompt, EnCat and CA-E render a clear tiger in a lab coat with a detailed laboratory, while CA-I produces an awkward full body pose and the standard image is less detailed. For the Hunyuan-DiT prompt, EnCat shows the woman lying down amid piles of paintings on the floor, while the other selection seat the subject upright or keep the paintings hung on the wall. For the PixArt-$\Sigma$ prompt, EnCat frames the man's reflected face inside the side mirror and CA-E shows his face through the window, while the standard and CA-I images show him from behind with no visible reflection. Overall, these examples align with Table~\ref{table:downstream_bon}: The CA-E and EnCat predictor architectures are better at downstream BoN noise optimization. %

\subsection{RQ3: Which Preference Metrics are Suitable for Optimization?}
\label{sec:metric_results}

We now perform a lateral comparison of the different HPMs %
considered in this paper. The goal is to quantify to what extent performing initial noise optimization using our predictor $\Phi$ on one preference metric will translate to other preference metrics.

To do this, we consider the images generated by our SDXL predictors trained on HPSv2 using LambdaLoss+MAE for the 100 images in the Pick-a-Pick validation set. %
For each generated image, we compare the score of the predictor-selected output against the standard diffusion baseline under HPSv2, HPSv3, ImageReward, and PickScore. We then count how often a method improves over the standard baseline under individual metrics, pairs of metrics, triples of metrics, and all four metrics. This agreement-based view is useful because each reward model captures a different approximation of human preference, and a method that improves only one metric may not correspond to a broadly preferred output.

%\begin{table}[!t]
\begin{wraptable}{rt}{0.6\textwidth}
\begin{center}
\vspace{-5mm}
\caption{Metric-agreement counts for SDXL HPSv2-trained predictors on 100 Pick-a-Pic validation set prompts. Each row reports the number of prompts, out of 100, for which the predictor-selected image improves over the Standard baseline under every metric in the listed set. Single metrics, pairs, triples, and all-four agreement are separated by horizontal rules.}
\label{table:metric_agreement}
\scalebox{\scaleboxratio}{
\begin{tabular}{lccc}
\toprule
Human Preference Metrics & CAM-I & CAM-E & EnCat \\
\midrule
HPSv2 & 52 & 55 & 58 \\
HPSv3 & 54 & 57 & 61 \\
PS & 50 & 53 & 56 \\
IR & 39 & 41 & 43 \\
\midrule
HPSv2 + HPSv3 & 37 & 40 & 43 \\
HPSv2 + PS & 35 & 38 & 41 \\
HPSv3 + PS & 36 & 39 & 42 \\
HPSv2 + IR & 26 & 28 & 30 \\
HPSv3 + IR & 25 & 27 & 29 \\
PS + IR & 23 & 25 & 27 \\
\midrule
HPSv2 + HPSv3 + PS & 29 & 32 & 34 \\
HPSv2 + HPSv3 + IR & 18 & 20 & 22 \\
HPSv2 + PS + IR & 17 & 19 & 21 \\
HPSv3 + PS + IR & 16 & 18 & 20 \\
\midrule
All & 12 & 14 & 16 \\
\bottomrule
\end{tabular}
}
\end{center}
%\end{table}
\vspace{-5mm}
\end{wraptable}

Table~\ref{table:metric_agreement} tallies the results.  %
The single-metric rows show that all three predictors can improve over the standard baseline under HPSv2, HPSv3, and PickScore for at least half of the prompts. %
However, that is not the case when considering ImageReward (IR), which deems the standard baseline image to be superior more than half the time for each predictor.

\begin{figure}[t!]
    \centering
    \setlength{\tabcolsep}{2pt}
    \renewcommand{\arraystretch}{1.15}
    \small

    \begin{tabular}{c c c c}
        Standard & CA-I & CA-E & EnCat \\

        \includegraphics[width=0.24\textwidth]{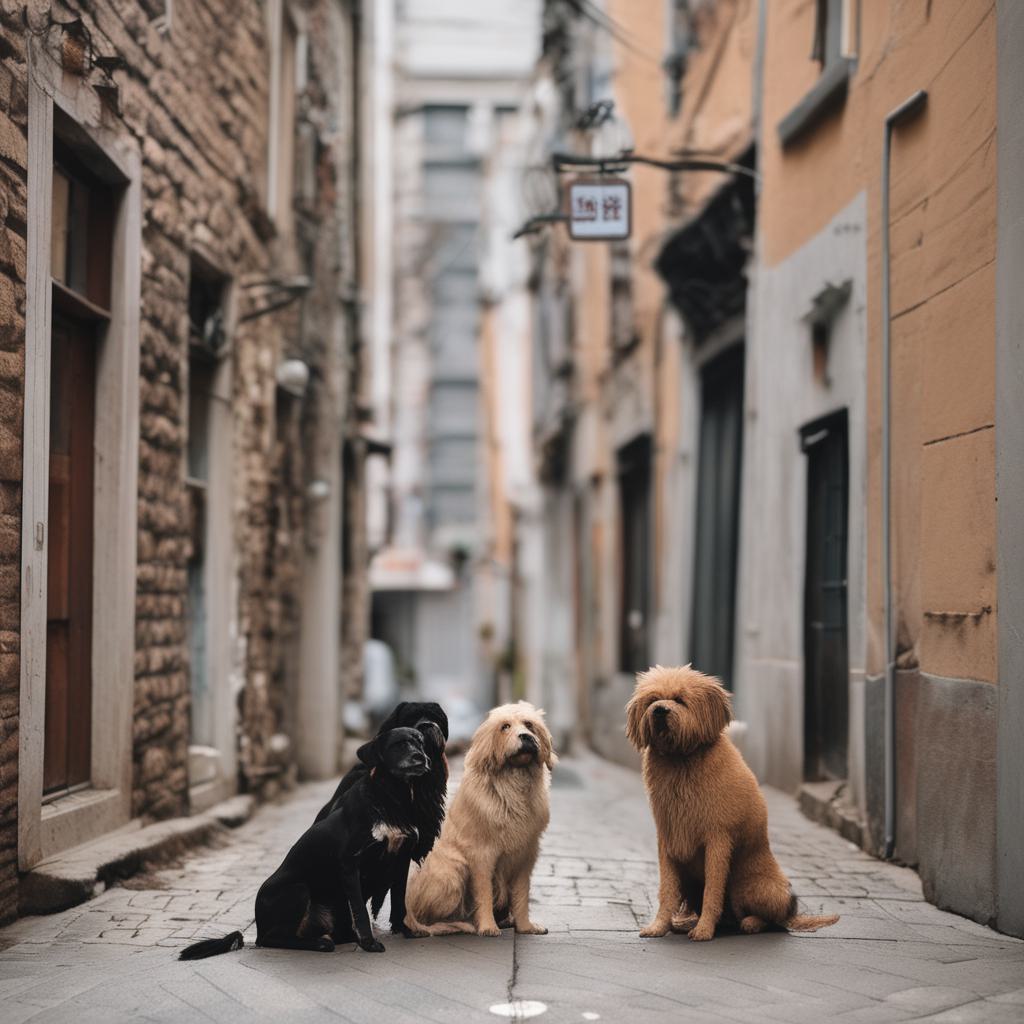} &
        \includegraphics[width=0.24\textwidth]{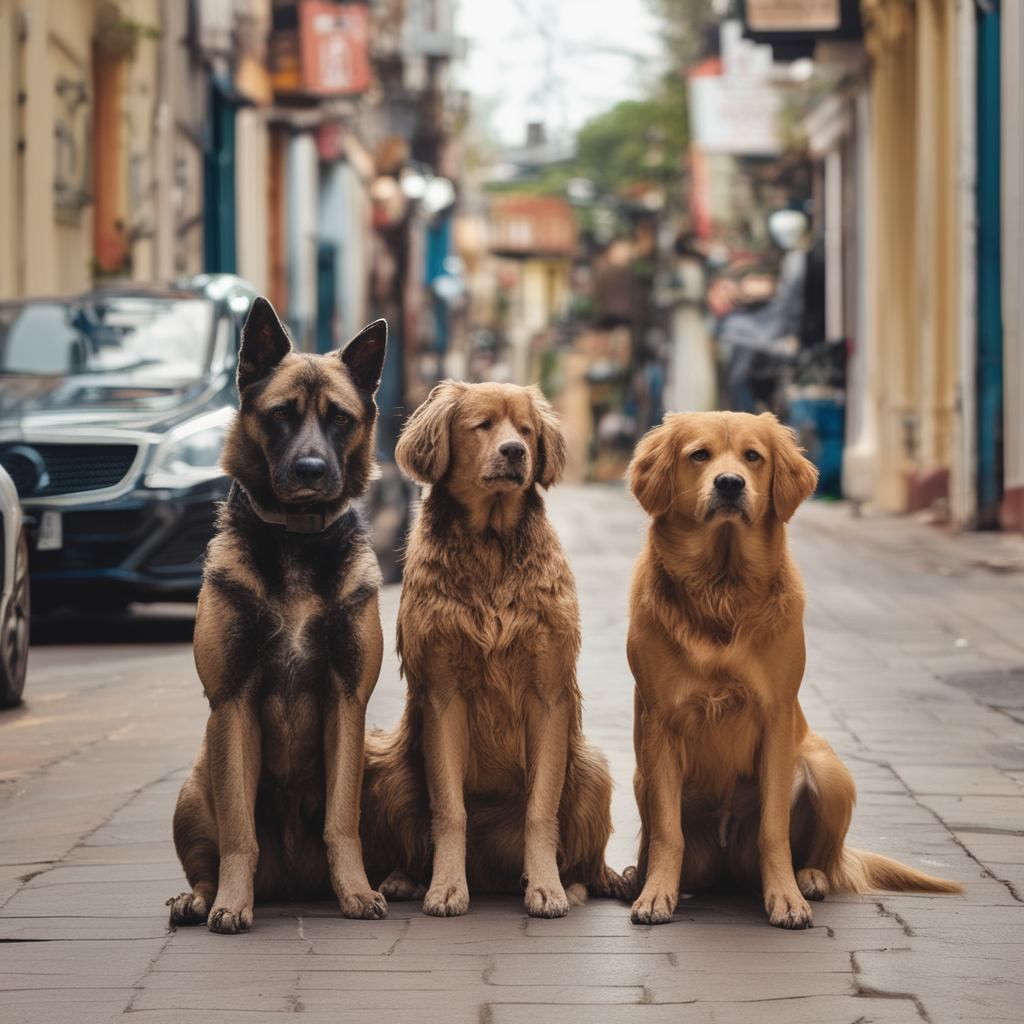} &
        \includegraphics[width=0.24\textwidth]{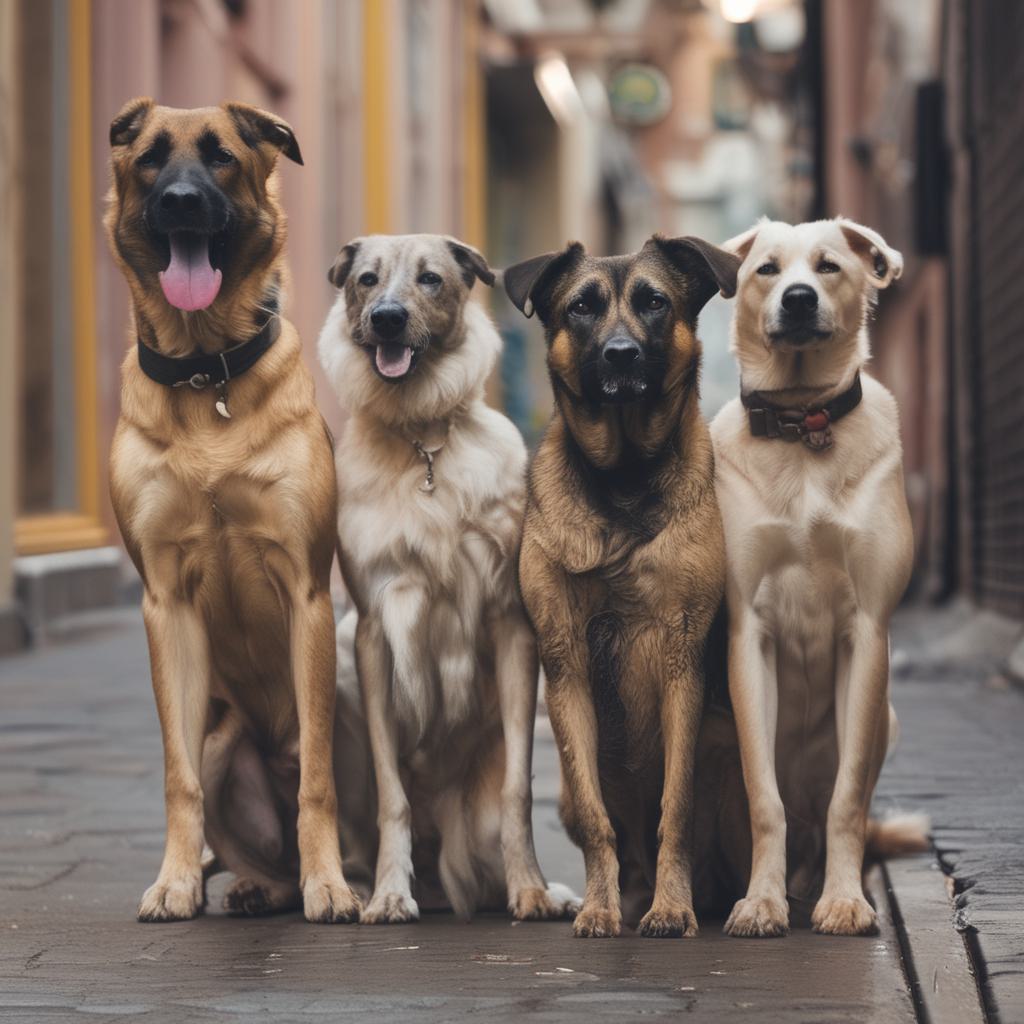} &
        \includegraphics[width=0.24\textwidth]{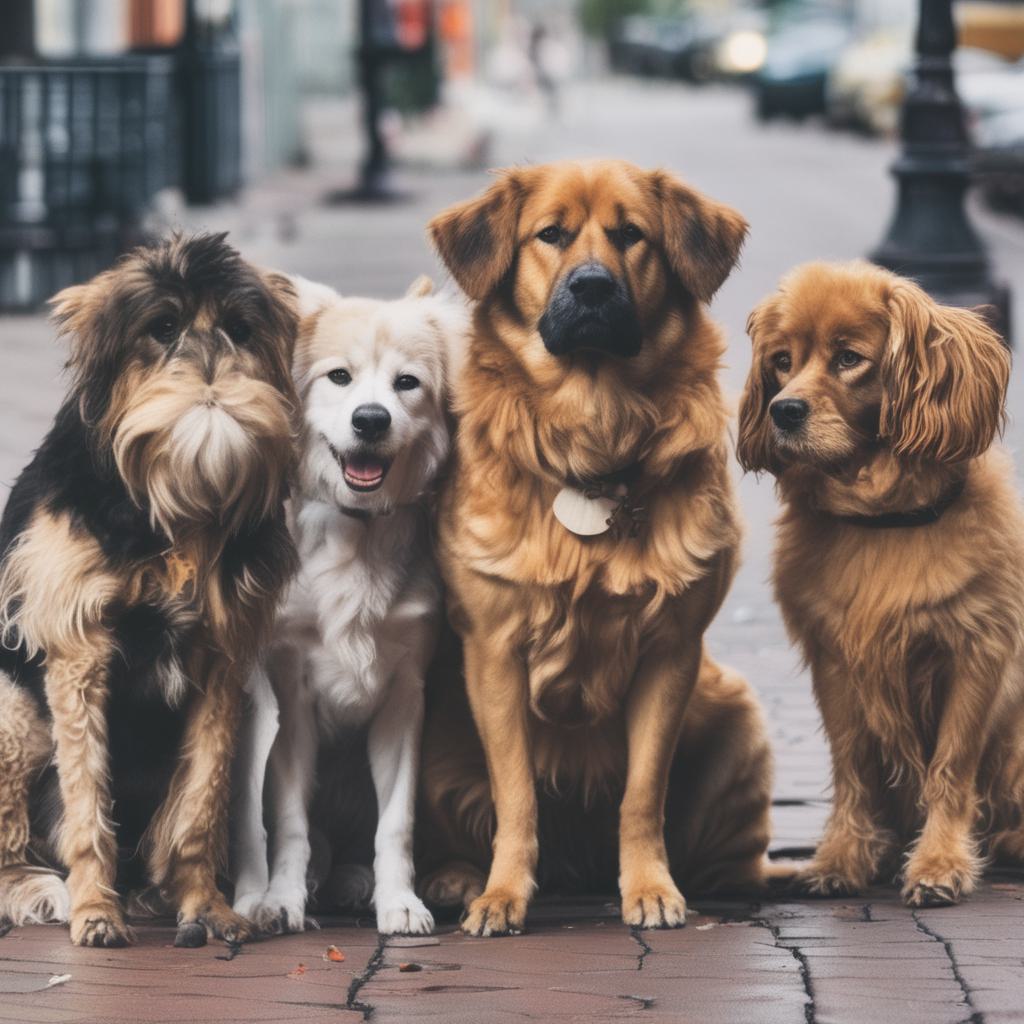} \\

        \footnotesize\begin{tabular}{@{}l@{~~}r@{}}
            HPSv2 & 0.2581 \\ HPSv3 & 5.6969 \\ IR & 1.0351 \\ PS & 22.6994
        \end{tabular} &
        \footnotesize\begin{tabular}{@{}l@{~~}r@{}}
            HPSv2 & 0.2981 \\ HPSv3 & 9.1720 \\ IR & \textbf{1.6239} \\ PS & \textbf{24.9392}
        \end{tabular} &
        \footnotesize\begin{tabular}{@{}l@{~~}r@{}}
            HPSv2 & \textbf{0.3398} \\ HPSv3 & 10.9810 \\ IR & 1.4739 \\ PS & 24.5920
        \end{tabular} &
        \footnotesize\begin{tabular}{@{}l@{~~}r@{}}
            HPSv2 & 0.3379 \\ HPSv3 & \textbf{12.9293} \\ IR & 0.6849 \\ PS & 24.4301
        \end{tabular} \\

    \end{tabular}

    \caption{Qualitative example for the preference-metric analysis on the prompt ``Four dogs on the street.'' \textbf{Bold} marks the best method per metric.}
    \label{fig:pref_metrics_examples}
    \vspace{-5mm}
\end{figure}

These results further cascade when we consider groups of two-or-more HPMs %
at once and establish a clear pattern: Whenever ImageReward is factored in, judgment is skewed towards the standard baseline. This is not the case for when we pair either HPSv2 or v3 with PS, or consider all three of them together. This is further borne out %
in Figure~\ref{fig:pref_metrics_examples} for the prompt ``Four dogs on the street'': EnCat and CA-E render the requested four dogs and top HPSv2 and HPSv3, yet ImageReward ranks EnCat the lowest of all methods, below the Standard baseline. PickScore stands as a bit of an outlier, ranking CA-I the highest even though it only renders three dogs, but we nevertheless note that the PS score values for CA-E and EnCat are much closer to CA-I than they are to the Standard baseline.

Further, among the predictor families, EnCat and CA-E generally preserve higher agreement counts than CA-I. This is consistent with our prior results: %
CA-I can occasionally improve a metric, but it is less reliable when agreement across multiple preference models is required. Overall, these results suggest that HPSv2 and PickScore are more suitable training targets for this task than ImageReward, since improvements under them more often transfer to other preference metrics. Finally, the corresponding PickScore-trained agreement table is provided in the supplementary material.

\subsection{Computational Cost Across Hardware}
\label{sec:cost_results}

%\begin{table}[!t]
\begin{wraptable}{lt}{0.8\linewidth}
\begin{center}
\vspace{-7mm}
\caption{Params and FLOPs comparison for predictor-guided Best-of-N %
with $N=1$. We separate predictor %
FLOPs %
from the FLOPs required to perform feature extraction of the %
cross-attention map for CA-I (FE FLOPs). `D.S.' shorthand for DreamShaper.} %
\label{table:flops}
\scalebox{\scaleboxratio}{
\begin{tabular}{clcccc}
\toprule
Diffusion Model & Method & Params & Predictor FLOPs & FE FLOPs & Total FLOPs \\
\midrule
\multirow{3}{*}{SDXL/D.S.}
& CA-I & 1.33M & 2.70M & 13.5T & 13.5T \\
& CA-E & 1.11M & 489M & 0 & 489M \\
& EnCat & 3.50M & 432M & 0 & 432M \\
\midrule
\multirow{3}{*}{Hunyuan-DiT}
& CA-I & 8.33M & 16.7M & 30.1T & 30.1T \\
& CA-E & 1.64M & 1.09G & 0 & 1.09G \\
& EnCat & 3.63M & 1.86G & 0 & 1.86G \\
\midrule
\multirow{3}{*}{PixArt-$\Sigma$}
& CA-I & 7.52M & 15.0M & 13.3T & 13.3T \\
& CA-E & 2.09M & 1.33G & 0 & 1.33G \\
& EnCat & 12.00M & 6.71G & 0 & 6.71G \\
\bottomrule
\end{tabular}%
}
\vspace{-5mm}
\end{center}
%\end{table}
\end{wraptable}
Next, we measure the computational overhead introduced by guided noise selection when integrated into a DM pipeline. %
Table~\ref{table:flops} provides a view of predictor FLOPs\\where we separate % a complementary view by separating 
FLOPs required by the HPM predictor itself, from %predictor FLOPs from 
feature-extraction FLOPs. Here, the true weakness of CA-I becomes clearer: The FLOPs cost required to perform partial inference on the DM denoiser to extract the cross-attention map raises the overall cost of evaluating a \textit{single} noise $X_T$ by several magnitudes - from gigaFLOPs to teraFLOPs - above what is required by CA-E and EnCat. 

Further, Table~\ref{table:runtime} provides a complementary view of %reports 
the wall-clock time consumed to perform BoN selection with $N=100$ across a trio of hardware devices. As we can see, both CA-E and EnCat add negligible overhead across all DMs and hardware. The issue is CA-I~\cite{guo2024initno}, which is substantially more expensive as it must partially invoke the DM denoiser to extract the first cross-attention map at $t=T$. This cost is especially true for the newer DiTs Hunyuan and PixArt-$\Sigma$, where selection time for CA-I can exceed the actual generation time multiple times over, but is also of significant concern for SDXL-based U-Nets.

\begin{table}[!t]
\begin{center}
\caption{Runtime comparison for predictor-guided noise optimization across varying Nvidia hardware. We consider BoN selection time for $N=100$ as well as total generation time for the DM to generate an image. All results in seconds. %
Recall that CA-I utilizes the internal cross-attention map from the first DM transformer block when $t=T$. 
}
\label{table:runtime}
\scalebox{\scaleboxratio}{
\begin{tabular}{cclcc}
\toprule
GPU Device & Diffusion Model & Method & Selection Time & +Generation \\
\midrule
\multirow{12}{*}{A100}
& \multirow{4}{*}{SDXL/DreamShaper}
& Standard & 0 & 6.3651 \\
& & CA-I & 13.8529 & 20.0833 \\
& & CA-E & 0.0763 & 6.3680 \\
& & EnCat & 0.0669 & 6.3648 \\
\cmidrule(lr){2-5}
& \multirow{4}{*}{Hunyuan-DiT}
& Standard & 0 & 17.863 \\
& & CA-I & 43.5509 & 61.331 \\
& & CA-E & 0.123 & 17.9027 \\
& & EnCat & 0.1069 & 17.8880 \\
\cmidrule(lr){2-5}
& \multirow{4}{*}{PixArt-$\Sigma$}
& Standard & 0 & 17.1243 \\
& & CA-I & 65.4173 & 77.3544 \\
& & CA-E & 4.3579 & 16.6388 \\
& & EnCat & 4.8550 & 16.8888 \\
\midrule
\multirow{12}{*}{RTX PRO 5000}
& \multirow{4}{*}{SDXL/DreamShaper}
& Standard & 0 & 5.4108 \\
& & CA-I & 10.3342 & 15.5247 \\
& & CA-E & 0.0201 & 5.1992 \\
& & EnCat & 0.044 & 5.1811 \\
\cmidrule(lr){2-5}
& \multirow{4}{*}{Hunyuan-DiT}
& Standard & 0 & 15.3909 \\
& & CA-I & 34.7874 & 50.4423 \\
& & CA-E & 0.0493 & 15.7822 \\
& & EnCat & 0.0615 & 15.2791 \\
\cmidrule(lr){2-5}
& \multirow{4}{*}{PixArt-$\Sigma$}
& Standard & 0 & 9.7646 \\
& & CA-I & 34.533 & 41.0288 \\
& & CA-E & 2.483 & 8.9693 \\
& & EnCat & 2.5451 & 9.1638 \\
\midrule
\multirow{12}{*}{DGX Spark}
& \multirow{4}{*}{SDXL/DreamShaper}
& Standard & 0 & 20.4274 \\
& & CA-I & 40.9684 & 60.2145 \\
& & CA-E & 0.0904 & 19.5241 \\
& & EnCat & 0.083 & 19.0025 \\
\cmidrule(lr){2-5}
& \multirow{4}{*}{Hunyuan-DiT}
& Standard & 0 & 60.4956 \\
& & CA-I & 137.3244 & 196.7365 \\
& & CA-E & 0.2662 & 59.7327 \\
& & EnCat & 0.1966 & 59.4129 \\
\cmidrule(lr){2-5}
& \multirow{4}{*}{PixArt-$\Sigma$}
& Standard & 0 & 8.2562 \\
& & CA-I & 38.9267 & 45.3572 \\
& & CA-E & 0.5166 & 6.9824 \\
& & EnCat & 0.803 & 7.2272 \\
\bottomrule
\end{tabular}%
}
\vspace{-5mm}
\end{center}
\end{table}

Thus, given the hardware results presented and prior evaluation results, we observe that CA-I is the least optimal form of predictor for initial noise optimization via human preference estimation. In contrast, both CA-E and EnCat have roughly comparable hardware costs and are around equally capable when it comes to downstream inference in an end-to-end DM pipeline. Notably, these two predictor styles are lightweight and impose limited negligible computational overhead cost. 

\section{Conclusion}
\label{sec:conclusion}

This paper questions whether it is possible to improve T2I DM generative performance by combining human preference metrics, initial noise optimization and Best-of-N (BoN) sorting. We address this with three research questions targeting the feasibility of estimating human preference metrics from DM initial conditions, whether initial optimization via a predictor and BoN can yield downstream improvements on existing DM pipelines, while also performing a lateral comparison of different preference metrics.

To answer these questions, we consider three distinct forms of predictor that models the human preference score from the prompt and initial noise. We find that two of these predictor designs are able to consistently improve end-to-end human preference metric benchmark scores even in a zero-shot setting. Ironically, we also find that the computationally inefficient form of predictor is the least effective at this task.

\bibliography{egbib}

@String(CVPR  = {IEEE Conf. Comput. Vis. Pattern Recog.})

@String(NeurIPS = {Adv. Neural Inform. Process. Syst.})

@String(ICML  = {Int. Conf. Mach. Learn.})

@String(ICLR  = {Int. Conf. Learn. Represent.})

@String(AAAI  = {AAAI})

@String(CVPR  = {CVPR})

@String(NeurIPS = {NeurIPS})

@String(ICML  = {ICML})

@String(ICLR  = {ICLR})

@inproceedings{goodfellow2014generative,
  title={Generative adversarial nets},
  author={Goodfellow, Ian and Pouget-Abadie, Jean and Mirza, Mehdi and Xu, Bing and Warde-Farley, David and Ozair, Sherjil and Courville, Aaron and Bengio, Yoshua},
  booktitle={Advances in neural information processing systems},
  pages={2672--2680},
  year={2014}
}

@inproceedings{vaswani2017attention,
 author = {Vaswani, Ashish and Shazeer, Noam and Parmar, Niki and Uszkoreit, Jakob and Jones, Llion and Gomez, Aidan N and Kaiser, \L ukasz and Polosukhin, Illia},
 booktitle = {Advances in Neural Information Processing Systems},
 editor = {I. Guyon and U. Von Luxburg and S. Bengio and H. Wallach and R. Fergus and S. Vishwanathan and R. Garnett},
 pages = {},
 publisher = {Curran Associates, Inc.},
 title = {Attention is All you Need},
 url = {https://proceedings.neurips.cc/paper_files/paper/2017/file/3f5ee243547dee91fbd053c1c4a845aa-Paper.pdf},
 volume = {30},
 year = {2017}
}

@article{wang2020picking,
  title={Picking winning tickets before training by preserving gradient flow},
  author={Wang, Chaoqi and Zhang, Guodong and Grosse, Roger},
  journal={arXiv preprint arXiv:2002.07376},
  year={2020}
}

@inproceedings{kirillov2019panoptic,
  title={Panoptic Feature Pyramid Networks},
  author={Kirillov, Alexander and Girshick, Ross and He, Kaiming and Doll{\'a}r, Piotr},
  booktitle={Proceedings of the IEEE/CVF Conference on Computer Vision and Pattern Recognition},
  pages={6399--6408},
  year={2019}
}

@article{white2021powerful,
  title={How powerful are performance predictors in neural architecture search?},
  author={White, Colin and Zela, Arber and Ru, Robin and Liu, Yang and Hutter, Frank},
  journal={NeurIPS},
  year={2021}
}

@inproceedings{he2016identity,
  title={Identity Mappings in Deep Residual Networks},
  author={He, Kaiming and Zhang, Xiangyu and Ren, Shaoqing and Sun, Jian},
  booktitle={European Conference on Computer Vision},
  pages={630--645},
  year={2016},
  organization={Springer}
}

@inproceedings{salameh2024autogo,
 shorthand={Salameh and Mills et al., 2023},
 author = {Salameh, Mohammad and Mills, Keith G. and Hassanpour, Negar and Han, Fred and Zhang, Shuting and Lu, Wei and Jui, Shangling and Zhou, Chunhua and Sun, Fengyu and Niu, Di},
 booktitle = {Advances in Neural Information Processing Systems},
 title = {AutoGO: Automated Computation Graph Optimization for Neural Network Evolution},
 year = {2023}
}

@inbook{han2023general,
    author = {Fred X. Han and Keith G. Mills and Fabian Chudak and Parsa Riahi and Mohammad Salameh and Jialin Zhang and Wei Lu and Shangling Jui and Di Niu},
    title = {A General-Purpose Transferable Predictor for Neural Architecture Search},
    booktitle = {Proceedings of the 2023 SIAM International Conference on Data Mining (SDM)},
    publisher = {Society for Industrial and Applied Mathematics},
    chapter = {},
    pages = {721-729},
    doi = {10.1137/1.9781611977653.ch81},
    URL = {https://epubs.siam.org/doi/abs/10.1137/1.9781611977653.ch81},
    year = {2023},
}

@inproceedings{lu2023pinat,
  title     = {PINAT: A Permutation INvariance Augmented Transformer for NAS Predictor},
  author    = {Lu, Shun and Hu, Yu and Wang, Peihao and Han, Yan and Tan, Jianchao and Li, Jixiang and Yang, Sen and Liu, Ji},
  booktitle = {Proceedings of the AAAI Conference on Artificial Intelligence (AAAI)},
  year      = {2023}
}

@InProceedings{Rombach_2022_CVPR,
    author    = {Rombach, Robin and Blattmann, Andreas and Lorenz, Dominik and Esser, Patrick and Ommer, Bj\"orn},
    title     = {High-Resolution Image Synthesis With Latent Diffusion Models},
    booktitle = {CVPR},
    year      = {2022},
}

@inproceedings{blondel2020fast,
  title={Fast differentiable sorting and ranking},
  author={Blondel, Mathieu and Teboul, Olivier and Berthet, Quentin and Djolonga, Josip},
  booktitle={International Conference on Machine Learning},
  pages={950--959},
  year={2020},
  organization={PMLR}
}

@article{heusel2017gans,
  title={Gans trained by a two time-scale update rule converge to a local nash equilibrium},
  author={Heusel, Martin and Ramsauer, Hubert and Unterthiner, Thomas and Nessler, Bernhard and Hochreiter, Sepp},
  journal={Advances in neural information processing systems},
  volume={30},
  year={2017}
}

@inproceedings{loshchilov19AdamW,
  author       = {Ilya Loshchilov and
                  Frank Hutter},
  title        = {Decoupled Weight Decay Regularization},
  booktitle    = {7th International Conference on Learning Representations, {ICLR} 2019,
                  New Orleans, LA, USA, May 6-9, 2019},
  publisher    = {OpenReview.net},
  year         = {2019},
  url          = {https://openreview.net/forum?id=Bkg6RiCqY7},
  bibsource    = {dblp computer science bibliography, https://dblp.org}
}

@inproceedings{peebles2023scalable,
  title={Scalable diffusion models with transformers},
  author={Peebles, William and Xie, Saining},
  booktitle={Proceedings of the IEEE/CVF International Conference on Computer Vision},
  pages={4195--4205},
  year={2023}
}

@inproceedings{podell2023sdxl,
  author       = {Dustin Podell and
                  Zion English and
                  Kyle Lacey and
                  Andreas Blattmann and
                  Tim Dockhorn and
                  Jonas M{\"{u}}ller and
                  Joe Penna and
                  Robin Rombach},
  title        = {{SDXL:} Improving Latent Diffusion Models for High-Resolution Image
                  Synthesis},
  booktitle    = {The Twelfth International Conference on Learning Representations,
                  {ICLR} 2024, Vienna, Austria, May 7-11, 2024},
  publisher    = {OpenReview.net},
  year         = {2024},
  url          = {https://openreview.net/forum?id=di52zR8xgf},
  bibsource    = {dblp computer science bibliography, https://dblp.org}
}

@inproceedings{esser2024scaling,
  author       = {Patrick Esser and
                  Sumith Kulal and
                  Andreas Blattmann and
                  Rahim Entezari and
                  Jonas M{\"{u}}ller and
                  Harry Saini and
                  Yam Levi and
                  Dominik Lorenz and
                  Axel Sauer and
                  Frederic Boesel and
                  Dustin Podell and
                  Tim Dockhorn and
                  Zion English and
                  Robin Rombach},
  title        = {Scaling Rectified Flow Transformers for High-Resolution Image Synthesis},
  booktitle    = {Forty-first International Conference on Machine Learning, {ICML} 2024,
                  Vienna, Austria, July 21-27, 2024},
  publisher    = {OpenReview.net},
  year         = {2024},
  url          = {https://openreview.net/forum?id=FPnUhsQJ5B},
  bibsource    = {dblp computer science bibliography, https://dblp.org}
}

@misc{von-platen-etal-2022-diffusers,
  shorthand={von Platen et al., 2022},
  author = {Patrick von Platen and Suraj Patil and Anton Lozhkov and Pedro Cuenca and Nathan Lambert and Kashif Rasul and Mishig Davaadorj and Dhruv Nair and Sayak Paul and William Berman and Yiyi Xu and Steven Liu and Thomas Wolf},
  title = {Diffusers: State-of-the-art diffusion models},
  year = {2022},
  publisher = {GitHub},
  journal = {GitHub repository},
  howpublished = {\url{https://github.com/huggingface/diffusers}}
}

@misc{bfl2024Flux,
  author = {{Black Forest Labs}},
  title = {FLUX},
  year = {2024},
  publisher = {GitHub},
  journal = {GitHub repository},
  howpublished = {\url{https://github.com/black-forest-labs/flux}}
}

@inproceedings{sohl2015deep,
  title={Deep unsupervised learning using nonequilibrium thermodynamics},
  author={Sohl-Dickstein, Jascha and Weiss, Eric and Maheswaranathan, Niru and Ganguli, Surya},
  booktitle={International conference on machine learning},
  pages={2256--2265},
  year={2015},
  organization={PMLR}
}

@article{ho2020denoising,
  title={Denoising diffusion probabilistic models},
  author={Ho, Jonathan and Jain, Ajay and Abbeel, Pieter},
  journal={Advances in neural information processing systems},
  volume={33},
  pages={6840--6851},
  year={2020}
}

@inproceedings{kingma2013auto,
  author       = {Diederik P. Kingma and
                  Max Welling},
  editor       = {Yoshua Bengio and
                  Yann LeCun},
  title        = {Auto-Encoding Variational Bayes},
  booktitle    = {2nd International Conference on Learning Representations, {ICLR} 2014,
                  Banff, AB, Canada, April 14-16, 2014, Conference Track Proceedings},
  year         = {2014},
  url          = {http://arxiv.org/abs/1312.6114},
  bibsource    = {dblp computer science bibliography, https://dblp.org}
}

@article{ho2022video,
  title={Video diffusion models},
  author={Ho, Jonathan and Salimans, Tim and Gritsenko, Alexey and Chan, William and Norouzi, Mohammad and Fleet, David J},
  journal={Advances in Neural Information Processing Systems},
  volume={35},
  pages={8633--8646},
  year={2022}
}

@inproceedings{chen2024pixartAlpha,
  author       = {Junsong Chen and
                  Jincheng Yu and
                  Chongjian Ge and
                  Lewei Yao and
                  Enze Xie and
                  Zhongdao Wang and
                  James T. Kwok and
                  Ping Luo and
                  Huchuan Lu and
                  Zhenguo Li},
  title        = {PixArt-{\(\alpha\)}: Fast Training of Diffusion Transformer for Photorealistic
                  Text-to-Image Synthesis},
  booktitle    = {The Twelfth International Conference on Learning Representations,
                  {ICLR} 2024, Vienna, Austria, May 7-11, 2024},
  publisher    = {OpenReview.net},
  year         = {2024},
  url          = {https://openreview.net/forum?id=eAKmQPe3m1},
  bibsource    = {dblp computer science bibliography, https://dblp.org}
}

@inproceedings{chen2024pixartsigma,
  title={{PixArt-{\(\Sigma\)}}: Weak-to-Strong Training of Diffusion Transformer for 4K Text-to-Image Generation},
  author={Chen, Junsong and Ge, Chongjian and Xie, Enze and Wu, Yue and Yao, Lewei and Ren, Xiaozhe and Wang, Zhongdao and Luo, Ping and Lu, Huchuan and Li, Zhenguo},
  booktitle={European Conference on Computer Vision},
  pages={74--91},
  year={2025},
  organization={Springer}
}

@misc{li2024hunyuandit,
      title={Hunyuan-DiT: A Powerful Multi-Resolution Diffusion Transformer with Fine-Grained Chinese Understanding}, 
      author={Zhimin Li and Jianwei Zhang and Qin Lin and Jiangfeng Xiong and Yanxin Long and Xinchi Deng and Yingfang Zhang and Xingchao Liu and Minbin Huang and Zedong Xiao and Dayou Chen and Jiajun He and Jiahao Li and Wenyue Li and Chen Zhang and Rongwei Quan and Jianxiang Lu and Jiabin Huang and Xiaoyan Yuan and Xiaoxiao Zheng and Yixuan Li and Jihong Zhang and Chao Zhang and Meng Chen and Jie Liu and Zheng Fang and Weiyan Wang and Jinbao Xue and Yangyu Tao and Jianchen Zhu and Kai Liu and Sihuan Lin and Yifu Sun and Yun Li and Dongdong Wang and Mingtao Chen and Zhichao Hu and Xiao Xiao and Yan Chen and Yuhong Liu and Wei Liu and Di Wang and Yong Yang and Jie Jiang and Qinglin Lu},
      year={2024},
      eprint={2405.08748},
      archivePrefix={arXiv},
      primaryClass={cs.CV}
}

@inproceedings{hessel2021clipscore,
    title = "{CLIPS}core: A Reference-free Evaluation Metric for Image Captioning",
    author = "Hessel, Jack  and
      Holtzman, Ari  and
      Forbes, Maxwell  and
      Le Bras, Ronan  and
      Choi, Yejin",
    editor = "Moens, Marie-Francine  and
      Huang, Xuanjing  and
      Specia, Lucia  and
      Yih, Scott Wen-tau",
    booktitle = "Proceedings of the 2021 Conference on Empirical Methods in Natural Language Processing",
    month = nov,
    year = "2021",
    address = "Online and Punta Cana, Dominican Republic",
    publisher = "Association for Computational Linguistics",
    url = "https://aclanthology.org/2021.emnlp-main.595/",
    doi = "10.18653/v1/2021.emnlp-main.595",
    pages = "7514--7528",
}

@inproceedings{dosovitskiy2020image,
  author       = {Alexey Dosovitskiy and
                  Lucas Beyer and
                  Alexander Kolesnikov and
                  Dirk Weissenborn and
                  Xiaohua Zhai and
                  Thomas Unterthiner and
                  Mostafa Dehghani and
                  Matthias Minderer and
                  Georg Heigold and
                  Sylvain Gelly and
                  Jakob Uszkoreit and
                  Neil Houlsby},
  title        = {An Image is Worth 16x16 Words: Transformers for Image Recognition
                  at Scale},
  booktitle    = {9th International Conference on Learning Representations, {ICLR} 2021,
                  Virtual Event, Austria, May 3-7, 2021},
  publisher    = {OpenReview.net},
  year         = {2021},
  url          = {https://openreview.net/forum?id=YicbFdNTTy},
  bibsource    = {dblp computer science bibliography, https://dblp.org}
}

@inproceedings{xie2024sana,
  author       = {Enze Xie and
                  Junsong Chen and
                  Junyu Chen and
                  Han Cai and
                  Haotian Tang and
                  Yujun Lin and
                  Zhekai Zhang and
                  Muyang Li and
                  Ligeng Zhu and
                  Yao Lu and
                  Song Han},
  title        = {{SANA:} Efficient High-Resolution Text-to-Image Synthesis with Linear
                  Diffusion Transformers},
  booktitle    = {The Thirteenth International Conference on Learning Representations,
                  {ICLR} 2025, Singapore, April 24-28, 2025},
  publisher    = {OpenReview.net},
  year         = {2025},
  url          = {https://openreview.net/forum?id=N8Oj1XhtYZ},
  bibsource    = {dblp computer science bibliography, https://dblp.org}
}

@article{wu2023human,
  title={Human preference score v2: A solid benchmark for evaluating human preferences of text-to-image synthesis},
  author={Wu, Xiaoshi and Hao, Yiming and Sun, Keqiang and Chen, Yixiong and Zhu, Feng and Zhao, Rui and Li, Hongsheng},
  journal={arXiv preprint arXiv:2306.09341},
  year={2023}
}

@inproceedings{mills2025qua2sedimo,
        title = {Qua$^{2}$SeDiMo: Quantifiable Quantization Sensitivity of Diffusion Models},
        author = {Mills, Keith G. and Salameh, Mohammad and Chen, Ruichen and Hassanpour, Negar Lu, Wei and Niu, Di},
        booktitle={AAAI},
        year={2025}
}

@article{soro2024diffusion,
  title={Diffusion-based neural network weights generation},
  author={Soro, Bedionita and Andreis, Bruno and Lee, Hayeon and Jeong, Wonyong and Chong, Song and Hutter, Frank and Hwang, Sung Ju},
  journal={arXiv preprint arXiv:2402.18153},
  year={2024}
}

@article{ghasemabadi2025guided,
  title={Guided by Gut: Efficient Test-Time Scaling with Reinforced Intrinsic Confidence},
  author={Ghasemabadi, Amirhosein and Mills, Keith G and Li, Baochun and Niu, Di},
  journal={arXiv preprint arXiv:2505.20325},
  year={2025}
}

@article{chen2025fp4dit,
  title={Fp4dit: Towards effective floating point quantization for diffusion transformers},
  author={Chen, Ruichen and Mills, Keith G and Niu, Di},
  journal={arXiv preprint arXiv:2503.15465},
  year={2025}
}

@inproceedings{chen2025rettention,
        title = {Re-ttention: Ultra Sparse Visual Generation via Attention Statistical Reshape},
        author = {Chen, Ruichen and Mills, Keith G. and Jiang, Liyao and Gao, Chao and Niu, Di},
        booktitle={NeurIPS},
        year={2025}
}

@article{yang2024cogvideox,
  author       = {Zhuoyi Yang and
                  Jiayan Teng and
                  Wendi Zheng and
                  Ming Ding and
                  Shiyu Huang and
                  Jiazheng Xu and
                  Yuanming Yang and
                  Wenyi Hong and
                  Xiaohan Zhang and
                  Guanyu Feng and
                  Da Yin and
                  Yuxuan Zhang and
                  Weihan Wang and
                  Yean Cheng and
                  Bin Xu and
                  Xiaotao Gu and
                  Yuxiao Dong and
                  Jie Tang},
  title        = {CogVideoX: Text-to-Video Diffusion Models with An Expert Transformer},
  booktitle    = {The Thirteenth International Conference on Learning Representations,
                  {ICLR} 2025, Singapore, April 24-28, 2025},
  publisher    = {OpenReview.net},
  year         = {2025},
  url          = {https://openreview.net/forum?id=LQzN6TRFg9},
  bibsource    = {dblp computer science bibliography, https://dblp.org}
}

@article{ho2022classifier,
  title={Classifier-free diffusion guidance},
  author={Ho, Jonathan and Salimans, Tim},
  journal={arXiv preprint arXiv:2207.12598},
  year={2022}
}

@article{eyring2024reno,
  title={Reno: Enhancing one-step text-to-image models through reward-based noise optimization},
  author={Eyring, Luca and Karthik, Shyamgopal and Roth, Karsten and Dosovitskiy, Alexey and Akata, Zeynep},
  journal={Advances in Neural Information Processing Systems},
  volume={37},
  pages={125487--125519},
  year={2024}
}

@article{xu2023imagereward,
  title={Imagereward: Learning and evaluating human preferences for text-to-image generation},
  author={Xu, Jiazheng and Liu, Xiao and Wu, Yuchen and Tong, Yuxuan and Li, Qinkai and Ding, Ming and Tang, Jie and Dong, Yuxiao},
  journal={Advances in Neural Information Processing Systems},
  volume={36},
  pages={15903--15935},
  year={2023}
}

@inproceedings{wang2025silent,
  title={The silent assistant: Noisequery as implicit guidance for goal-driven image generation},
  author={Wang, Ruoyu and Huang, Huayang and Zhu, Ye and Russakovsky, Olga and Wu, Yu},
  booktitle={Proceedings of the IEEE/CVF International Conference on Computer Vision},
  pages={17618--17628},
  year={2025}
}

@inproceedings{zhou2025golden,
  title={Golden noise for diffusion models: A learning framework},
  author={Zhou, Zikai and Shao, Shitong and Bai, Lichen and Zhang, Shufei and Xu, Zhiqiang and Han, Bo and Xie, Zeke},
  booktitle={Proceedings of the IEEE/CVF International Conference on Computer Vision},
  pages={17688--17697},
  year={2025}
}

@inproceedings{guo2024initno,
  title={Initno: Boosting text-to-image diffusion models via initial noise optimization},
  author={Guo, Xiefan and Liu, Jinlin and Cui, Miaomiao and Li, Jiankai and Yang, Hongyu and Huang, Di},
  booktitle={Proceedings of the IEEE/CVF Conference on Computer Vision and Pattern Recognition},
  pages={9380--9389},
  year={2024}
}

@article{karras2022elucidating,
  title={Elucidating the design space of diffusion-based generative models},
  author={Karras, Tero and Aittala, Miika and Aila, Timo and Laine, Samuli},
  journal={Advances in neural information processing systems},
  volume={35},
  pages={26565--26577},
  year={2022}
}

@misc{pixartalpha_prompts,
  author       = {PixArt-Alpha Team},
  title        = {PixArt-Alpha Prompt List},
  howpublished = {\url{https://github.com/PixArt-alpha/PixArt-alpha/blob/master/asset/samples.txt}},
  note         = {Accessed: 2026-01-15},
  year         = {2024}
}

@article{song2020denoising,
  title={Denoising diffusion implicit models},
  author={Song, Jiaming and Meng, Chenlin and Ermon, Stefano},
  journal={arXiv preprint arXiv:2010.02502},
  year={2020}
}

@article{brown2024large,
  title={Large language monkeys: Scaling inference compute with repeated sampling},
  author={Brown, Bradley and Juravsky, Jordan and Ehrlich, Ryan and Clark, Ronald and Le, Quoc V and R{\'e}, Christopher and Mirhoseini, Azalia},
  journal={arXiv preprint arXiv:2407.21787},
  year={2024}
}

@inproceedings{ma2025hpsv3,
  title={Hpsv3: Towards wide-spectrum human preference score},
  author={Ma, Yuhang and Wu, Xiaoshi and Sun, Keqiang and Li, Hongsheng},
  booktitle={Proceedings of the IEEE/CVF International Conference on Computer Vision},
  pages={15086--15095},
  year={2025}
}

@misc{comfyui_contributors2025,
  title={ComfyUI: A Node-Based GUI for Stable Diffusion},
  author={ComfyUI Contributors},
  year={2025},
  url={https://github.com/Comfy-Org/ComfyUI}
}

@misc{autoamtic1111-stable-diffusion-webui,
  author = {AUTOMATIC1111},
  title = {Stable Diffusion WebUI},
  year = {2022},
  publisher = {GitHub},
  journal = {GitHub repository},
  howpublished = {\url{https://github.com/AUTOMATIC1111/stable-diffusion-webui}}
}

@misc{civitaiJuggernautRagnarok_by_RunDiffusion,
	author = {KandooAI},
	title = {{J}uggernaut {X}{L} - {R}agnarok\_by\_{R}un{D}iffusion | {S}table {D}iffusion {X}{L} {C}heckpoint | {C}ivitai --- civitai.com},
	howpublished = {\url{https://civitai.com/models/133005/juggernaut-xl}},
	year = {2025},
	note = {[Accessed 27-01-2026]},
}

@misc{civitaiCyberRealistic_Pony,
	author = {Cyberdelia},
	title = {{C}yber{R}ealistic {P}ony - | {S}table {D}iffusion {X}{L} {C}heckpoint | {C}ivitai --- civitai.com},
	howpublished = {\url{https://civitai.com/models/443821/cyberrealistic-pony}},
	year = {2026},
	note = {[Accessed 02-03-2026]},
}

@misc{civitaiIllustriousV20,
	author = {ONOMAAI},
	title = {{I}llustrious {X}{L} 2.0 - v2.0 | {I}llustrious {C}heckpoint | {C}ivitai --- civitai.com},
	howpublished = {\url{https://civitai.com/models/1369089/illustrious-xl-20}},
	year = {2025},
	note = {[Accessed 27-01-2026]},
}

@article{kirstain2023pick,
  title={Pick-a-pic: An open dataset of user preferences for text-to-image generation},
  author={Kirstain, Yuval and Polyak, Adam and Singer, Uriel and Matiana, Shahbuland and Penna, Joe and Levy, Omer},
  journal={Advances in neural information processing systems},
  volume={36},
  pages={36652--36663},
  year={2023}
}

@article{li2025noisear,
  title={Noisear: Autoregressing initial noise prior for diffusion models},
  author={Li, Zeming and Liu, Xiangyue and Zhang, Xiangyu and Tan, Ping and Shum, Heung-Yeung},
  journal={arXiv preprint arXiv:2506.01337},
  year={2025}
}

@article{eyring2025noise,
  title={Noise hypernetworks: Amortizing test-time compute in diffusion models},
  author={Eyring, Luca and Karthik, Shyamgopal and Dosovitskiy, Alexey and Ruiz, Nataniel and Akata, Zeynep},
  journal={arXiv preprint arXiv:2508.09968},
  year={2025}
}

@article{venkatraman2025outsourced,
  title={Outsourced diffusion sampling: Efficient posterior inference in latent spaces of generative models},
  author={Venkatraman, Siddarth and Hasan, Mohsin and Kim, Minsu and Scimeca, Luca and Sendera, Marcin and Bengio, Yoshua and Berseth, Glen and Malkin, Nikolay},
  journal={arXiv preprint arXiv:2502.06999},
  year={2025}
}

@article{kalaivanan2025ess,
  title={ESS-Flow: Training-free guidance of flow-based models as inference in source space},
  author={Kalaivanan, Adhithyan and Zhao, Zheng and Sj{\"o}lund, Jens and Lindsten, Fredrik},
  journal={arXiv preprint arXiv:2510.05849},
  year={2025}
}

@article{wang2025source,
  title={Source-Guided Flow Matching},
  author={Wang, Zifan and Harting, Alice and Barreau, Matthieu and Zavlanos, Michael M and Johansson, Karl H},
  journal={arXiv preprint arXiv:2508.14807},
  year={2025}
}

@article{smith2025calibrating,
  title={Calibrating Generative Models},
  author={Smith, Henry D and Diamant, Nathaniel L and Trippe, Brian L},
  journal={arXiv preprint arXiv:2510.10020},
  year={2025}
}

@article{om2025posterior,
  title={Posterior Inference in Latent Space for Scalable Constrained Black-box Optimization},
  author={Om, Kiyoung and Sim, Kyuil and Yun, Taeyoung and Kang, Hyeongyu and Park, Jinkyoo},
  journal={arXiv preprint arXiv:2507.00480},
  year={2025}
}

@inproceedings{chen2025sana,
  title={Sana-sprint: One-step diffusion with continuous-time consistency distillation},
  author={Chen, Junsong and Xue, Shuchen and Zhao, Yuyang and Yu, Jincheng and Paul, Sayak and Chen, Junyu and Cai, Han and Han, Song and Xie, Enze},
  booktitle={Proceedings of the IEEE/CVF International Conference on Computer Vision},
  pages={16185--16195},
  year={2025}
}

@misc{lykon2023dreamshaper,
  author = {Lykon},
  title = {DreamShaper - Stable Diffusion Fine-tune},
  year = {2023},
  publisher = {Civitai},
  howpublished = {\url{https://civitai.com/models/4384/dreamshaper}},
}

@misc{jiang2026raise,
    title={RAISE: Requirement-Adaptive Evolutionary Refinement for Training-Free Text-to-Image Alignment}, 
    author={Liyao Jiang and Ruichen Chen and Chao Gao and Di Niu},
    year={2026},
    eprint={2603.00483},
    archivePrefix={arXiv},
    primaryClass={cs.CV},
    url={https://arxiv.org/abs/2603.00483}, 
}

@article{daras2024survey,
  title={A survey on diffusion models for inverse problems},
  author={Daras, Giannis and Chung, Hyungjin and Lai, Chieh-Hsin and Mitsufuji, Yuki and Ye, Jong Chul and Milanfar, Peyman and Dimakis, Alexandros G and Delbracio, Mauricio},
  journal={arXiv preprint arXiv:2410.00083},
  year={2024}
}

@article{kong2024hunyuanvideo,
  title={Hunyuanvideo: A systematic framework for large video generative models},
  author={Kong, Weijie and Tian, Qi and Zhang, Zijian and Min, Rox and Dai, Zuozhuo and Zhou, Jin and Xiong, Jiangfeng and Li, Xin and Wu, Bo and Zhang, Jianwei and others},
  journal={arXiv preprint arXiv:2412.03603},
  year={2024}
}

@inproceedings{huang2023make,
  title={Make-an-audio: Text-to-audio generation with prompt-enhanced diffusion models},
  author={Huang, Rongjie and Huang, Jiawei and Yang, Dongchao and Ren, Yi and Liu, Luping and Li, Mingze and Ye, Zhenhui and Liu, Jinglin and Yin, Xiang and Zhao, Zhou},
  booktitle={International Conference on Machine Learning},
  pages={13916--13932},
  year={2023},
  organization={PMLR}
}

@article{wang2025diffusion,
  title={Diffusion models for molecules: A survey of methods and tasks},
  author={Wang, Liang and Song, Chao and Liu, Zhiyuan and Rong, Yu and Liu, Qiang and Wu, Shu},
  journal={arXiv preprint arXiv:2502.09511},
  year={2025}
}

@article{saxena2023surprising,
  title={The surprising effectiveness of diffusion models for optical flow and monocular depth estimation},
  author={Saxena, Saurabh and Herrmann, Charles and Hur, Junhwa and Kar, Abhishek and Norouzi, Mohammad and Sun, Deqing and Fleet, David J},
  journal={Advances in Neural Information Processing Systems},
  volume={36},
  pages={39443--39469},
  year={2023}
}

@inproceedings{dong2016accelerating,
  title={Accelerating the super-resolution convolutional neural network},
  author={Dong, Chao and Loy, Chen Change and Tang, Xiaoou},
  booktitle={European conference on computer vision},
  pages={391--407},
  year={2016},
  organization={Springer}
}

@article{wu2026densedpo,
  title={Densedpo: Fine-grained temporal preference optimization for video diffusion models},
  author={Wu, Ziyi and Kag, Anil and Skorokhodov, Ivan and Menapace, Willi and Mirzaei, Ashkan and Gilitschenski, Igor and Tulyakov, Sergey and Siarohin, Aliaksandr},
  journal={Advances in Neural Information Processing Systems},
  volume={38},
  pages={171632--171668},
  year={2026}
}

@InProceedings{Wallace_2024_CVPR,
    author    = {Wallace, Bram and Dang, Meihua and Rafailov, Rafael and Zhou, Linqi and Lou, Aaron and Purushwalkam, Senthil and Ermon, Stefano and Xiong, Caiming and Joty, Shafiq and Naik, Nikhil},
    title     = {Diffusion Model Alignment Using Direct Preference Optimization},
    booktitle = {Proceedings of the IEEE/CVF Conference on Computer Vision and Pattern Recognition (CVPR)},
    month     = {June},
    year      = {2024},
    pages     = {8228-8238}
}

@InProceedings{Zhang_2024_CVPR,
    author    = {Zhang, Sixian and Wang, Bohan and Wu, Junqiang and Li, Yan and Gao, Tingting and Zhang, Di and Wang, Zhongyuan},
    title     = {Learning Multi-Dimensional Human Preference for Text-to-Image Generation},
    booktitle = {Proceedings of the IEEE/CVF Conference on Computer Vision and Pattern Recognition (CVPR)},
    month     = {June},
    year      = {2024},
    pages     = {8018-8027}
}

@article{tishby2000information,
  title={The information bottleneck method},
  author={Tishby, Naftali and Pereira, Fernando C and Bialek, William},
  journal={arXiv preprint physics/0004057},
  year={2000}
}

@inproceedings{tishby2015deep,
  title={Deep learning and the information bottleneck principle},
  author={Tishby, Naftali and Zaslavsky, Noga},
  booktitle={2015 ieee information theory workshop (itw)},
  pages={1--5},
  year={2015},
  organization={Ieee}
}

@article{shannon1948mathematical,
  title={A mathematical theory of communication},
  author={Shannon, Claude Elwood},
  journal={The Bell system technical journal},
  volume={27},
  number={3},
  pages={379--423},
  year={1948},
  publisher={Nokia Bell Labs}
}

@article{zammit2025neural,
  title={Neural methods for amortized inference},
  author={Zammit-Mangion, Andrew and Sainsbury-Dale, Matthew and Huser, Rapha{\"e}l},
  journal={Annual Review of Statistics and Its Application},
  volume={12},
  number={1},
  pages={311--335},
  year={2025},
  publisher={Annual Reviews}
}

@inproceedings{wang2018lambdaloss,
  title={The lambdaloss framework for ranking metric optimization},
  author={Wang, Xuanhui and Li, Cheng and Golbandi, Nadav and Bendersky, Michael and Najork, Marc},
  booktitle={Proceedings of the 27th ACM international conference on information and knowledge management},
  pages={1313--1322},
  year={2018}
}

\newpage
\appendix

\section*{Supplementary Materials}

We provide additional implementation and training details, architecture specifications for each predictor, baseline implementation details, further ablation studies, and additional qualitative image samples.

\section{Additional Implementation Details}
\label{sec:supp_impl}

We provide further details on how we train our three predictor families---CA-I, CA-E, and EnCat (Sec.~\ref{sec:predictor_design})---along with their architecture specifications and the implementation of the standard baseline.

\subsection{Training Details}
\label{sec:supp_training}

\noindent\textbf{Dataset.} Each predictor is fit to a per-DM, per-target corpus of $100$k examples, assembled by drawing $5$k Pick-a-Pic~\cite{kirstain2023pick} training prompts and pairing every prompt with $20$ independently sampled Gaussian noise tensors. Every resulting image is scored by the target reward model in advance, so training consumes precomputed scalars and never invokes the generator. To make the predictor inputs cheap to load, we also precompute and cache each prompt embedding (from the DM's own text encoder) and its noise tensors to disk. We partition the $100$k examples $80\%/10\%/10\%$ at the prompt level, so a prompt never spans two splits. Targets are standardized to zero mean and unit variance using statistics from the training split only; we store these statistics in the checkpoint to recover native units at inference.

\noindent\textbf{Grouped batching.} Since the ranking term acts within a prompt, a batch must contain several noises that share the same prompt. We therefore assemble each batch from $k{=}12$ prompts at a time; with ${\sim}20$ noises available per prompt this yields ${\sim}240$ examples per batch, supplying enough same-prompt pairs for the within-group ranking objective to be well defined.

\noindent\textbf{Loss function.} For our main results we instantiate the ranking--regression objective of Eq.~\ref{eq:loss} with LambdaLoss@$5$~\cite{wang2018lambdaloss} as the ranking term and the MAE between predicted and normalized scores as the regression term, weighting the regression term by $\alpha=0.05$. Since each training prompt has 20 candidate noises, LambdaLoss@$5$ optimizes the top quarter of the within-prompt candidate list. For the SRCC variant studied in the ablations (Sec.~\ref{sec:supp_config}), we instead use the differentiable SRCC of Blondel~\etal~\cite{blondel2020fast} as the ranking term, computed \emph{per prompt group} and averaged over groups within the batch (regularization strength $10^{-2}$), again combined with MAE.

\noindent\textbf{Optimization.} Predictors are optimized with AdamW~\cite{loshchilov19AdamW} (learning rate $10^{-4}$, weight decay $10^{-8}$) under gradient-norm clipping at $1.0$. The learning rate is halved whenever the validation primary metric does not improve for five consecutive epochs. Training runs for at most $80$ epochs, and we retain the epoch that maximizes that metric (NDCG for LambdaLoss runs, SRCC otherwise).

\noindent\textbf{Validation metrics.} We track three quantities on the held-out validation prompts. A global Spearman correlation (SRCC) summarizes how well the predicted scores order the entire validation pool, while per-prompt NDCG@$K$ ($K\in\{3,5\}$), macro-averaged across prompts, checks whether the noises ranked highest within each prompt are the correct ones. We additionally report MAE in de-normalized units to capture absolute regression error.

\subsection{Predictor Architectures}
\label{sec:supp_arch}

\noindent\textbf{CA-I.} We read the first internal cross-attention map and reduce each attention row to per-head, per-token entropy, standard deviation, and maximum statistics. A head compression factor $r$ ($h \bmod r = 0$) sums adjacent groups of $r$ heads, reducing the head dimension from $h$ to $\sfrac{h}{r}$; the flattened statistics ($\sfrac{h}{r}\cdot|p_{tok}|\cdot n_{\text{stat}}$) feed the same MLP head as EnCat. The main configuration uses all three statistics with $r{=}1$.

\noindent\textbf{CA-E.} A patchify convolution ($P_k{=}4$) maps the latent to a patch sequence; a linear layer projects prompt tokens to dimension $256$; an $8$-head cross-attention block attends patches to prompt tokens. The attention map is reduced to entropy statistics with head compression $r{=}2$ and scored by the same MLP head, without invoking the diffusion model.

\noindent\textbf{EnCat.} The noise encoder is a lightweight convolutional network that downsamples the $4{\times}128{\times}128$ latent to a $1024$-d vector. In parallel, the text encoder applies a shared per-token MLP that maps each prompt-token embedding to a scalar, producing a length-$|p_{tok}|$ vector. The two vectors are concatenated and passed to a small MLP head (hidden widths $512$, $256$, $64$; SiLU activations; dropout $0.1$) that outputs the scalar score. The embedding dimension $d_c$ and sequence length $|p_{tok}|$ follow each DM's native text encoder.

\subsection{Baseline Implementation}
\label{sec:supp_baselines}

The standard baseline runs the Diffusers~\cite{von-platen-etal-2022-diffusers} pipeline with no noise selection. We seed the noise generator from each prompt's integer index and apply this same seeding rule to every method, so the baseline and all predictors observe an identical candidate-noise sequence for a given prompt.

\section{Additional Ablation Studies}
\label{sec:supp_ablations}

This section collects the ablations behind our design and data choices: the per-predictor configuration sweeps (Sec.~\ref{sec:supp_config}), the fixed-budget prompt/noise study (Sec.~\ref{sec:supp_budget}), and the metric-agreement counts for PickScore-trained predictors (Sec.~\ref{sec:supp_metric_agreement_ps}).

\subsection{Predictor Configuration Ablations}
\label{sec:supp_config}

We sweep the design choices for each predictor under the MAE+SRCC objective and retrain the selected configuration under the main-paper LambdaLoss objective for the results in the main text. The configuration used in the main paper is indicated in each table's caption.

\noindent\textbf{CA-I.} Table~\ref{table:camap_i_ablation} sweeps the attention statistic set and head compression on SDXL and PixArt-$\Sigma$; we use entropy+std+max with $r{=}1$.

\begin{table}[htbp]
\begin{center}
\caption{CA-I internal cross-attention configuration ablation, trained with MAE+SRCC. We vary the attention statistic set (E: entropy; Esm: entropy+std+max) and the head compression $r\in\{1,2\}$. Bold marks the configuration used in the main paper (Esm, $r{=}1$). MAE is in each target's native score scale.}
\label{table:camap_i_ablation}
\scalebox{\scaleboxratio}{
\begin{tabular}{cclccccc}
\toprule
Model & Target & Config & Params & SRCC & NDCG@$3$ & NDCG@$5$ & MAE \\
\midrule
\multirow{8}{*}{SDXL}
& \multirow{4}{*}{HPSv2}
& E, $r{=}1$ & 0.54M & 0.098 & 0.594 & 0.622 & 0.0331 \\
& & E, $r{=}2$ & 0.34M & 0.133 & 0.584 & 0.608 & 0.0330 \\
& & Esm, $r{=}1$ & 1.33M & $-0.023$ & 0.578 & 0.608 & 0.0331 \\
& & Esm, $r{=}2$ & 0.74M & 0.019 & 0.579 & 0.610 & 0.0333 \\
\cmidrule(lr){2-8}
& \multirow{4}{*}{PickScore}
& E, $r{=}1$ & 0.54M & 0.350 & 0.576 & 0.603 & 1.119 \\
& & E, $r{=}2$ & 0.34M & 0.311 & 0.573 & 0.601 & 1.120 \\
& & Esm, $r{=}1$ & 1.33M & 0.279 & 0.562 & 0.593 & 1.121 \\
& & Esm, $r{=}2$ & 0.74M & 0.349 & 0.572 & 0.602 & 1.119 \\
\midrule
\multirow{8}{*}{PixArt-$\Sigma$}
& \multirow{4}{*}{HPSv2}
& E, $r{=}1$ & 2.61M & 0.087 & 0.567 & 0.596 & 0.0348 \\
& & E, $r{=}2$ & 1.38M & 0.016 & 0.569 & 0.599 & 0.0349 \\
& & Esm, $r{=}1$ & 7.52M & 0.108 & 0.566 & 0.596 & 0.0348 \\
& & Esm, $r{=}2$ & 3.83M & 0.165 & 0.565 & 0.595 & 0.0349 \\
\cmidrule(lr){2-8}
& \multirow{4}{*}{PickScore}
& E, $r{=}1$ & 2.61M & 0.009 & 0.547 & 0.579 & 1.242 \\
& & E, $r{=}2$ & 1.38M & $-0.010$ & 0.537 & 0.575 & 1.242 \\
& & Esm, $r{=}1$ & 7.52M & 0.069 & 0.548 & 0.579 & 1.243 \\
& & Esm, $r{=}2$ & 3.83M & 0.024 & 0.548 & 0.580 & 1.245 \\
\bottomrule
\end{tabular}
}
\end{center}
\end{table}

\noindent\textbf{CA-E.} Table~\ref{table:camap_e_ablation} sweeps the statistic set, head compression, and patch size; we use entropy-only with $P_k{=}4$ and $r{=}2$.

\begin{table}[htbp]
\begin{center}
\caption{CA-E external patch cross-attention configuration ablation, trained with MAE+SRCC. We vary the attention statistic set (E: entropy; Esm: entropy+std+max), the patch size $P_k\in\{4,8,16\}$, and the head compression $r$. The configuration used in the main paper is E, $P_k{=}4$, $r{=}2$. MAE is in each target's native score scale.}
\label{table:camap_e_ablation}
\scalebox{\scaleboxratio}{
\begin{tabular}{ccccccccc}
\toprule
Model & Target & Stat & $P_k$ & $r$ & Params & SRCC & NDCG@$5$ & MAE \\
\midrule
\multirow{12}{*}{SDXL}
& \multirow{6}{*}{HPSv2}
& E & 4 & 2 & 1.11M & 0.663 & 0.611 & 0.0250 \\
& & Esm & 4 & 1 & 1.90M & 0.656 & 0.613 & 0.0250 \\
& & E & 8 & 2 & 1.16M & 0.641 & 0.618 & 0.0259 \\
& & Esm & 8 & 1 & 1.95M & 0.633 & 0.616 & 0.0259 \\
& & E & 16 & 2 & 1.36M & 0.606 & 0.619 & 0.0273 \\
& & Esm & 16 & 1 & 2.15M & 0.602 & 0.624 & 0.0265 \\
\cmidrule(lr){2-9}
& \multirow{6}{*}{PickScore}
& E & 4 & 2 & 1.11M & 0.598 & 0.607 & 0.899 \\
& & Esm & 4 & 1 & 1.90M & 0.590 & 0.606 & 0.917 \\
& & E & 8 & 2 & 1.16M & 0.589 & 0.604 & 0.905 \\
& & Esm & 8 & 1 & 1.95M & 0.577 & 0.606 & 0.923 \\
& & E & 16 & 2 & 1.36M & 0.563 & 0.607 & 0.939 \\
& & Esm & 16 & 1 & 2.15M & 0.565 & 0.606 & 0.938 \\
\midrule
\multirow{12}{*}{PixArt-$\Sigma$}
& \multirow{6}{*}{HPSv2}
& E & 4 & 2 & 2.09M & 0.623 & 0.610 & 0.0273 \\
& & Esm & 4 & 1 & 5.17M & 0.639 & 0.597 & 0.0267 \\
& & E & 8 & 2 & 2.14M & 0.604 & 0.594 & 0.0283 \\
& & Esm & 8 & 1 & 5.21M & 0.648 & 0.588 & 0.0266 \\
& & E & 16 & 2 & 2.34M & 0.610 & 0.592 & 0.0278 \\
& & Esm & 16 & 1 & 5.41M & 0.625 & 0.588 & 0.0272 \\
\cmidrule(lr){2-9}
& \multirow{6}{*}{PickScore}
& E & 4 & 2 & 2.09M & 0.642 & 0.572 & 0.943 \\
& & Esm & 4 & 1 & 5.17M & 0.618 & 0.583 & 0.977 \\
& & E & 8 & 2 & 2.14M & 0.614 & 0.583 & 0.978 \\
& & Esm & 8 & 1 & 5.21M & 0.601 & 0.579 & 1.007 \\
& & E & 16 & 2 & 2.34M & 0.596 & 0.581 & 0.999 \\
& & Esm & 16 & 1 & 5.41M & 0.614 & 0.570 & 0.974 \\
\bottomrule
\end{tabular}
}
\end{center}
\end{table}

\subsection{Prompt/Noise Allocation}
\label{sec:supp_budget}
Table~\ref{table:dataset_budget} allocates a fixed $100$K-sample training budget between prompt diversity and per-prompt noise coverage. We train an EnCat predictor on SDXL with the PickScore target under the LambdaLoss@$5$+$0.05\,$MAE objective, split prompts by ID, and select on NDCG@$10$. For each budget we report offline NDCG@$5$, NDCG@$10$, SRCC, and MAE (in the native PickScore scale) on the held-out test split. 

\begin{table}[t!]
\begin{center}
\caption{Prompt/noise budget ablation under a fixed 100{,}000-sample training set. Bold marks the budget used in the main paper.}
\label{table:dataset_budget}
\scalebox{\scaleboxratio}{
\begin{tabular}{ccccccc}
\toprule
Prompts & Noises/prompt & Total & NDCG@$5$ & NDCG@$10$ & SRCC & MAE \\
\midrule
1{,}000  & 100 & 100K & 0.3202 & 0.3299 & 0.0092 & 2.3727 \\
2{,}000  & 50  & 100K & 0.4282 & 0.4466 & 0.0059 & 2.5035 \\
\textbf{5{,}000} & \textbf{20} & \textbf{100K} & \textbf{0.6275} & \textbf{0.6636} & \textbf{0.0098} & \textbf{0.811} \\
10{,}000 & 10  & 100K & 0.5976 & 0.7935 & $-0.0048$ & 1.1917 \\
\bottomrule
\end{tabular}
}
\end{center}
\end{table}

\begin{table}[t!]
\begin{center}
\caption{Metric-agreement counts for SDXL PickScore-trained predictors on 100 Pick-a-Pic prompts. Each row reports the number of prompts, out of 100, for which the predictor-selected image improves over the Standard baseline under every metric in the listed set.}
\label{table:metric_agreement_pickscore}
\scalebox{\scaleboxratio}{
\begin{tabular}{lccc}
\toprule
Human Preference Metrics & CA-I & CA-E & EnCat \\
\midrule
HPSv2 & 50 & 53 & 55 \\
HPSv3 & 52 & 55 & 58 \\
PS & 48 & 51 & 54 \\
IR & 42 & 44 & 46 \\
\midrule
HPSv2 + HPSv3 & 35 & 38 & 40 \\
HPSv2 + PS & 33 & 35 & 37 \\
HPSv3 + PS & 34 & 36 & 39 \\
HPSv2 + IR & 27 & 29 & 31 \\
HPSv3 + IR & 28 & 30 & 32 \\
PS + IR & 26 & 28 & 30 \\
\midrule
HPSv2 + HPSv3 + PS & 26 & 28 & 30 \\
HPSv2 + HPSv3 + IR & 19 & 21 & 23 \\
HPSv2 + PS + IR & 18 & 20 & 22 \\
HPSv3 + PS + IR & 20 & 22 & 24 \\
\midrule
All & 13 & 15 & 17 \\
\bottomrule
\end{tabular}%
}
\end{center}
\end{table}

\subsection{Metric Agreement for PickScore-Trained Predictors}
\label{sec:supp_metric_agreement_ps}

Table~\ref{table:metric_agreement_pickscore} reports the metric-agreement counts for the SDXL predictors trained on PickScore, the counterpart to the HPSv2 table in the main paper (Table~\ref{table:metric_agreement}). The same pattern holds: EnCat and CA-E preserve higher agreement than CA-I, and ImageReward remains the least concordant metric.

\section{Additional Qualitative Examples}
\label{sec:supp_qual}

\begin{figure}[htbp]
    \centering
    \setlength{\tabcolsep}{2pt}
    \renewcommand{\arraystretch}{1.15}
    \small

    \newcommand{\imgw}{0.24\textwidth}

    \begin{tabular}{@{}c c c c@{}}
        Standard & CA-I & CA-E & EnCat \\

        \includegraphics[width=\imgw]{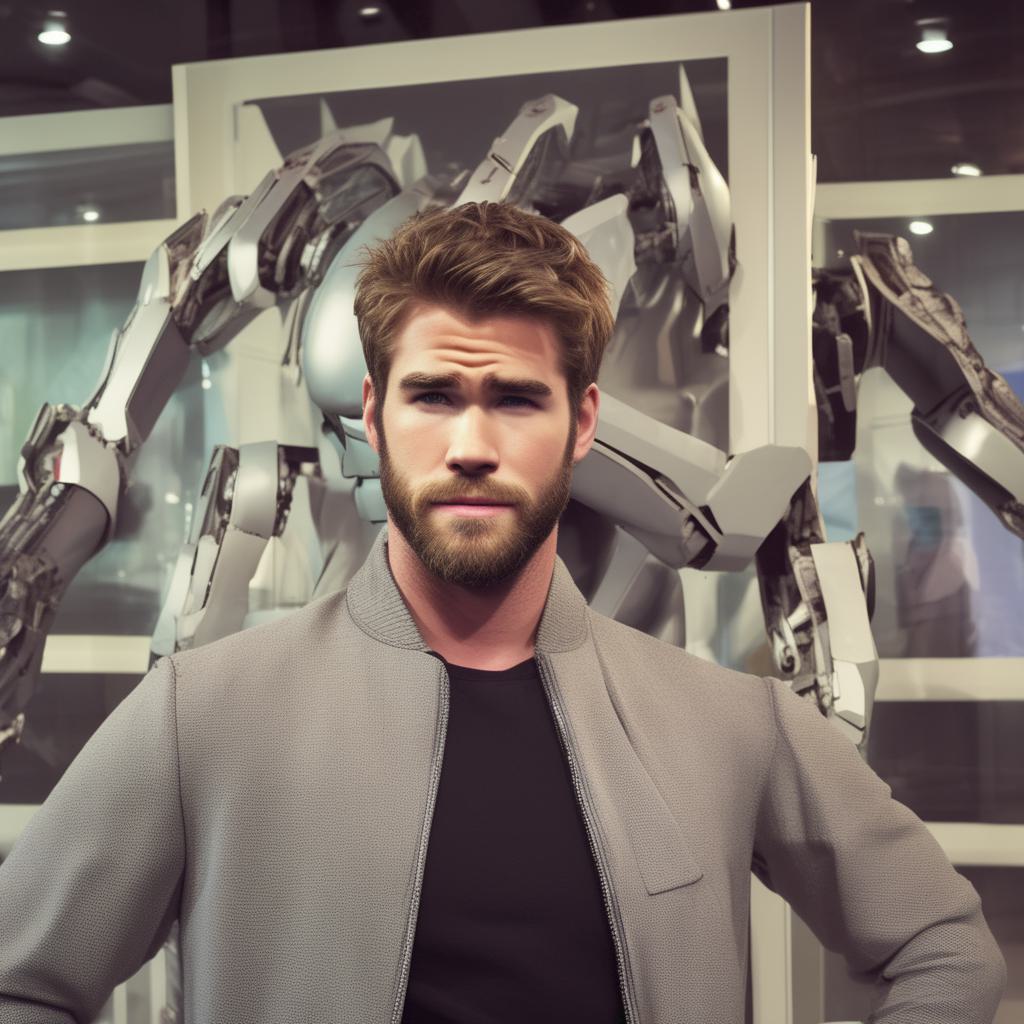} &
        \includegraphics[width=\imgw]{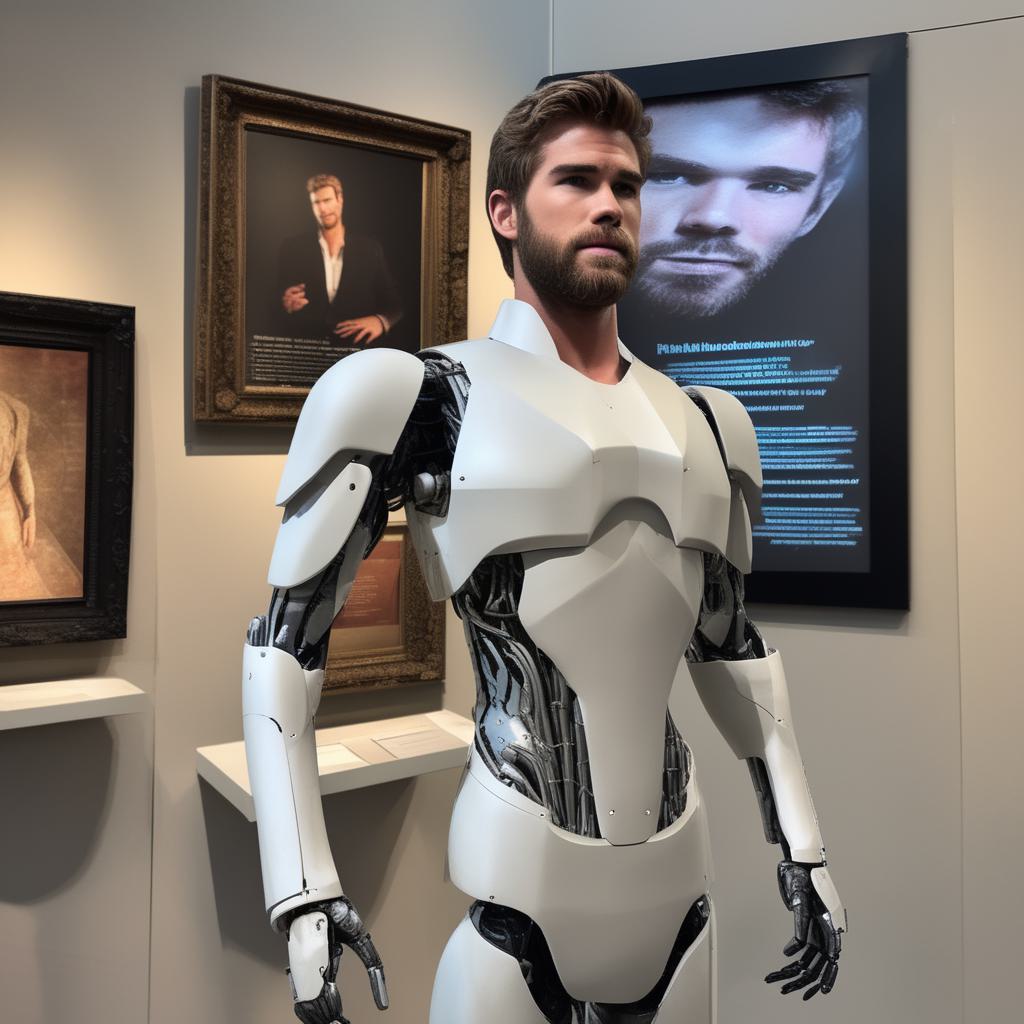} &
        \includegraphics[width=\imgw]{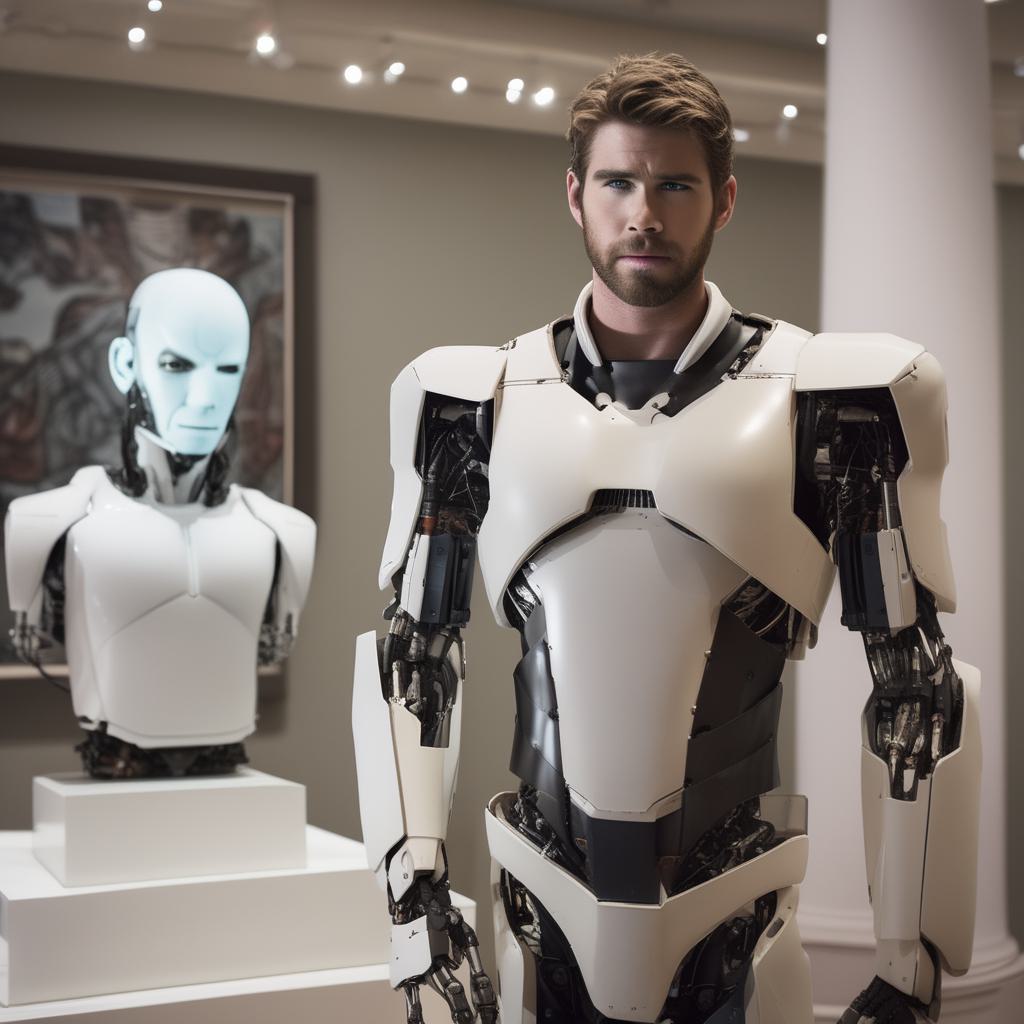} &
        \includegraphics[width=\imgw]{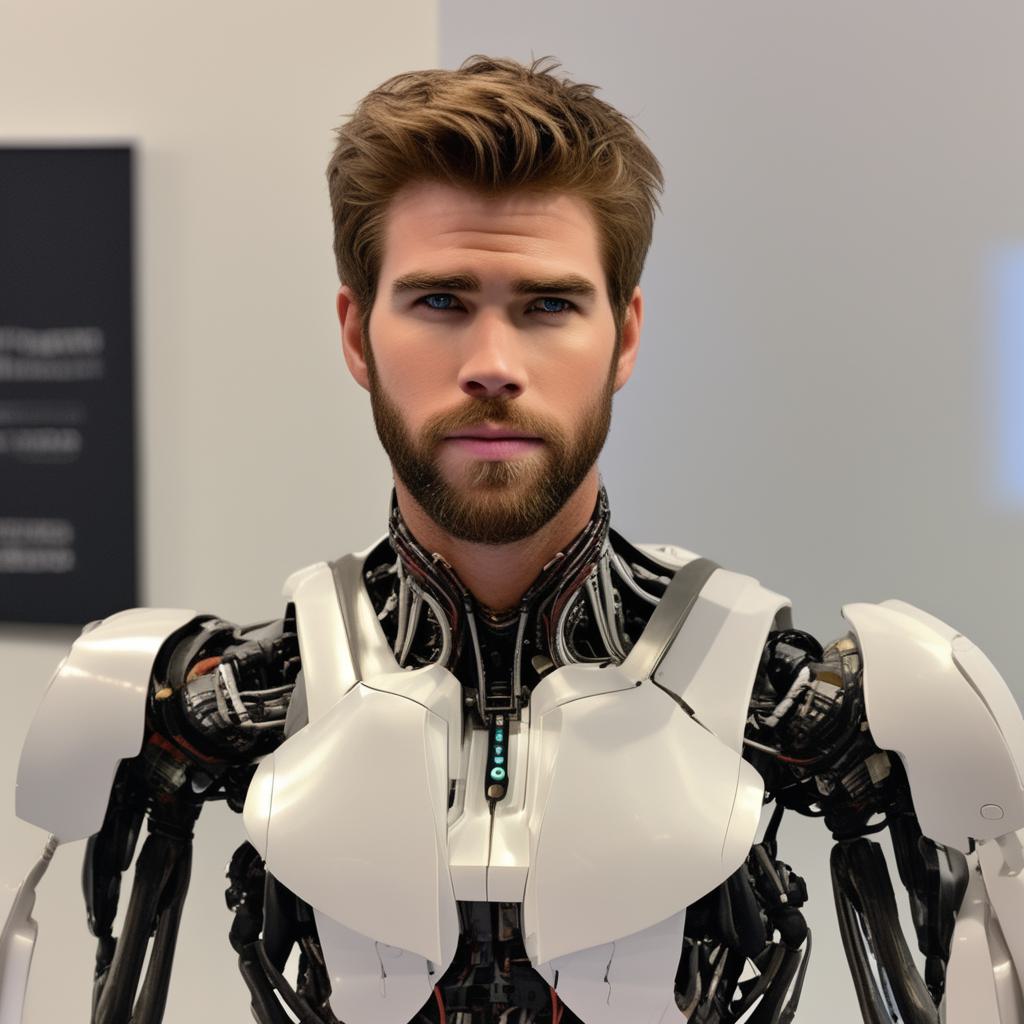} \\
        \multicolumn{4}{@{}p{0.99\textwidth}@{}}{
            SDXL: ``A photo of a male android, half robot and half humanoid, resembling actor Liam Hemsworth, posing stoically on display at a museum.''
        } \\[2mm]

        \includegraphics[width=\imgw]{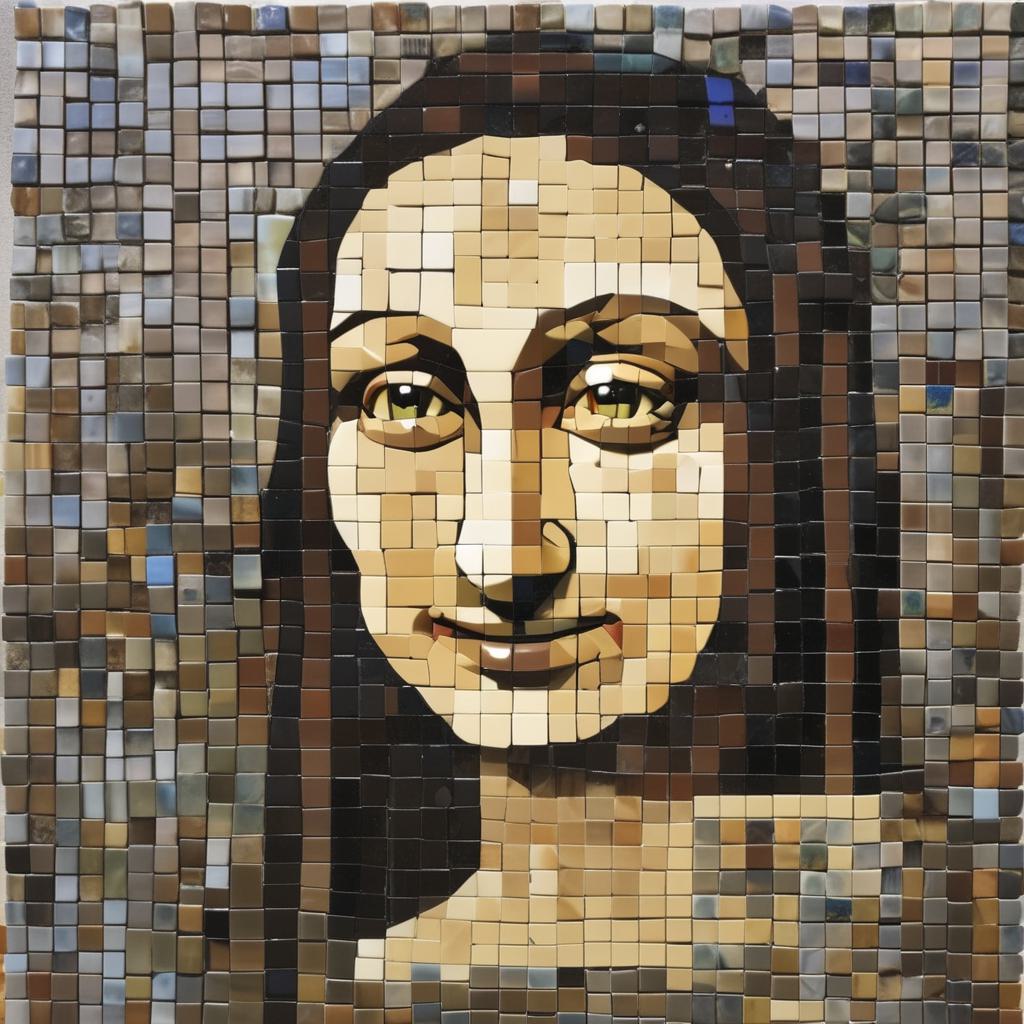} &
        \includegraphics[width=\imgw]{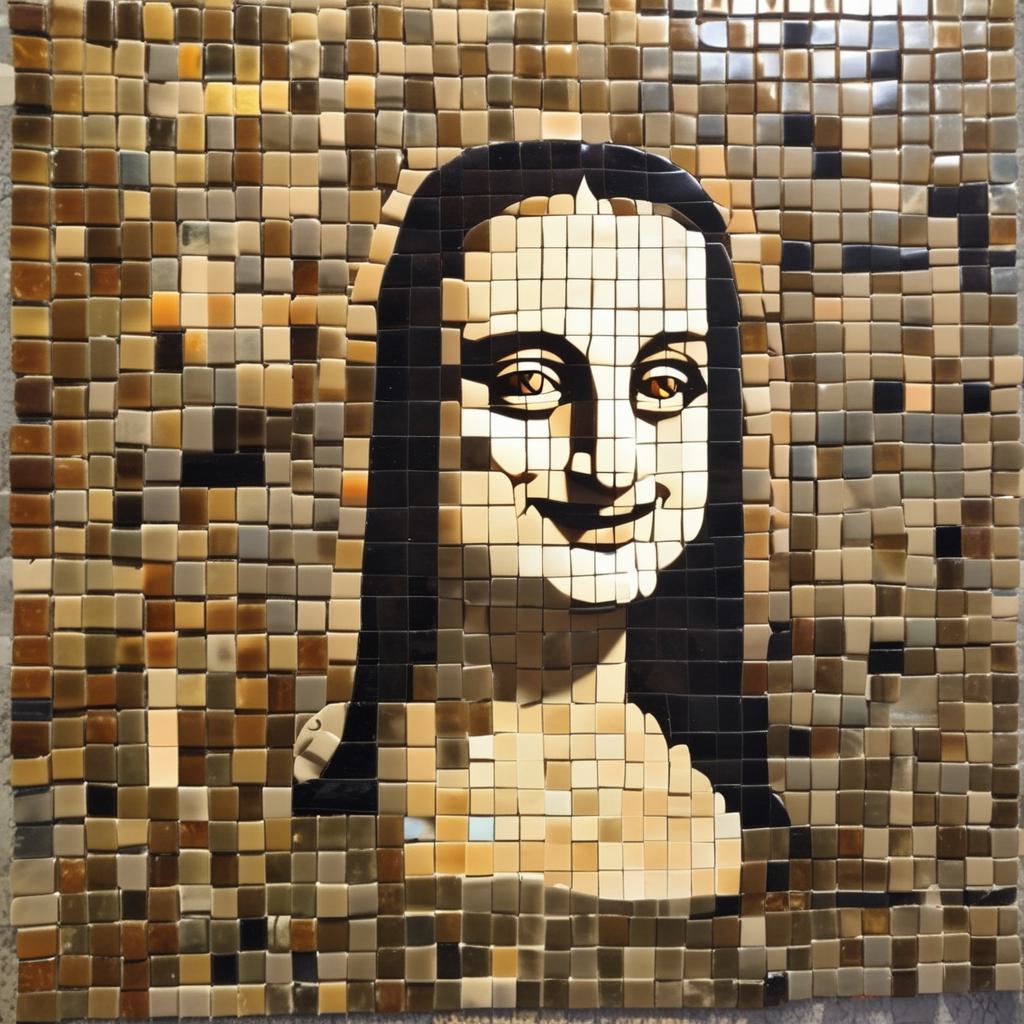} &
        \includegraphics[width=\imgw]{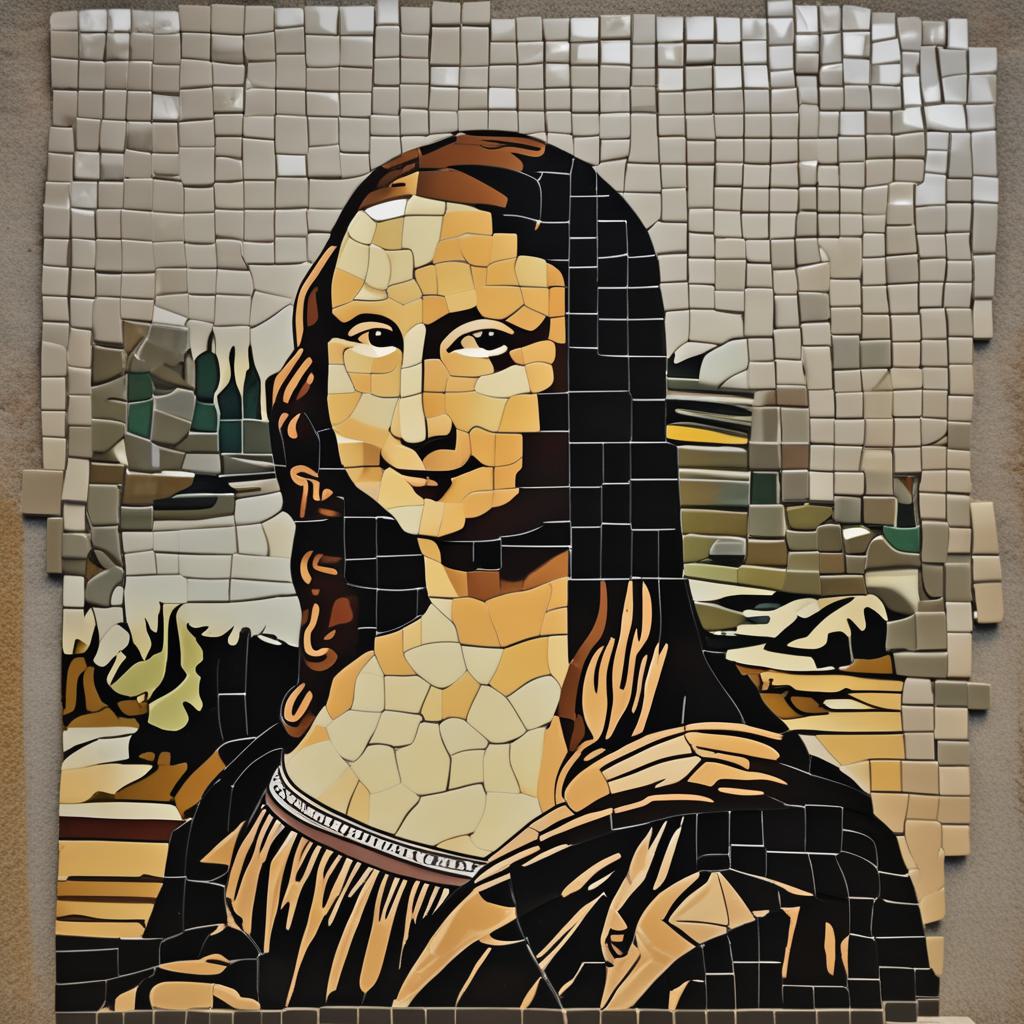} &
        \includegraphics[width=\imgw]{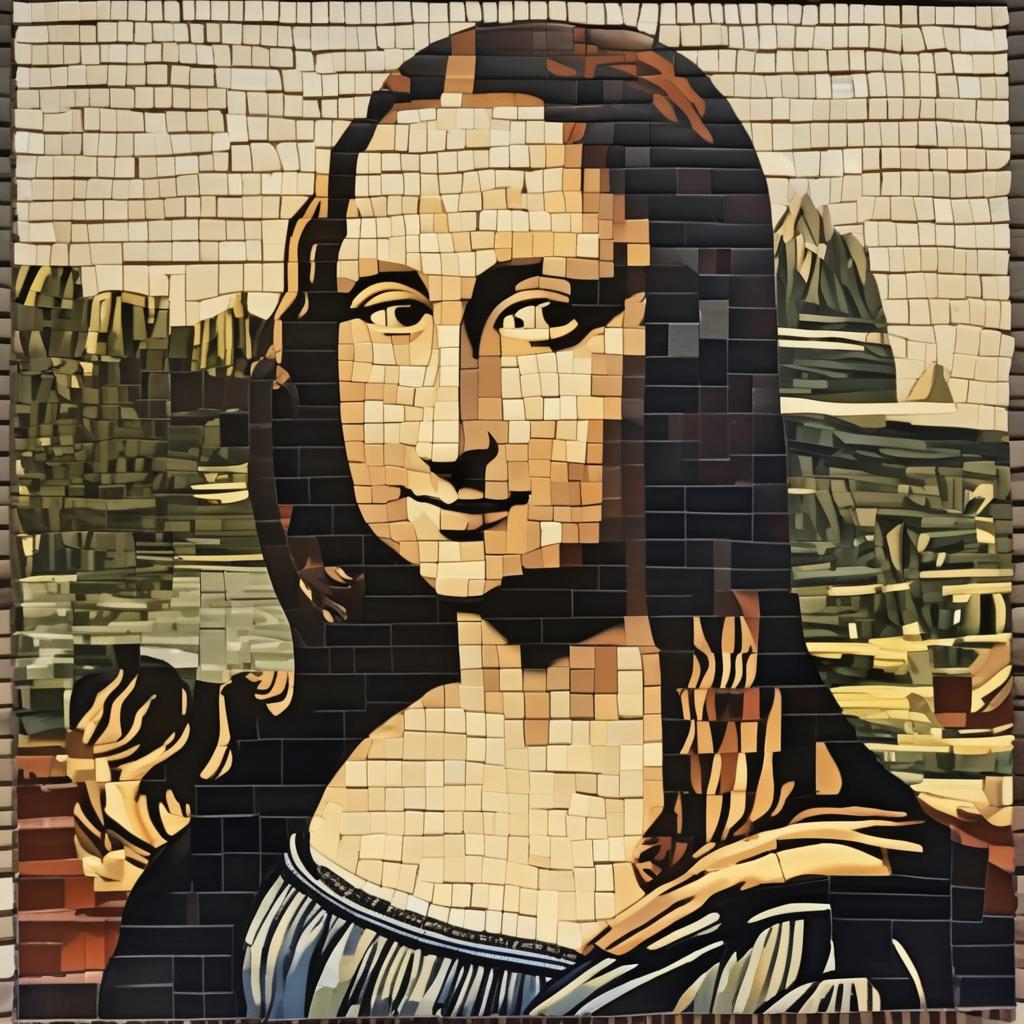} \\
        \multicolumn{4}{@{}p{0.99\textwidth}@{}}{
            SDXL: ``A ceramic glass mosaic depicts Mona Lisa's smile.''
        } \\[2mm]

        \includegraphics[width=\imgw]{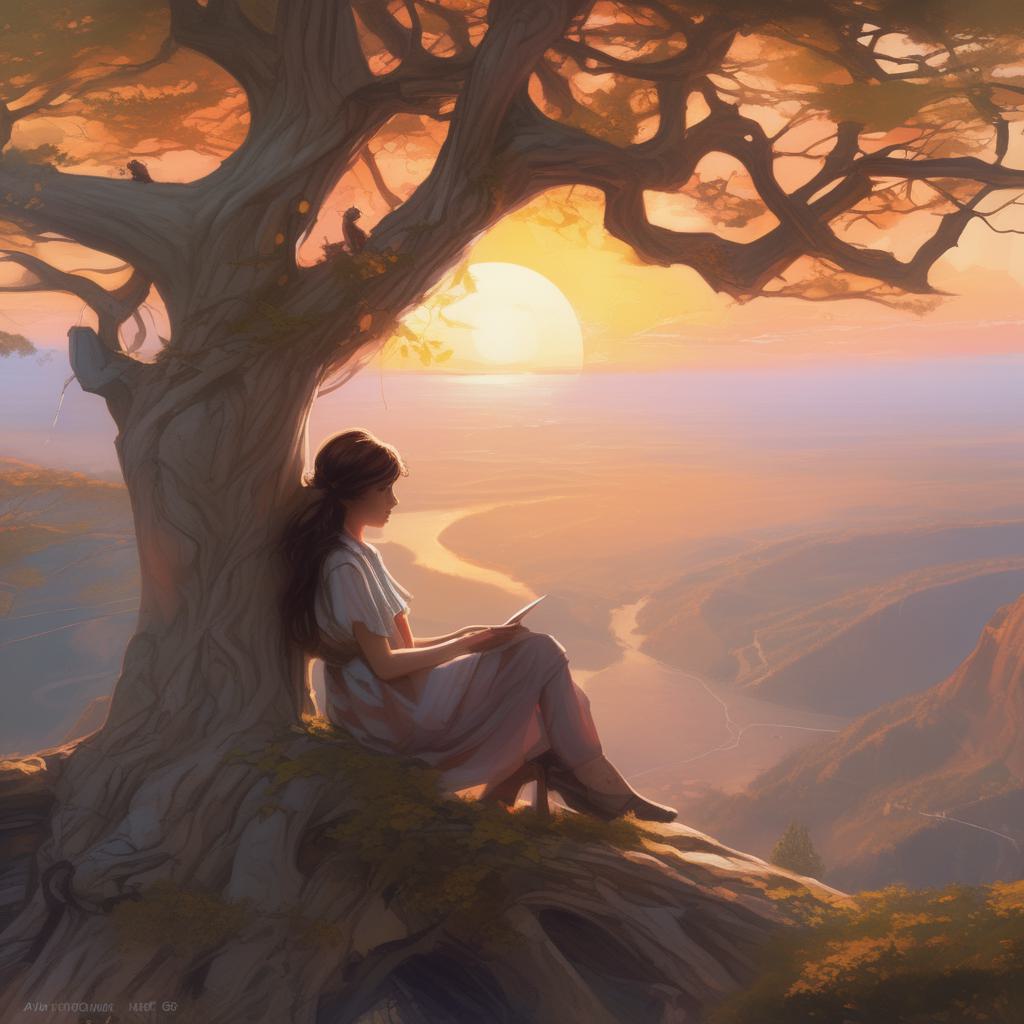} &
        \includegraphics[width=\imgw]{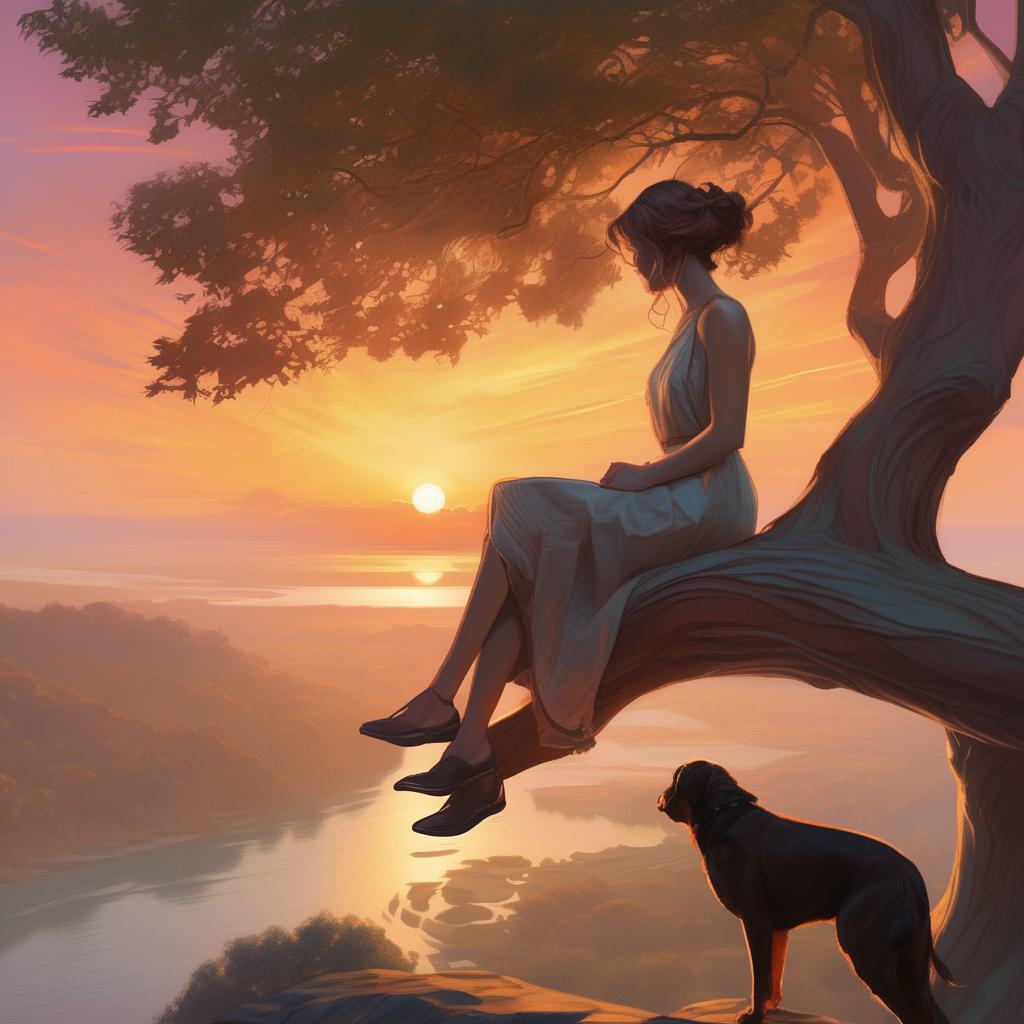} &
        \includegraphics[width=\imgw]{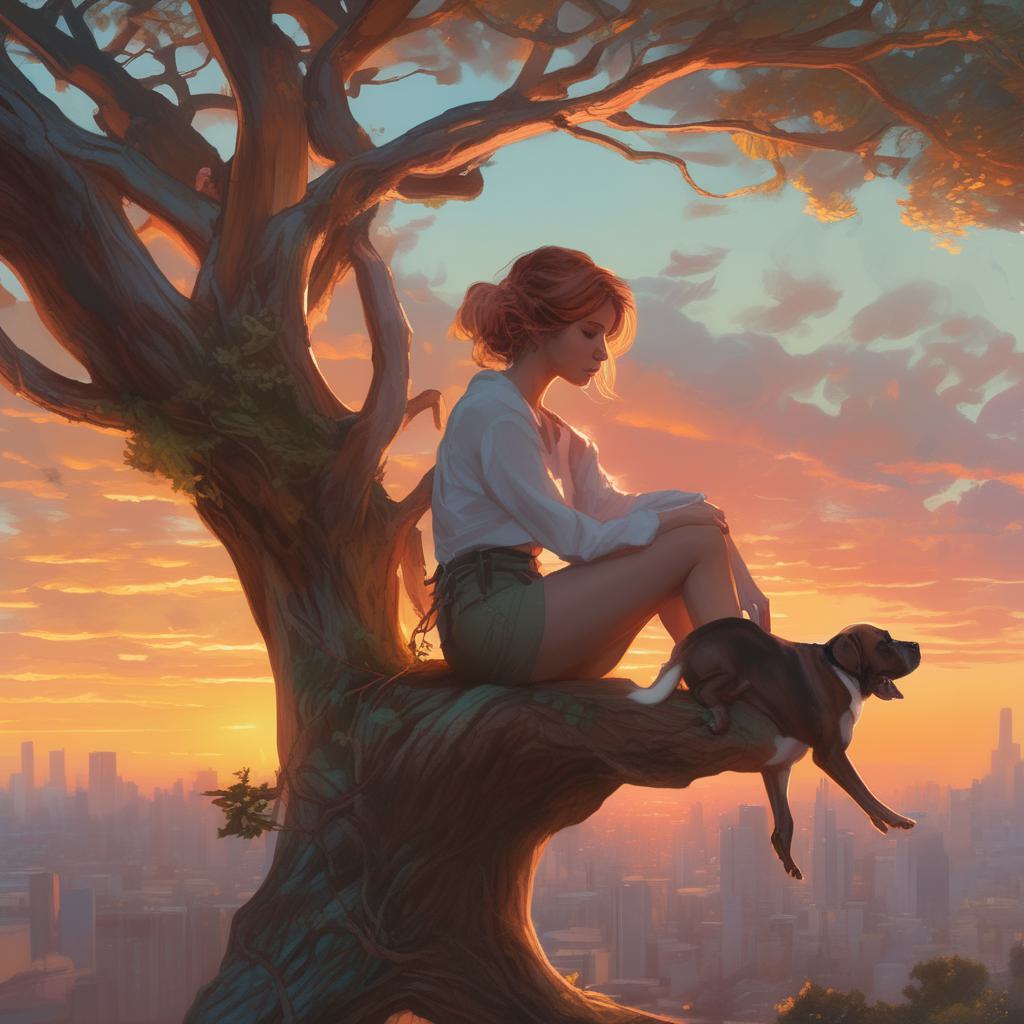} &
        \includegraphics[width=\imgw]{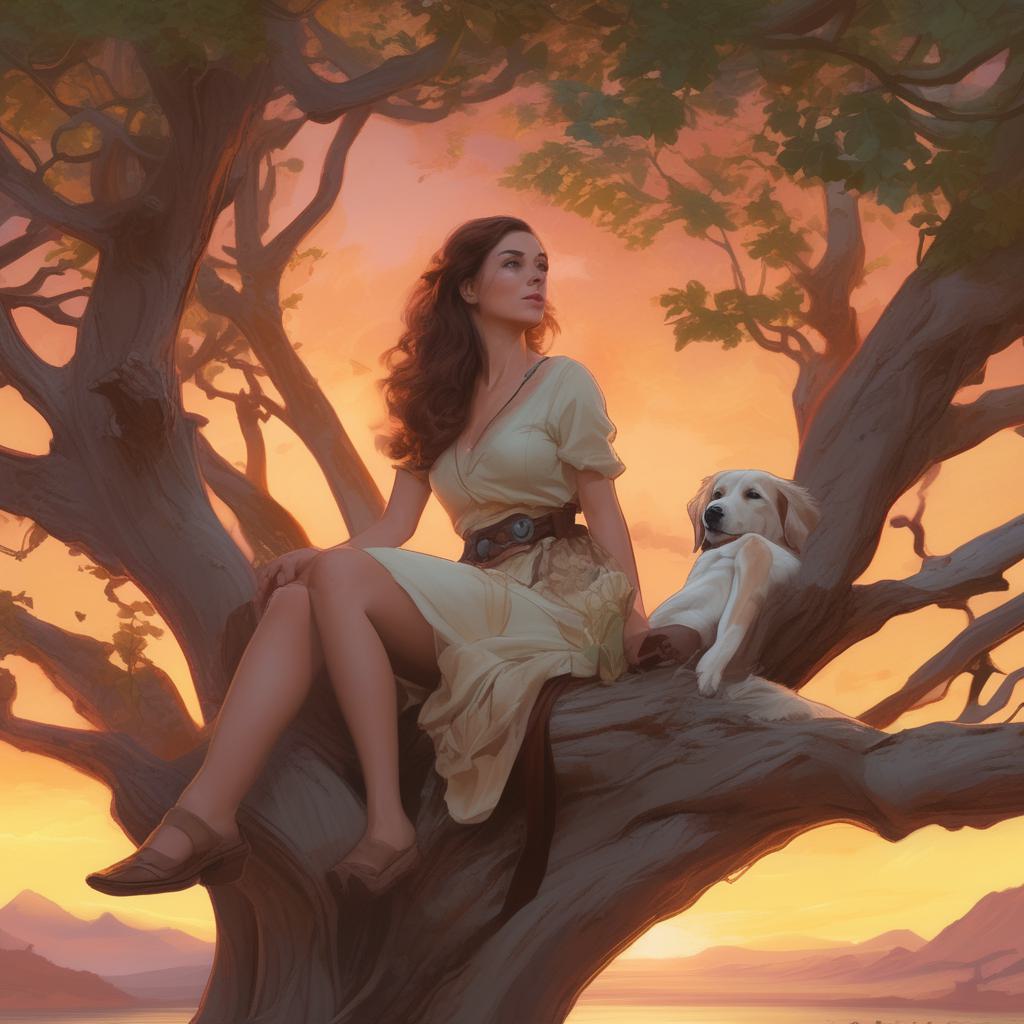} \\
        \multicolumn{4}{@{}p{0.99\textwidth}@{}}{
            SDXL: ``A woman and her dog sit on a tree and watch the sunset in a digital painting by artgerm, greg rutkowski, and Alphonse Mucha''
        } \\[2mm]

        \includegraphics[width=\imgw]{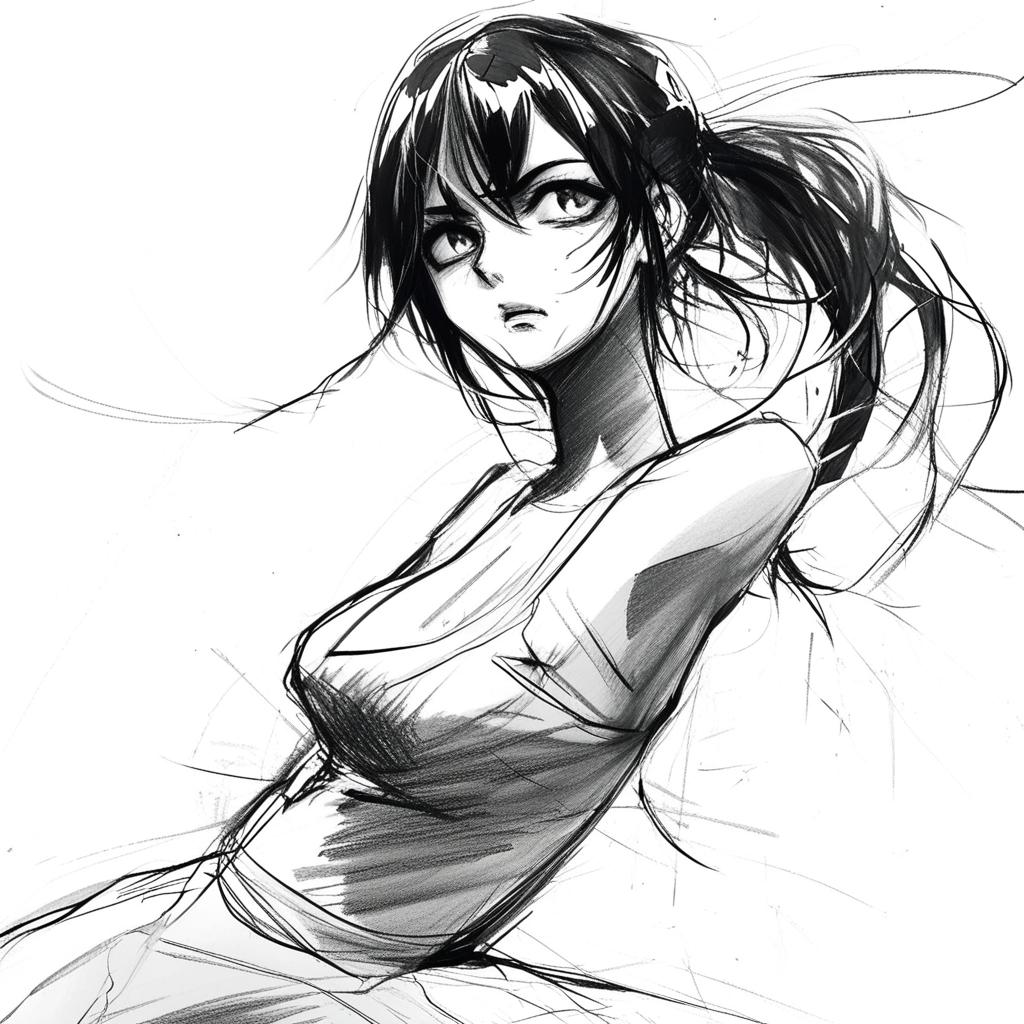} &
        \includegraphics[width=\imgw]{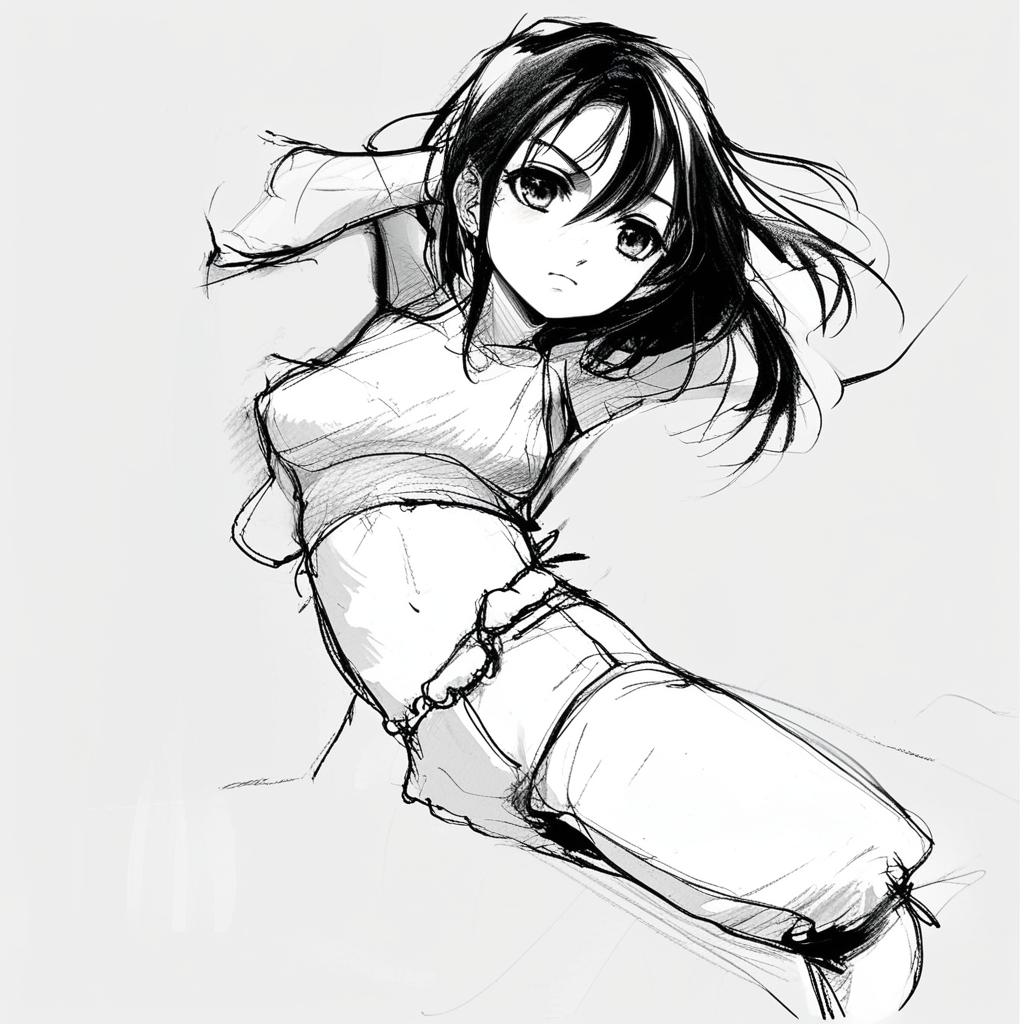} &
        \includegraphics[width=\imgw]{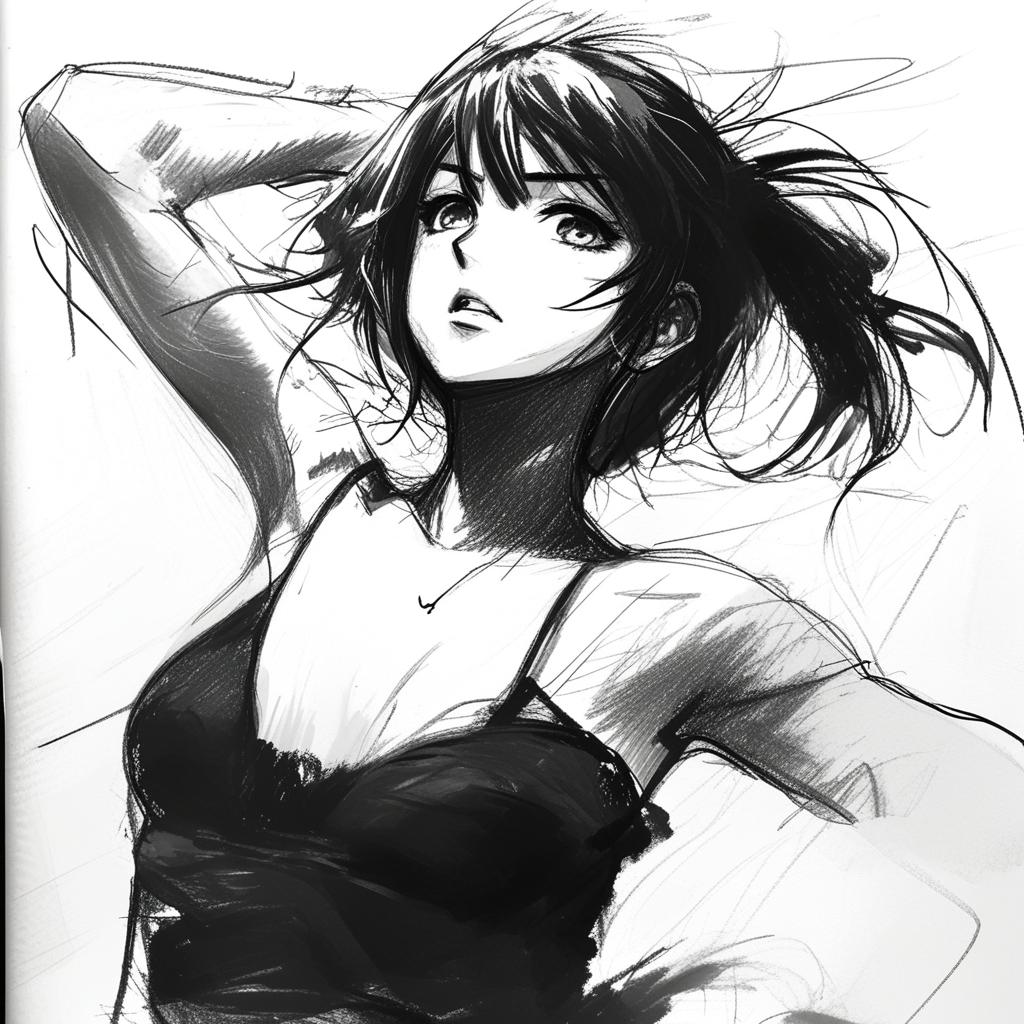} &
        \includegraphics[width=\imgw]{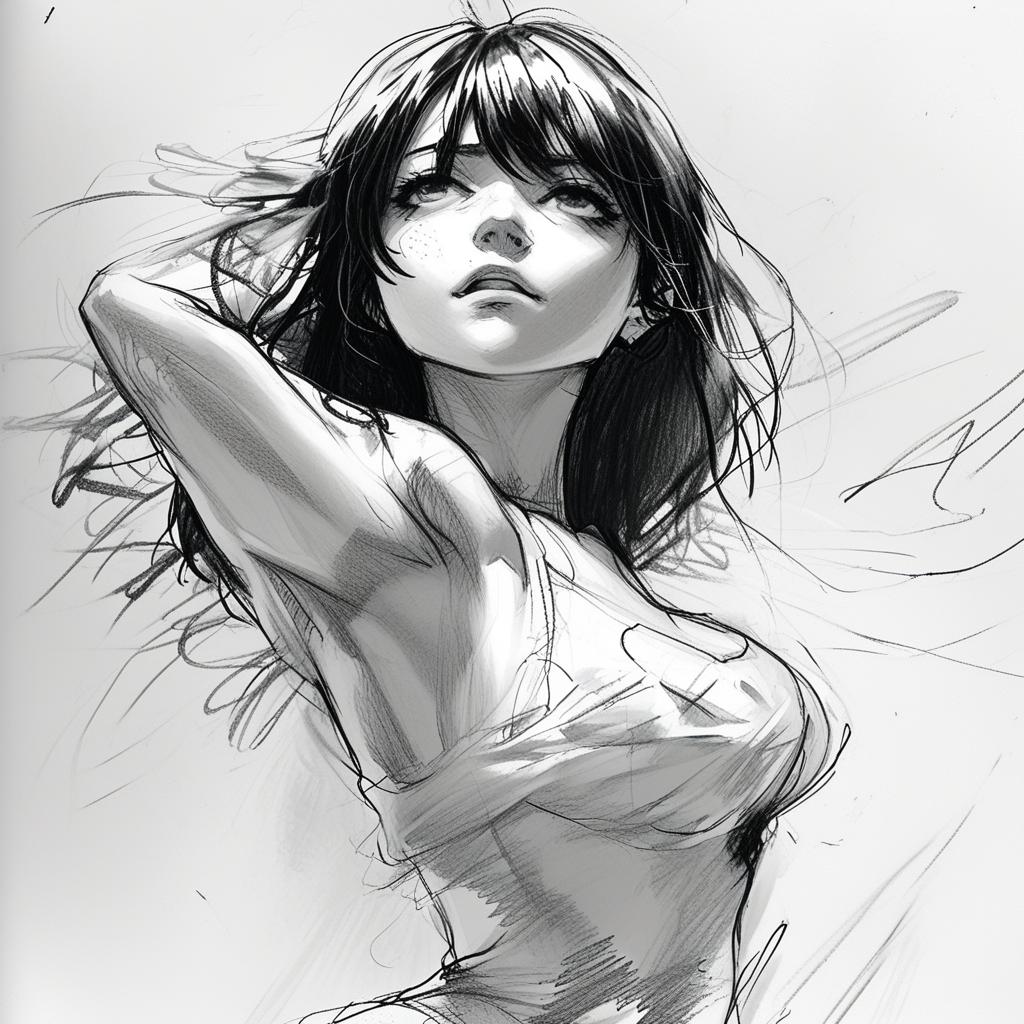} \\
        \multicolumn{4}{@{}p{0.99\textwidth}@{}}{
            PixArt-$\Sigma$: ``Sketch of manga girl in dramatic pose.''
        } \\[2mm]

        \includegraphics[width=\imgw]{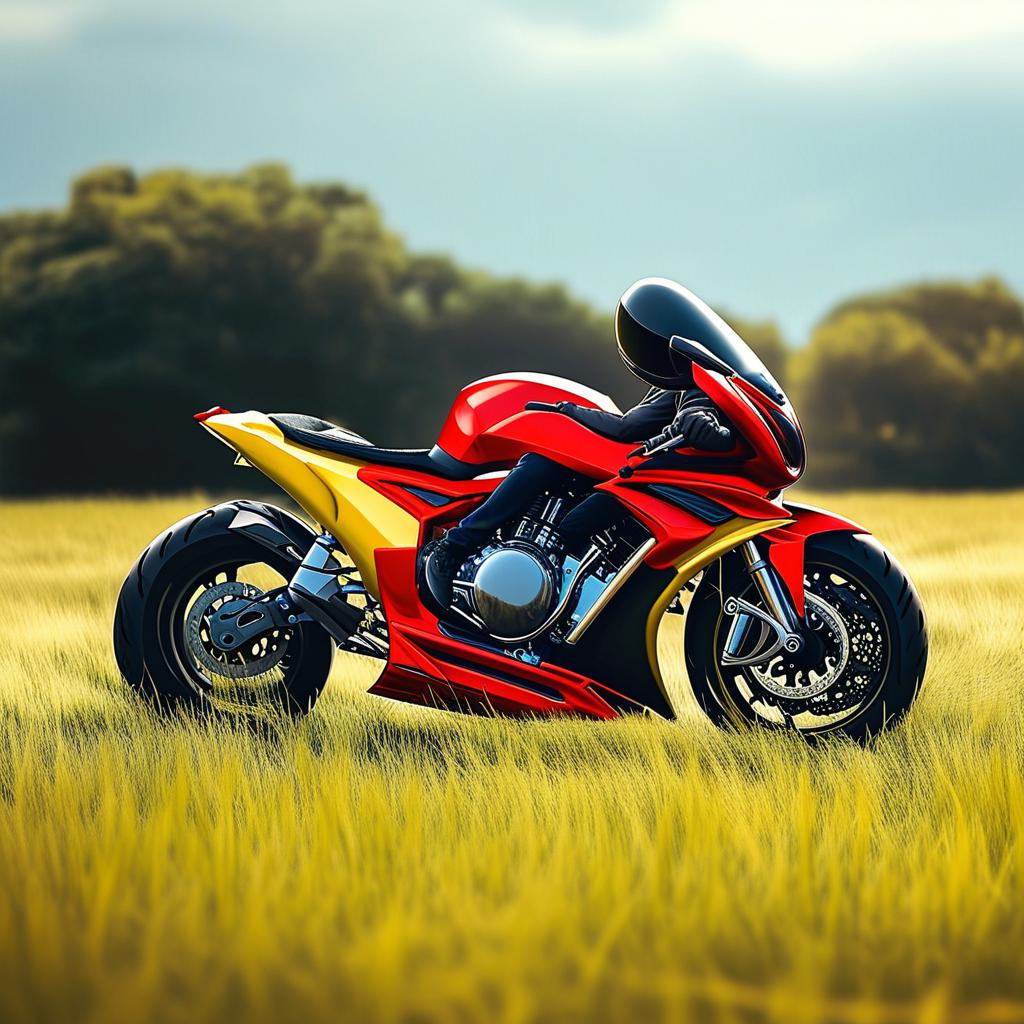} &
        \includegraphics[width=\imgw]{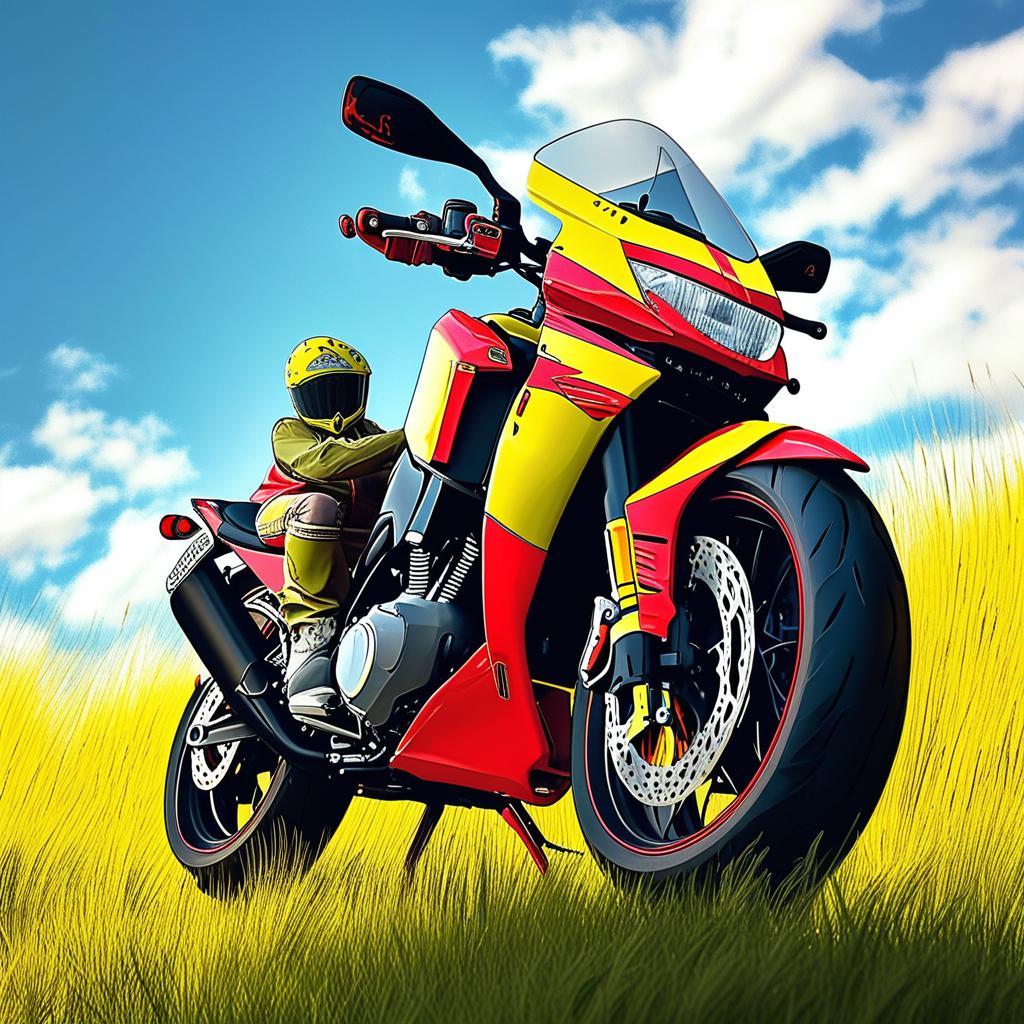} &
        \includegraphics[width=\imgw]{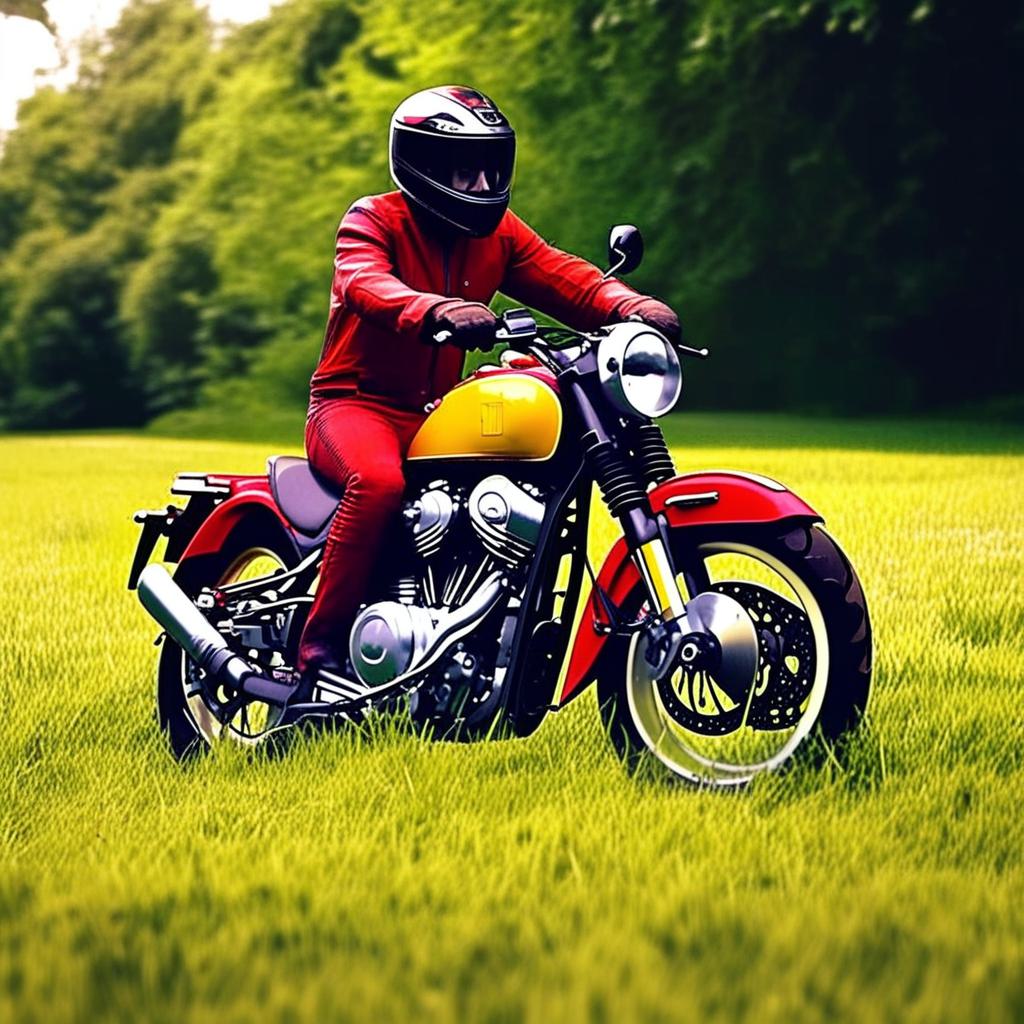} &
        \includegraphics[width=\imgw]{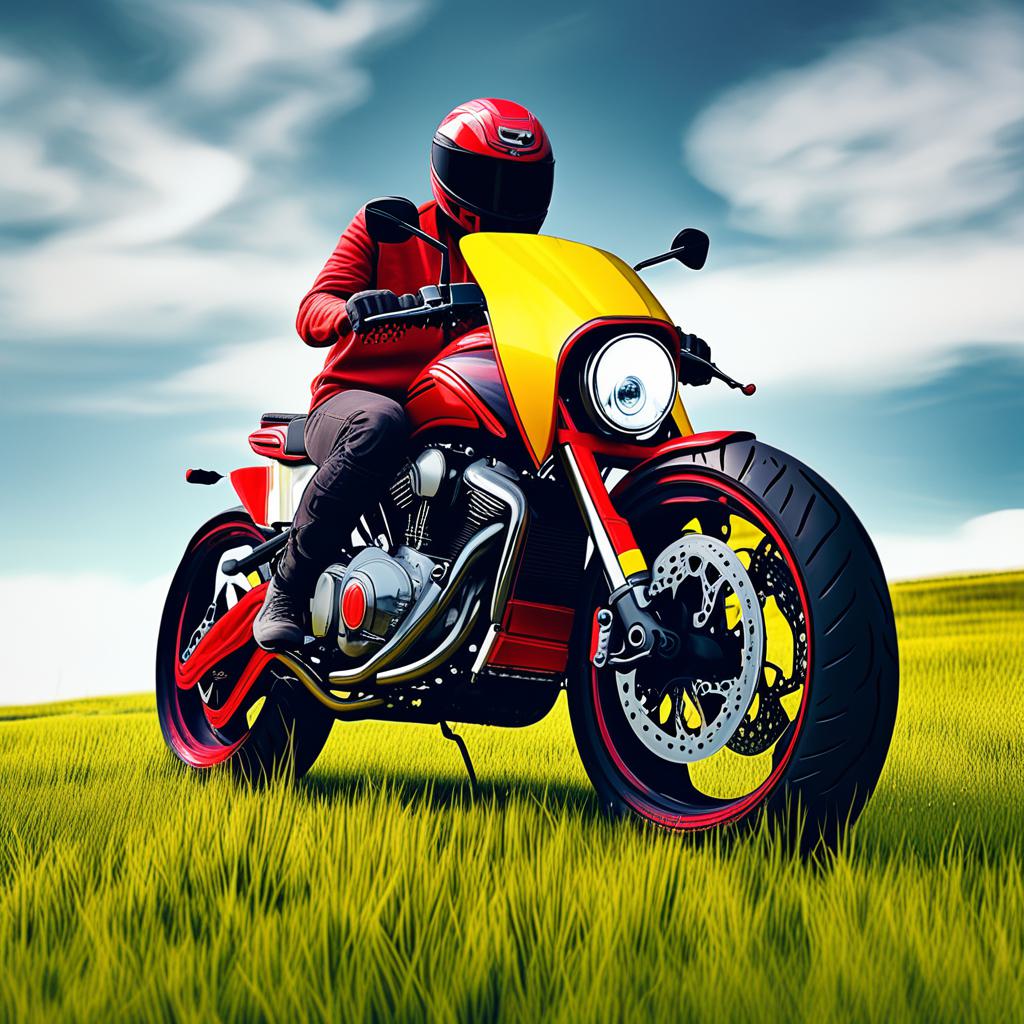} \\
        \multicolumn{4}{@{}p{0.99\textwidth}@{}}{
            PixArt-$\Sigma$: ``Yellow and red motorcycle with a man riding on it next to grass.''
        } \\

    \end{tabular}

    \caption{Additional qualitative image examples.}
    \label{fig:supp_examples}
    \vspace{-5mm}
\end{figure}

We provide additional qualitative examples on SDXL and PixArt-$\Sigma$ in Fig.~\ref{fig:supp_examples}. Consistent with the quantitative results, CA-E and EnCat track the prompt more coherently than the Standard baseline and CA-I. They render the half-robot, half-humanoid android, place both the woman \emph{and} the dog together on a tree, and seat a rider on the motorcycle, whereas the Standard and CA-I selections more often drop a requested element or produce an awkward composition. Across rows, the noises selected by CA-E and EnCat yield images that are both more prompt-faithful and more visually coherent.

\end{document}